%% file: main.tex
\documentclass[10pt,journal,compsoc]{IEEEtran}

\ifCLASSOPTIONcompsoc
\else
\fi

\usepackage{microtype}
\usepackage{graphicx}
\usepackage{subfigure}
\usepackage{color}
\usepackage{booktabs}
\usepackage{enumitem}
\usepackage[table]{xcolor}
\definecolor{Gray}{gray}{0.9}
\definecolor{LightCyan}{rgb}{0.88,1,1}
\usepackage{amsmath,amssymb} 
\usepackage{caption} 
\usepackage[pagebackref=true, colorlinks=true, allcolors=red, citecolor=green]{hyperref}

\usepackage[american]{babel}
\usepackage{xspace}

\definecolor{DarkGreen}{RGB}{1,50,32}
\definecolor{dr}{RGB}{47,85,151}
\definecolor{dreg}{RGB}{112,48,160}
\definecolor{dn}{RGB}{255, 117, 143}
\definecolor{pr}{RGB}{228,197,111}
\definecolor{kd}{RGB}{251,133,0}
\definecolor{mr}{RGB}{84,130,53}
\definecolor{tbs}{RGB}{2,48,71}

\definecolor{DeepRed}{RGB}{225,127,88}
\definecolor{DeepBlue}{RGB}{173,216,230}
\definecolor{Yellow}{RGB}{204,204,0}
\definecolor{Purple}{RGB}{128,0,128}
\definecolor{Green}{RGB}{0,128,0}

\input{math_definition}

\ifCLASSINFOpdf
\else
\fi

\begin{document}
	
\title{Representation Learning for Tabular Data: \\A Comprehensive Survey}

\author{Jun-Peng Jiang, Si-Yang Liu, Hao-Run Cai, Qile Zhou, Han-Jia Ye
	\IEEEcompsocitemizethanks{
		\IEEEcompsocthanksitem J.-P. Jiang, S.-Y Liu, H.-R Cai, Q. Zhou, and H.-J. Ye are with School of Artificial Intelligence, Nanjing University, and National Key Laboratory for Novel Software Technology, Nanjing University, Nanjing, 210023, China.
		E-mail: \{jiangjp,liusy,zhouql,yehj\}@lamda.nju.edu.cn, caihr@smail.nju.edu.cn}
}

\markboth{Journal of \LaTeX\ Class Files,~Vol.~Xx, No.~X, Xxxx~20Xx}%
{Ye \MakeLowercase{\textit{et al.}}: Deep Learning with Tabular Data: A Comprehensive Survey}

\IEEEtitleabstractindextext{%
\input{abstract}
}

\makeatletter
\DeclareRobustCommand\onedot{\futurelet\@let@token\@onedot}
\def\@onedot{\ifx\@let@token.\else.\null\fi\xspace}

\def\eg{\emph{e.g}\onedot} \def\Eg{\emph{E.g}\onedot}
\def\ie{\emph{i.e}\onedot} \def\Ie{\emph{I.e}\onedot}
\def\cf{\emph{c.f}\onedot} \def\Cf{\emph{C.f}\onedot}
\def\etc{\emph{etc}\onedot} \def\vs{\emph{vs}\onedot}
\def\wrt{w.r.t\onedot} \def\dof{d.o.f\onedot}
\def\etal{\emph{et al}\onedot}
\makeatother
\newtheorem{remark}{Remark}

\maketitle

\IEEEpeerreviewmaketitle


\input{intro}
\input{background}

\input{reason}

\input{taxonomy2}

\input{taxonomy}
\input{transfer}
\input{general}
\input{Ensemble}
\input{extensions}
\input{discussions}
\input{conclusion}



\bibliographystyle{IEEEtran}
\bibliography{main}

\end{document}

%% file: math_definition.tex
\usepackage{amssymb}
\usepackage{amsmath,amsfonts}
\usepackage{amsopn}
\usepackage{bm} 
\usepackage{multirow}
\usepackage{flushend}
\usepackage{tabularx}

\newcommand{\vct}[1]{\boldsymbol{#1}} 
\newcommand{\mat}[1]{\boldsymbol{#1}} 

\newcommand{\field}[1]{\mathbb{#1}}
\newcommand{\R}{\field{R}} 

\newcommand{\ProbOpr}[1]{\mathbb{#1}}

\newcommand{\expect}[2]{%
\ifthenelse{\equal{#2}{}}{\ProbOpr{E}_{#1}}
{\ifthenelse{\equal{#1}{}}{\ProbOpr{E}\left[#2\right]}{\ProbOpr{E}_{#1}\left[#2\right]}}} 

\newcommand{\e}{{\vct{e}}}
\newcommand{\x}{{\vct{x}}}

\newcommand{\vtheta}{\vct{\theta}}
\newcommand{\vTheta}{\vct{\Theta}}

\newcommand{\vb}{\vct{b}}

\newcommand{\mW}{\mat{W}}

\newcommand{\mT}{\mat{T}}

\newcommand{\sD}{\mathcal{D}}

\newcommand{\eat}[1]{}

\newcommand{\bbR}{\mathbb{R}}

%% file: abstract.tex
\begin{abstract}
Tabular data, structured as rows and columns, is among the most prevalent data types in machine learning classification and regression applications. Models for learning from tabular data have continuously evolved, with Deep Neural Networks (DNNs) recently demonstrating promising results through their capability of representation learning. 
In this survey, we systematically introduce the field of tabular representation learning, covering the background, challenges, and benchmarks, along with the pros and cons of using DNNs.
We organize existing methods into three main categories according to their generalization capabilities: specialized, transferable, and general models. Specialized models focus on tasks where training and evaluation occur within the same data distribution. We introduce a hierarchical taxonomy for specialized models based on the key aspects of tabular data---features, samples, and objectives---and delve into detailed strategies for obtaining high-quality feature- and sample-level representations.
Transferable models are pre-trained on one or more datasets and subsequently fine-tuned on downstream tasks, leveraging knowledge acquired from homogeneous or heterogeneous sources, or even cross-modalities such as vision and language. 
General models, also known as tabular foundation models, extend this concept further, allowing direct application to downstream tasks without additional fine-tuning. We group these general models based on the strategies used to adapt across heterogeneous datasets.
Additionally, we explore ensemble methods, which integrate the strengths of multiple tabular models. Finally, we discuss representative extensions of tabular learning, including open-environment tabular machine learning, multimodal learning with tabular data, and tabular understanding tasks. More information can be found in the following repository: \url{https://github.com/LAMDA-Tabular/Tabular-Survey}.
\end{abstract} 

%% file: intro.tex
\section{Introduction}\label{sec:intro}
Tabular data, characterized by structured rows and columns, is one of the most prevalent data formats in real-world machine learning applications, spanning diverse domains such as finance~\cite{kovalerchuk2005data}, healthcare~\cite{hyland2020early}, education~\cite{romero2010educational}, recommendation systems~\cite{amatriain2010data}, and scientific research. In particular, AI for scientific research (AI4science) has increasingly relied on tabular data, as numerous prominent datasets---such as those from genomics~\cite{tibshirani2002diagnosis}, chemistry~\cite{ivanciuc2007applications}, and climate science~\cite{ahmed2010empirical,allen2002towards}---naturally adopt tabular forms.

Tabular data inherently organizes information in a {\em structured, table-like format}. 
In this survey, we focus primarily on supervised tabular machine learning tasks, specifically classification and regression.
\begin{figure}[t]
    \includegraphics[width=0.99\columnwidth]{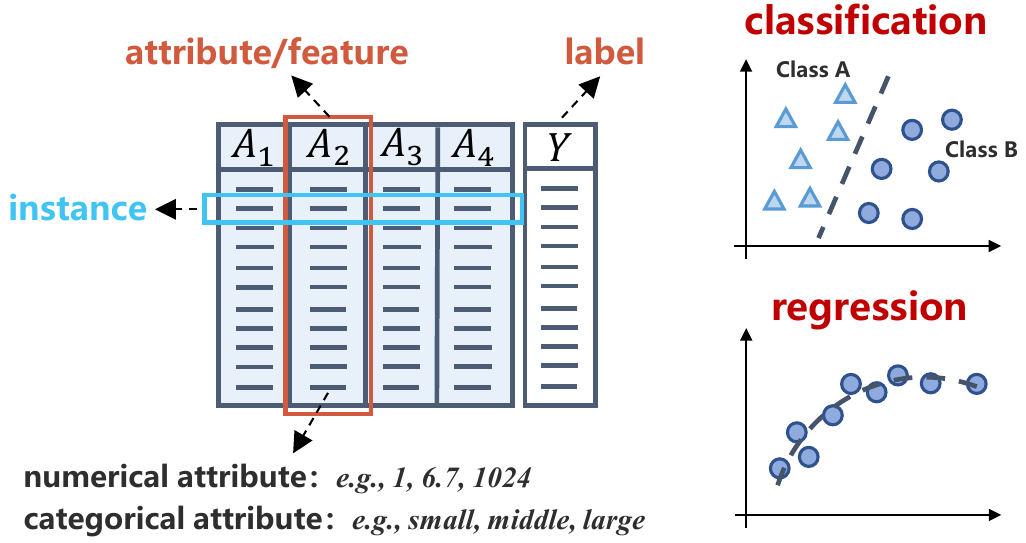}
    \caption{A brief introduction to tabular data and associated learning tasks. Each row represents an instance and each column corresponds to a specific attribute or feature, which can be numerical or categorical. The most common tabular machine learning tasks are classification and regression as shown in the right side of the the figure.} 
    \label{figure:intro}
    \vspace{-5mm}
\end{figure}
Beyond their structured organization, tabular datasets frequently include heterogeneous attributes~\cite{Borisov2024Deep}, encompassing numerical, categorical, or mixed data types that may be either dense or sparse. Additionally, many tabular datasets present quality challenges, such as noisy measurements, missing values, outliers, inaccuracies~\cite{Aggarwal15DMBook}, and privacy constraints~\cite{Ji2014DP}, all of which complicate the modeling process. The most common supervised tabular tasks are classification and regression, where the goal is to learn mappings from training data to discrete or continuous targets, respectively. As illustrated in \autoref{figure:intro}, each row represents an instance (with its corresponding label), while each column corresponds to a specific attribute or feature~\cite{DelgadoCBA14}. Ideally, learned mappings should generalize effectively, accurately predicting outcomes for new instances drawn from the same underlying distribution.

\begin{figure*}[t]
    \includegraphics[width=0.99\textwidth]{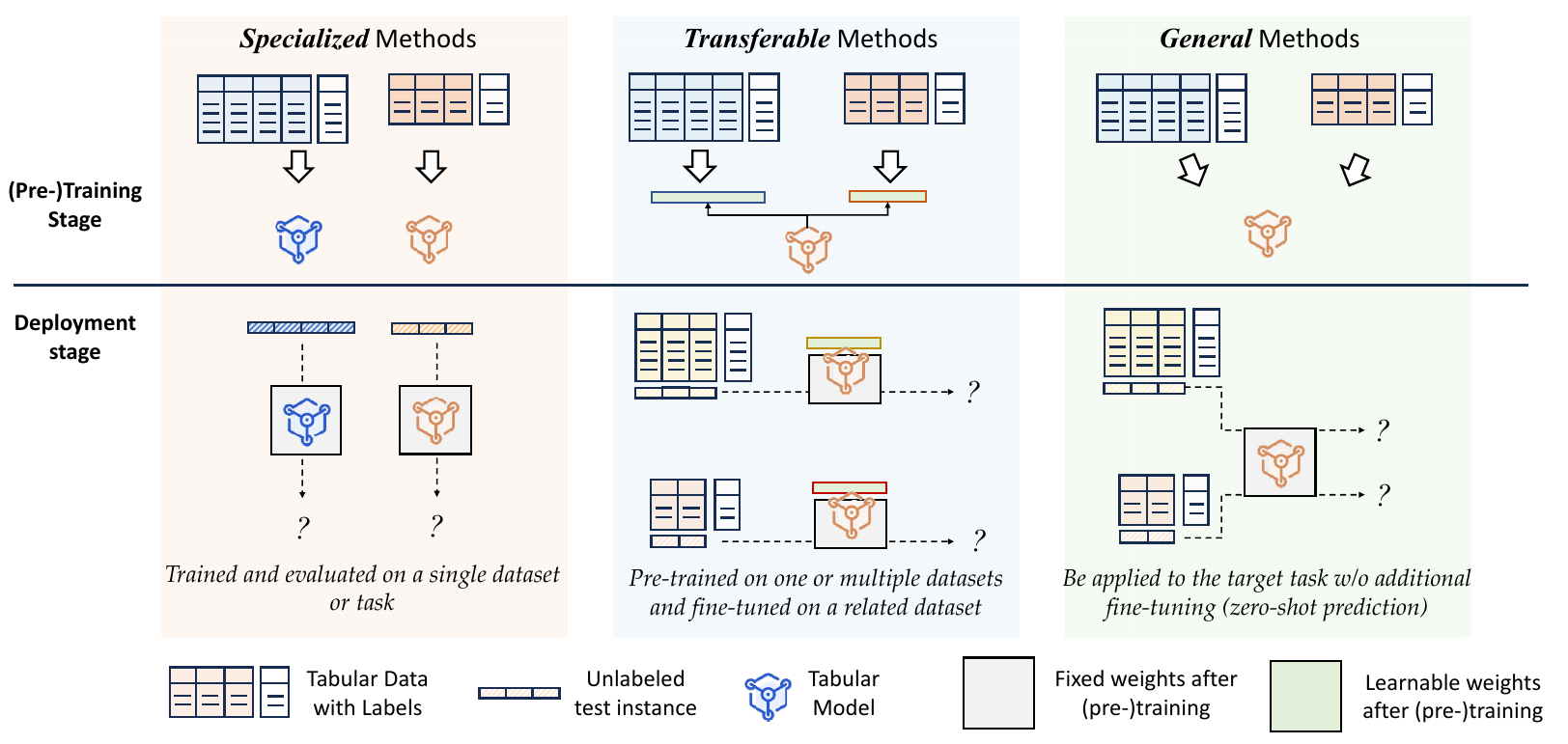}
    \caption{
    We organize existing tabular classification/regression methods into three categories according to their generalization capabilities: specialized (left), transferable (middle), and general (right) models.
    Specialized models focus on tasks where training and evaluation occur within the same data distribution.
    Transferable models are pre-trained on one or more datasets and subsequently fine-tuned on downstream tasks.
    General models, also known as tabular foundation models, extend this concept further, allowing direct application to downstream tasks without additional fine-tuning.} 
    \label{figure:intro2} 
    \vspace{-5mm}
\end{figure*}

Machine learning methods for tabular data have evolved significantly over the years~\cite{bishop2006pattern,HastieTF09ESL,Mohri2012FoML,Murphy2012PML}. Recently, the rise of deep learning has profoundly impacted domains like computer vision~\cite{voulodimos2018deep} and natural language processing~\cite{otter2020survey}, where Deep Neural Networks (DNNs) extract semantic \emph{representations} directly from raw inputs~\cite{bengio2013representation,lecun2015deep,goodfellow2016deep}. These learned representations have not only improved generalization but have also facilitated knowledge transfer across related tasks~\cite{Donahue2014Decaf}. The flexibility of DNNs in modeling complex feature interactions and learning rich hierarchical structures has inspired significant interest in adapting deep learning techniques to tabular data.

Indeed, DNNs were applied to tabular data decades ago, initially targeting dimensionality reduction and visualization tasks~\cite{hinton2006reducing,weston2008deep,van2009learning,min2010deep}, yet they typically struggled to match tree-based methods on standard classification and regression problems. Later advances in DNNs have led to significant improvements across various tabular-related applications, such as click-through rate prediction~\cite{ZhangDW16Deep,Cheng2016Wide}, anomaly detection~\cite{mehrotra2017anomaly}, recommendation systems~\cite{isinkaye2015recommendation}, and time series forecasting~\cite{Rangapuram2018DeepSSM,lim2021time}. Modern deep learning approaches, benefiting from better-designed architectures, optimized training strategies, high-quality representations, have revitalized DNN performance on tabular data, often rivaling or surpassing traditional tree-based models~\cite{GorishniyRKB21Revisiting,David2024RealMLP,Ye2025ModernNCA}. Given the wide variety of approaches emerging in deep tabular modeling, a systematic overview that revisits critical factors and current methodologies in representation learning for tabular data has become increasingly necessary.

This survey begins by introducing the background of tabular data learning, highlighting the challenges involved and critically examining the advantages and limitations of utilizing DNNs compared to classical---particularly tree-based---methods~\cite{Grinsztajn2022Why,ZivA22Tabular,EgeInductiveBias,McElfreshKVCRGW23when}. Given the observed instability of method performance across different tabular datasets, we also discuss comprehensive strategies for dataset collection, evaluation, and analysis, aiming to establish robust criteria for aggregating performance metrics across multiple datasets~\cite{ye2019learning,Jesus2022Turning,kohli2024towards,Tschalzev2024DataCentric}.

We broadly categorize deep tabular methods into three types: {\em specialized methods, transferable methods, and general methods}, distinguished by the scope of datasets on which they are trained and deployed, as well as their corresponding generalization capabilities (illustrated in~\autoref{figure:intro2}). Specialized tabular methods align closely with classical supervised models, typically trained and evaluated on data drawn from the same distribution. In contrast, transferable methods leverage knowledge from models pre-trained on one or multiple source datasets, subsequently fine-tuning these models on target datasets; the primary challenge here lies in addressing the heterogeneity between pre-trained sources and target tasks. The recently proposed general tabular methods---motivated by the remarkable ``zero-shot'' generalization abilities demonstrated by large language models (LLMs)---exhibit exceptional versatility. These general models can directly apply their learned representations to downstream tabular datasets without additional fine-tuning, achieving robust generalization due to advanced pre-training strategies. Although the generalization ability tends to increase from specialized to general models, it does not imply that specialized or transferable methods are less valuable; specialized models remain superior on large-scale datasets, and fine-tuning general models can further improve their predictive performance. Additionally, the first two types of methods provide foundational insights and valuable components that contribute significantly to advancements in general tabular models.

For {\bf specialized methods}, numerous designs have been proposed from diverse perspectives, and previous papers have often categorized these methods based primarily on their architectural characteristics or behaviors. Existing taxonomies~\cite{Ye2024Closer}, for example, group specialized methods into feature-preprocessing-based~\cite{GorishniyRKB21Revisiting,Gorishniy2022On}, data-augmentation-based~\cite{UcarHE21SubTab,BahriJTM22Scarf,YoonZJS20VIME,Wu2024SwitchTab}, MLP variants~\cite{Kadra2021Well,David2024RealMLP}, specialized DNN architectures~\cite{WangFFW17DCN,KlambauerUMH17Self,ke2018tabnn,WangSCJLHC21DCNv2,ChenLWCW22DAN,Chen2023TabCaps,Chen2024Team,Xu2024BiSHop}, tree-mimic approaches~\cite{Badirli2020GrowNet,PopovMB20Neural,Chang0G22NODEGAM}, token-based techniques~\cite{SongS0DX0T19AutoInt,Huang2020TabTransformer,GorishniyRKB21Revisiting,Zhou2023TabToken,Chen2023Excel}, regularization-driven methods~\cite{jeffares2023tangos,PTARL}, and neighborhood-based strategies~\cite{NaderSL22DNNR,gorishniy2023tabr,Ye2025ModernNCA}. However, such categorizations can appear scattered, making it difficult to connect the core ideas between methods placed in distinct groups.
In contrast, this survey introduces a hierarchical taxonomy based on the key aspects of tabular data---features, samples, and objectives---providing a cohesive organizational framework. Our approach emphasizes detailed strategies for obtaining high-quality representations at both feature- and sample-levels. This unified perspective helps bridge core ideas across diverse methods, facilitating clearer comparative discussions and potentially guiding the design of future, more advanced tabular models.

Instead of training a model from scratch on a single tabular dataset, {\bf transferable models} leverage knowledge encoded in a pre-trained model from another dataset, which can significantly enhance the training process, especially when data or computational resources for the target task are limited. A major challenge in transferring knowledge across tabular tasks lies in the inherent heterogeneity between the source and target datasets, particularly differences in their feature and label spaces.
In this survey, we adopt a broad perspective on transferable tabular models, categorizing methods based on the sources of their pre-trained knowledge. Specifically, we discuss models pre-trained on homogeneous tabular domains, such as self-supervised methods with additional pre-training steps on the target dataset itself~\cite{Somepalli2021SAINT,Rubachev2022revisiting}; models pre-trained across heterogeneous tabular domains~\cite{Onishi2023TabRet,shen2023cross,Zhou2023TabToken}; and methods transferring knowledge from other modalities, such as vision-based pre-trained models~\cite{Zhu2021IGTD,Lee2023TablEye,Mamdouh2025Tab2Visual}. Additionally, since incorporating attribute semantics (when available) is a common strategy for bridging heterogeneous attribute spaces across tabular datasets~\cite{Wang2022TransTab,Yan2024Making,Ye2024Towards}, we also explore approaches leveraging language models in the final category. In particular, we further organize these language model-based strategies according to the methods they use to extract knowledge and the types of language models involved---ranging from small-scale language models to Large Language Models (LLMs)~\cite{Hegselmann2022TabLLM,Wen2024Generative,Hollmann2023CAAFE,Han2024FeatLLM}.

Inspired by recent advancements in foundation models from vision and language domains~\cite{zhou2024comprehensive,LiangWNJ0SPW24FoundationTS}, {\bf general models}---also known as tabular foundation models---expand the concept of transferable tabular models by enabling direct application to downstream tasks without additional fine-tuning. This capability, commonly referred to as the model's ``zero-shot'' ability, significantly enhances the model’s usability across diverse tabular datasets.
In contrast to transferable models, which primarily focus on bridging knowledge gaps between source and target datasets, general models aim to construct highly adaptive architectures capable of handling a wide array of heterogeneous datasets simultaneously. We categorize these general models based on the strategies used to achieve adaptiveness across diverse tabular tasks, specifically examining adaptations from both data-centric~\cite{Ye2023TabPTM} and model-centric perspectives~\cite{BonetMGI2024HyperFast,Muller2023MotherNet}. Furthermore, we discuss critical branches of general tabular models in detail: the TabPFN variants leveraging in-context learning~\cite{Hollmann2022TabPFN,Thomas2024LocalPFN,hollmann2025tabpfn}, and methods utilizing attribute and task semantics to unify heterogeneous tasks within a common representation framework~\cite{tabula8b,WenZZXB24From,Wen2025ICL}.

Additionally, ensemble methods~\cite{gorishniy2024tabm,Liu2025Beta,hollmann2025tabpfn} are introduced, which improve the generalization ability based on the strengths of multiple tabular models.
Finally, we briefly overview other relevant extensions of tabular learning, including clustering~\cite{SvirskyL24Interpretable,Rauf2024TableDC}, anomaly detection~\cite{Han2022ADBench,ShenkarW22Anomaly,YinQZW024MCM}, data generation and imputation~\cite{HansenSSP23Reimagining,HouGXQ23Incremental,VeroBV24CuTS}, interpretability~\cite{Huang2020TabTransformer,ArikP21TabNet,Chang0G22NODEGAM}, multimodal learning~\cite{hager2023best,JiangYW00Z24Tabular}, open-environment tabular machine learning~\cite{Diao2024OEBench,Rubachev2024TabRed,gardner2024benchmarking,zhou2024core}, and tabular understanding~\cite{jin2022survey,fang2024large}. By summarizing the state of the field and identifying open challenges, we aim to guide future research and applications in tabular data representation learning.

%% file: background.tex
\section{Background}\label{sec:background}
This section presents the (supervised) tabular machine learning task, including the notation of tabular data learning, the history of tabular data, the challenges of learning from tabular data, evaluation metrics, and tabular benchmarks.

\subsection{Learning with Tabular Data}
A supervised tabular dataset is formatted as $N$ examples and $d$ features/attributes corresponding to $N$ rows and $d$ columns in the table. An instance $\x_i\in\bbR^d$ is depicted by its $d$ feature values. 
Assume $x_{i,j}$ as the $j$-th feature of instance $\x_i$, it could be a numerical (continuous) one $x_{i,j}^{\textit{\rm num}}\in\bbR$, like the temperature of a region or the density of the object. 
$\x_i$ can also be a categorical (discrete) value $x_{i,j}^{\textit{\rm cat}}$, like one of multiple colors, the location of a person, or even some textual descriptions of the instance. 
Each instance is associated with a label $y_i$, where $y_i\in \{1,-1\}$ in a binary classification task, $y_i\in [C]=\{1,\ldots,C\}$ in a multi-class classification task, and $y_i\in \bbR$ in a regression task. 

\begin{remark}
    Ordinal regression~\cite{winship1984regression,GutierrezPSFH16}, also called ordinal classification, is a type of regression analysis used to predict an ordinal variable. It can be considered an intermediate problem between regression and classification. However, this survey primarily focuses on standard classification and regression tasks and does not specifically discuss ordinal regression.
\end{remark}

Given a tabular dataset $\sD=\{(\x_i, y_i)\}_{i=1}^N$, we aim to learn a mapping $f$ on $\sD$ that maps $\x_i$ to its label $y_i$. In other words, the model predicts $\x_i$ with $\hat{y}_i=f(\x_i)$. The general objective learning $f$ follows the structural risk minimization:
\begin{equation}
\min_f \; \sum_{(\x_i, y_i)\in\sD}  \ell(y, \;\hat{y}_i=f(\x_i)) + \Omega(f)\;.\label{eq:objective}
\end{equation}
$\ell(\cdot, \cdot)$ measures the discrepancy between the predicted label $\hat{y}_i$ and the true label $y_i$, \eg, cross-entropy in classification and mean square error in regression. 
$\Omega(\cdot)$ is the regularization on the model, which restricts the complexity of $f$.
We expect the learned $f$ is able to extend its ability to {\em unseen} instances sampled from the same distribution as $\sD$. 

Tabular methods differ in their strategies to implement $f$. The ``dummy'' approach makes predictions based on training labels $\{y_i\}_{i=1}^N$ directly, which outputs the major class in the training set for classification and the average of all labels for regression, respectively.

In a $C$-class classification task, classical parametric methods implement $f$ with a linear mapping, \ie, $f(\x_i) = \mW^\top \x_i + \vb$, where the classifier $\mW\in\mathbb{R}^{d\times C}$ and $\vb\in\mathbb{R}^C$ is the bias. With different loss functions, we can implement Logistic Regression, SVM, or even AdaBoost. In contrast, non-parametric methods implement the prediction via $f(\x_i)=f(\x_i, \sD)$, depending on the whole training set. For example. KNN searches neighbors in the training set $\sD$ with the $K$ smallest distance w.r.t. $\x_i$. KNN can be viewed as a specific label smoother, with a dynamic local region for every instance. \cite{Jeffares2024Telescoping} links KNN and Random Forest from their ways of smoothing training labels in their predictions. 

Deep tabular methods implement $f$ with a deep neural network, \eg. Most deep learning models could be decomposed into two parts, \ie, $f(\x_i) = \mW^\top \phi(\x_i) + \vb$. Similar to the linear model, $\mW$ and $\vb$ are the components of linear classifier, with $\mW\in\mathbb{R}^{d'\times C}$. $\phi$ maps the input vector $\x_i$ into the $d'$ dimension space, which extracts semantic embeddings for the given tabular input. $\phi$ could be implemented with MLP or residual network. 

\subsection{History of Tabular Data}
Historically, classical machine learning tasks were predominantly formulated with tabular data, or datasets readily transformed into a tabular representation without explicitly designating them as ``tabular.'' In early literature, the term ``tabular'' typically referred to tables within relational databases~\cite{Cormode2002Mining}, CSV files on the web~\cite{Adelfio2013Schema}, or tables in documents~\cite{Arias1996Efficient}. Relevant tasks included table extraction~\cite{Wang2000Semantic}, parsing~\cite{Nederhof94Optimal}, understanding~\cite{Arias1996Interpreting}, and discovering association rules~\cite{Richards2001Discovery}. With the expansion of machine learning applications into other modalities such as images, texts, audio, and video, the classical vector-based data representations have come to be explicitly termed ``tabular data.'' 

Early statistical approaches such as linear regression, logistic regression, linear discriminant analysis, and K-Nearest Neighbors (KNN) predate artificial intelligence. Classical learning methods further expanded across various paradigms, including decision trees~\cite{quinlan1986induction,breiman1984classification}, multi-layer perceptrons (MLPs), support vector machines (SVMs), and nearest centroid classifiers~\cite{tibshirani2002diagnosis,HastieTF09ESL}. Ensemble methods enhanced predictive performance by aggregating outputs from multiple base learners~\cite{freund1995desicion,Breiman01RandomForest}. More recently, gradient boosting frameworks~\cite{friedman2001greedy,friedman2002stochastic}, such as XGBoost~\cite{chen2016xgboost}, LightGBM~\cite{ke2017lightgbm}, and CatBoost~\cite{Prokhorenkova2018Catboost}, have become prominent due to their effectiveness and efficiency in tabular data applications and competitions~\cite{nielsen2016tree,makridakis2022m5}.

With the development of deep learning, DNNs were applied to tabular classification and regression tasks decades ago, utilizing architectures such as stacked Restricted Boltzmann Machines and denoising autoencoders~\cite{larochelle2007empirical,salakhutdinov2007learning,min2009deep}. Early representation learning efforts primarily focused on dimensionality reduction and data visualization tasks~\cite{hinton2006reducing,weston2008deep,van2009learning,min2010deep}, yet these models struggled to surpass traditional tree-based methods in terms of generalization. However, advancements in neural network architectures and representation learning strategies have recently led to promising results in related tabular domains, including click-through rate prediction~\cite{ZhangDW16Deep,Cheng2016Wide}, anomaly detection~\cite{ahmed2016survey,mehrotra2017anomaly}, recommendation systems~\cite{lu2012recommender,isinkaye2015recommendation}, and time series forecasting~\cite{Rangapuram2018DeepSSM,salinas2020deepar,lim2021time,huang2025seqfusion}. Innovations such as convolutional layers and learnable feature embeddings have improved the ability of deep models to capture high-order attribute relationships~\cite{Liu2015CCPM,GuoTYLH17DeepFM}. While early deep tabular methods lagged behind ensemble tree-based models, recent techniques have demonstrated competitive or superior performance~\cite{GorishniyRKB21Revisiting,David2024RealMLP,Ye2025ModernNCA}, affirming deep representation learning as a promising direction for tabular data modeling.

While several survey papers have been published~\cite{Borisov2024Deep,somvanshi2024survey}, the field of tabular data has witnessed remarkable progress over the past two years. On one hand, the emergence of new specialized methods has introduced significant shifts in the landscape, motivating the need for our comprehensive taxonomy. On the other hand, the rise of transferable and general approaches has greatly enhanced the generality and applicability of tabular data modeling, which has been overlooked in previous works.

\subsection{Challenges of Learning from Tabular Data} 
Different from other types of data sources, \eg, images and texts, there exist several challenges dealing with tabular datasets due to their characteristics.

\noindent{\bf Heterogeneity of Features.} Unlike continuous image data or token-based textual data, tabular datasets often contain both numerical and categorical attributes, each requiring distinct handling methods~\cite{Borisov2024Deep,lane2003introduction}. Numerical features frequently exhibit varying ranges and distributions, necessitating normalization or scaling. Categorical features differ in cardinality and semantic interpretation, requiring encoding methods like one-hot vectors or embeddings. Consequently, tabular models must carefully handle these mixed data types to preserve the usability of each feature.

\noindent{\bf Lack of Spatial Relationships.} Tabular data inherently lacks spatial or sequential relationships that are naturally found in other modalities~\cite{Zhu2021IGTD,Kadra2021Well}. The order of columns has no semantic or spatial meaning, making tabular data permutation-invariant regarding features. Moreover, standard tabular machine learning assumes rows are independently and identically distributed ({\em i.i.d.}), further eliminating temporal or sequential correlations present in data such as video or time series. This absence of inherent spatial or sequential structure challenges deep learning architectures traditionally designed to exploit such dependencies.

\noindent{\bf Low-quality and Missing Data.} Compared to image or text data, where contextual or spatial redundancies help manage missing or corrupted values, tabular data is more vulnerable to incomplete or erroneous entries~\cite{karr2006data,sanchez2020improving}. Missing values in tabular datasets can introduce significant biases and degrade prediction quality. Additionally, noisy or incorrect values can considerably affect model reliability. Data preprocessing steps, including data cleaning and imputation, become crucial to maintaining accuracy and robustness in tabular machine learning.

\noindent{\bf Importance of Feature Engineering.} Effective tabular models heavily depend on the quality of their input features~\cite{Gorishniy2022On,chicco2022eleven}. Unlike image or textual data, where DNNs inherently learn feature representations from raw data, tabular methods often require domain-specific knowledge and meticulous manual feature engineering. Identifying and modeling complex, nonlinear interactions among tabular features frequently demands sophisticated transformations and expert insight, significantly impacting the predictive performance of models~\cite{luo2019autocross}.

\noindent{\bf Class Imbalance.} Tabular datasets frequently exhibit imbalanced label distributions, especially in classification tasks, where certain categories are underrepresented~\cite{he2009learning,he2013imbalanced}. Class imbalance complicates model learning, leading to biased outcomes toward majority classes and poor performance on minority classes. Specialized methods such as oversampling, undersampling, or tailored loss functions (\eg, focal loss~\cite{LinGGHD17-FocalLoss}) are required to address this imbalance effectively. Evaluation criteria like the AUC or F1-score further help assess model quality in imbalanced settings. Recent research highlights differences between deep and classical models in handling imbalance, emphasizing the need for careful consideration~\cite{johnson2019survey,engelmann2021conditional,sauber2022use,Jesus2022Turning}.
\begin{remark}
    Class imbalance has long been a known issue in the tabular domain, even before the rise of deep learning~\cite{liu2008exploratory}, and methods such as SMOTE~\cite{Chawla2002SMOTE,Alberto2018SMOTE} can easily be extended to deep learning methods during preprocessing. 
    However, Current deep tabular methods primarily assume that the training and testing data come from the same distribution, even in cases involving class imbalance. 
    In addition, some class imbalance methods in visual domain can be readily extended to the tabular data learning~\cite{cao2019learning,cui2019class}. Therefore, we do not delve into class imbalance in this survey.
\end{remark}

\noindent{\bf Scalability to Large Datasets.} Tabular datasets can become large-scale and high-dimensional, presenting computational and generalization challenges~\cite{xie2021fives}. With increasing dimensionality, the risk of overfitting increases, especially when the number of features significantly surpasses the number of samples. Consequently, efficient training algorithms, memory management strategies, and sufficient computational resources become essential. Effectively scaling tabular models to handle large datasets while maintaining generalization ability remains a challenging but critical research area~\cite{hu2024annotatedtables}.

\noindent{\bf Model Selection and Hyperparameter Tuning.} Tabular models are particularly sensitive to hyperparameter settings~\cite{klein2019tabular,pokhrel2023comparison}. Selecting an appropriate model architecture and tuning hyperparameters, such as learning rate, layer depth, or number of trees, can be computationally expensive and time-consuming. Despite the advancement of automated machine learning (AutoML) techniques~\cite{hutter2019automated,he2021automl,feurer2022auto}, efficiently identifying optimal configurations for deep tabular methods under practical constraints remains challenging and critical for achieving high predictive performance.

\noindent{\bf Domain-Specific Constraints.} Certain application domains, such as healthcare or finance, impose additional regulatory or ethical requirements on model development~\cite{mennella2024ethical}. For example, healthcare applications must comply with privacy standards like HIPAA~\cite{moore2019HIPAA} and provide explainability to clinicians. Financial models similarly must adhere to fairness regulations and industry standards. These constraints can restrict algorithm selection, necessitate interpretable predictions, and require additional validation, explainability, and auditability procedures~\cite{sittig2011legal,amann2020explainability,caffo2022explainable}.

\subsection{Evaluation of a Tabular Method}
We present the evaluation of tabular methods, ranging from traditional to modern, to provide a comprehensive evaluation across different aspects.
For a given model on a dataset $\sD$, we employ standard metrics that quantify the discrepancy between the predicted label $\hat{y}_i$ and the true label $y_i$.

\noindent{\bf Evaluation on A Single Task.} 
For classification tasks, Accuracy (or Error Rate) is commonly employed as the primary metric.  
AUC and F1 scores are further used to address imbalanced label distributions, while Expected Calibration Error (ECE)~\cite{guo2017calibration,helli2024drift} calculates the weighted average error of the estimated probabilities. All criteria are the higher, the better, except the error rate and ECE. 
For regression tasks, common metrics include Mean Squared Error (MSE), Mean Absolute Error (MAE), and Root Mean Squared Error (RMSE), with MAE and RMSE sharing the scale of the original labels. Lower values denote superior performance. Additionally, the coefficient of determination (R$^2$) is employed, with higher values indicating a better fit.

In tabular machine learning, the diversity of datasets makes it difficult for any single model to consistently excel across all scenarios. Therefore, evaluating models requires not only assessing their performance on individual datasets but also employing aggregated metrics that capture their overall effectiveness across multiple datasets.

\noindent{\bf Evaluation on A Set of Tasks.} 
Early research predominantly relied on \textit{Average Rank} (Friedman Rank)~\cite{DelgadoCBA14,McElfreshKVCRGW23when}, often used in conjunction with \textit{Critical Difference Comparisons}, to evaluate model performance across multiple datasets. Models are ranked per dataset based on a chosen metric (\eg, accuracy, AUC, RMSE), and the average rank is computed across datasets. To ensure statistical robustness, hypothesis tests were employed to assess the significance of ranking differences, providing a more reliable comparative analysis. For multiple comparisons, tests such as the Wilcoxon-Holm, Fredman, and Nemiyi tests are employed~\cite{Demsar06Statistical}. To address the potential degradation of average rank by poor performance on some datasets, the \textit{Probability of Achieving the Maximum Accuracy}~(PAMA)~\cite{DelgadoCBA14} is defined as the fraction of datasets in which a model attains the highest accuracy. An alternative to PAMA accounts for near-optimal performance, \textit{P95} quantifies the likelihood of a model attaining at least 95\% of the maximum accuracy, which is computed as the ratio of datasets where the classifier achieves at least 95\% of the maximum accuracy to the total number of datasets.

As research progressed, more diverse evaluation metrics were introduced. The \textit{Arithmetic Mean} of a chosen metric provides a direct comparison across datasets, but variations in the scales of evaluation metrics across datasets can distort results. To mitigate this issue, performance metrics are often normalized before aggregation, with \textit{normalized Accuracy} applied to classification tasks and \textit{normalized RMSE} (nRMSE) used for regression~\cite{Grinsztajn2022Why,David2024RealMLP}. Depending on the evaluation framework, \textit{Mean Normalized Error} can be used, but its dependence on normalization can hinder independent optimization. To further address these limitations, the \textit{Shifted Geometric Mean} (SGM) error was introduced, which aggregates errors multiplicatively, reducing sensitivity to extreme values and ensuring more stable cross-datasets/splits comparisons~\cite{David2024RealMLP}.

Beyond absolute performance, relative comparisons are also important. The \textit{Relative Improvement} metric quantifies a model’s performance gain over a baseline (\eg, a simple MLP), offering insight into efficiency relative to simpler alternatives~\cite{Yury2024TabM}. More recently, drawing inspiration from the ELO rating system\cite{glickman1999rating,hvattum2010using}, ELO-based evaluation has been introduced~\cite{Ma2024TabDPT}, modeling model-to-model comparisons as pairwise competitions across datasets. The \textit{ELO Score} iteratively adjusts rankings based on relative performance, providing a more dynamic, fine-grained assessment.

\subsection{Tabular Benchmarks and Datasets}
This section introduces existing benchmarks and datasets, along with associated considerations for constructing the benchmarks and evaluation protocols. 

\subsubsection{Popular Tabular Benchmarks and Datasets}
We first introduce several benchmarks based on raw features constructed from various aspects. Then, we present datasets with rich semantics, following some tabular toolboxes and evaluation protocols.

\noindent{\bf Standard Benchmarks.}
Methods for tabular data have preferences depending on the dataset, and evaluating them on limited datasets can be easily influenced by randomness or other factors. Therefore, it's important to consider various aspects to ensure a more comprehensive and reliable benchmark evaluation.

A comprehensive benchmark should cover a diverse set of datasets to test the model's generalization capabilities across different tasks and feature types. The benchmark should include datasets from different task types, including binary classification, multi-class classification, and regression tasks. \cite{DelgadoCBA14} evaluates 179 classifiers across 17 families on 121 datasets, concluding that Random Forest variants were the most likely to perform best overall. 
\cite{Kadra2021Well} explores MLPs with parameterized techniques, such as ensembling and data augmentation, over 40 classification datasets. Similarly, \cite{GorishniyRKB21Revisiting} demonstrates the effectiveness of MLPs, ResNets, and transformer-based models on 11 datasets.
\cite{Grinsztajn2022Why} conducts experiments on 45 datasets, investigating the differences between tree-based and DNN-based methods. 

The benchmark should cover datasets with varying sizes, including datasets with a large number of samples and features as well as smaller datasets. The diversity of dataset sizes helps evaluate the scalability and efficiency of different models. \cite{McElfreshKVCRGW23when} includes 176 classification datasets and evaluate 19 methods, comprising 8 classical and 11 deep methods. In this study, the pre-trained TabPFN model~\cite{Hollmann2022TabPFN} emerges as the top performer on average, even when limited to randomly sampled training sets of 3000 examples. However, limited trials for hyperparameter tuning and strict time constraints in~\cite{McElfreshKVCRGW23when} may have led to suboptimal evaluations for some deep tabular methods~\cite{tschalzev2025unreflected}.

To ensure robustness and generalization, datasets from multiple domains should be included. Common domains for tabular data include healthcare, biology, finance, education, and physics. Additionally, some datasets are derived from other domains, such as image or speech data, by feature extraction. \cite{Attention-and-contrastive-benchmark} evaluates attention mechanisms and contrastive learning methods across 28 tabular datasets, comparing their performance with traditional deep learning and machine learning approaches.
\cite{Ye2024Closer}, with a particular focus on DNN-based models, uses a benchmark of over 300 tabular datasets spanning a wide range of task types, sizes, and domains. A more diverse collection allows us to assess whether a tabular method can generalize across applications.

\noindent{\bf Semantic-Enriched Datasets.}
In addition, recent research has also focused on evaluating tabular data with rich semantics, such as incorporating meta information related to tasks or integrating attribute names. 
UniTabE~\cite{Yang2024UniTabE} introduces a 7TB dataset containing 13 billion tabular examples for tabular pre-training, covering domains with investing, time series analysis, finance, economics, and with numerical, categorical, text data types. 
CM2~\cite{Ye2024Towards} proposes OpenTabs for cross-table pre-training, which contains an extensive collection of large-scale tables with column name semantics, including approximately 46M tabular samples. TP-BERTa~\cite{Yan2024Making} filters the OpenTabs for datasets with at least 10,000 samples and no more than 32 features, resulting in 101 binary classification datasets and 101 regression datasets with about 10 million samples. GTL~\cite{Wen2024Generative} curates a collection of 384 public tabular datasets from Kaggle, which includes 176 classification and 208 regression tasks spanning a wide range of industrial domains.
TabLib collects a set of 627M tables totaling 69TiB, along with 867B tokens of context~\cite{Eggert2023TabLib}. TabLib was extracted from numerous file formats, including CSV, HTML, SQLite, PDF, Excel, and others, sourced from GitHub and Common Crawl. T4 (The Tremendous Tablib Trawl)~\cite{tabula8b} takes account of the inscrutable statistics and call sheets with personally identifiable information in TabLib and filters TabLib into a collection of 4M tables with 2.1B rows. 

Among these benchmarks and datasets, the semantic-rich ones are primarily used for pre-training LLMs on tabular data, while the others are mainly employed for evaluating standard methods. Besides, some toolboxes implement methods over tabular data, including those for classical methods, as well as those for deep tabular methods~\cite{deeptables,erickson2020autogluon,PyTorch_Tabular,Zaurin_pytorch-widedeep_A_flexible_2023,Liu2024Talent}. To establish a comprehensive tabular benchmark, several factors need to be considered, including the range of datasets and data quality.

\begin{remark}
    Recent studies have proposed alternative perspectives for tabular evaluations, such as focusing on dataset age~\cite{kohli2024towards}, leveraging expert-level feature engineering~\cite{Tschalzev2024DataCentric}, and considering dataset version~\cite{Ye2024Closer}.
    Studies have also highlighted generalization in open word environments in tabular datasets~\cite{Tschalzev2024DataCentric,Rubachev2024TabRed}, where the distributions of training, validation, and test sets differ significantly. More discussions are in~\autoref{sec:extension}. Incorporating diverse, high-quality datasets helps build a reliable benchmark for meaningful model comparisons.
\end{remark}

\subsubsection{Evaluation Protocols}
Given the strong sensitivity of tabular methods to data and the additional randomness in deep methods, robust evaluation is essential. Furthermore, due to the high computational cost of some methods, it is equally important to ensure evaluation efficiency.

\noindent{\bf Model Selection.}
Model selection on the validation set involves both hyperparameter tuning and early stopping, which are essential for reliable evaluation. Due to the large number of hyperparameters in deep methods, automated methods like Optuna~\cite{akiba2019optuna} are commonly used to explore hyperparameters through multiple trials~\cite{GorishniyRKB21Revisiting,gorishniy2023tabr}. During tuning, models are evaluated on the validation split, while models can also be trained with multiple random seeds, providing more reliable evaluations. In each trial and the final training, early stopping~\cite{morgan1989advances} often employed to prevent overfitting, and the epoch with the best validation performance is selected as the final model.

\noindent{\bf Performance Evaluation.}
To assess generalization and prevent overfitting, models are typically evaluated using separate train/val/test splits, with a typical split ratio of 64\%/16\%/20\%. However, such fixed splits may yield inconsistent results. With the rise of deep learning, researchers have proposed more robust evaluation protocols to better reflect model capabilities~\cite{arlot2009survey}. Two main approaches are commonly used: (1) fixing the data split and running multiple trials with different random seeds~\cite{WangSCJLHC21DCNv2,Badirli2020GrowNet,ArikP21TabNet,gorishniy2023tabr,SongS0DX0T19AutoInt,BonetMGI2024HyperFast,GorishniyRKB21Revisiting,Xu2024BiSHop,Chen2023Trompt,Chen2023Excel,Rubachev2022revisiting}; and (2) using cross-validation, where new train/val/test splits are generated in each fold~\cite{Huang2020TabTransformer,Hollmann2022TabPFN,Marton2024GRANDE,NaderSL22DNNR,David2024RealMLP}. A hybrid strategy combining both random seeds and cross-validation is also adopted~\cite{Jiang2024ProtoGate}.

Recent studies show that holdout-based hyperparameter tuning can be unstable and prone to overfitting to the validation set~\cite{CawleyT10,tschalzev2025unreflected}. \cite{tschalzev2025unreflected} found it ineffective on most TabZilla~\cite{McElfreshKVCRGW23when} datasets and instead used 5-fold cross-validation for more robust hyperparameter selection.
As a result, they found the key meta-feature findings reported in~\cite{McElfreshKVCRGW23when} no longer held. This observation was also discussed in~\cite{Ye2024Closer}, which further identified meta-features that have a greater impact on model performance. For small datasets, alternative strategies have been proposed~\cite{Dietterich98,Model_eval_Raschka,Constructing_Hornung}. 
However, this approach significantly reduces the efficiency of hyperparameter search. \cite{nagler2024reshuffling} showed that simply reshuffling data splits can often improve generalization, making holdout selection competitive with cross-validation while remaining more computationally efficient.

%% file: reason.tex
\section{From Classical to Deep Method}\label{sec:reason}
We present possible advantages of deep learning for tabular data, as well as the potential challenges of deep learning when compared with tree-based methods.

\subsection{Advantages of deep representation learning}
Deep tabular models offer several advantages beyond performance when compared with classical methods. 

\noindent{\bf Ability to Model Complex Feature Interactions.}
DNNs are particularly adept at capturing high-order, non-linear interactions between features, which may be challenging for traditional models like linear regression or decision trees~\cite{WangFFW17DCN,WangSCJLHC21DCNv2}. By learning a hierarchical representation of features, DNNs allow low-level feature interactions to be captured in the initial layers, while higher-order interactions are identified in deeper layers. This ability to automatically learn complex relationships makes DNNs highly effective in capturing intricate dependencies within tabular data.

\noindent{\bf End-to-End Learning.}
Unlike traditional machine learning methods, which often involve separate steps for feature engineering, preprocessing, and model tuning, DNNs can process raw features and automatically extract useful representations without complex manual transformations. This end-to-end learning approach reduces human bias and simplifies the workflow, making the process more efficient. DNNs are trained through gradient optimization, enabling a unified, streamlined solution for complex tasks~\cite{GorishniyRKB21Revisiting,JiangYW00Z24Tabular}. Additionally, deep models support multi-task learning, allowing related tasks to benefit from shared representations, enhancing both performance and efficiency~\cite{Feng0Z18_mGBDT,Somepalli2021SAINT,Wu2024SwitchTab}.

\noindent{\bf Integration with Other Modalities.}
Deep tabular methods excel in multi-modal pipelines, where tabular data is integrated with other modalities, such as images, audio, or text. In AI4science applications, for instance, tabular data might be combined with image data~\cite{hager2023best,JiangYW00Z24Tabular} (\eg, in medical imaging applications) or time-series data~\cite{padhi2021tabular,di2023explainable} (\eg, in forecasting tasks). DNNs are well-suited to model interactions between heterogeneous data types, improving the overall performance. By jointly learning from multiple data sources, DNNs enhance their ability to make more accurate and comprehensive predictions across domains.

\noindent{\bf Flexibility with Dynamic Environments.}
DNN-based methods benefit from the flexibility of gradient-based optimization, which allows efficient and iterative training. This flexibility makes DNNs adaptable to changing objectives without significant modifications, unlike tree-based models that often require specialized methods for different tasks~\cite{Borisov2024Deep}. Moreover, DNNs excel in dynamic environments, such as real-time predictions, financial analysis, and decision-making systems, where feature relationships may shift. This adaptability makes them suitable for online learning or incremental training, where new data is continuously integrated without retraining from scratch~\cite{van2022three,zhou2024class}.

\noindent{\bf Long-Term Knowledge Transfer and Learning.}
DNNs are capable of long-term learning and knowledge transfer, which allows them to retain valuable knowledge gained from training on diverse tasks~\cite{yosinski2014transferable}. Once trained on a broad set of tasks, DNNs can transfer this knowledge to related domains, reducing the need for complete retraining~\cite{dar2020transfer}. This is especially advantageous in fields like AI4science, where a model trained on one type of scientific data can be adapted to other related domains, saving both time and computational resources. This ability to transfer knowledge across tasks is a key advantage of deep learning, enabling more efficient use of data and model capabilities over time.

\subsection{Debates between Tree-Based Methods and DNNs}
Although deep tabular methods have shown great potential in learning semantic representations and constructing nonlinear predictors, their initial performance often struggles to surpass that of classical tree-based ensemble methods, such as Gradient Boosted Decision Trees (GBDT). Many studies still treat GBDT approaches as strong baselines~\cite{Grinsztajn2022Why,McElfreshKVCRGW23when}, and in some cases, the advantages of deep tabular methods diminish as the number of evaluation datasets increases.

Several reasons contribute to why tree-based methods retain their advantages over DNNs in many tabular tasks:

\noindent{\bf Better Handling of High-Frequency Data.}
Tree-based methods, particularly GBDT models, are highly efficient at handling high-frequency data or dense datasets with many small variations~\cite{EgeInductiveBias}. These models build decision trees by recursively splitting the data at the most informative feature points, capturing both local and global patterns efficiently. DNNs, on the other hand, may not capture fine-grained patterns as effectively without extensive regularization or tuning~\cite{cao2019towards,basri2020frequency}. To address this limitation,~\cite{EgeInductiveBias} introduced frequency reduction as an inductive bias through the addition of scaling layers, while ~\cite{Gorishniy2022On} demonstrated that periodic activation functions can significantly enhance neural networks' ability to learn high-frequency functions.

\noindent{\bf Natural Handling of Mixed Data Types.}
Tabular data often includes a combination of numerical, categorical, and ordinal features~\cite{Borisov2024Deep,Ye2024Closer,NEURIPS2023_ac01e21b}. Tree-based models are particularly strong when working with mixed data types, as they can handle categorical features directly without requiring one-hot encoding or embeddings. This ability to work with raw categorical data simplifies the preprocessing pipeline significantly. DNNs, however, generally require encoding techniques (\eg, one-hot encoding or learned embeddings) for categorical features, adding complexity and potentially leading to suboptimal performance~\cite{Huang2020TabTransformer}.

\noindent{\bf Lower Computational Requirements for Training and Inference.}
For certain tasks, tree-based models tend to be more computationally efficient than DNNs~\cite{GorishniyRKB21Revisiting}. GBDTs and other decision tree-based models can train relatively quickly and are less computationally intensive than deep neural networks~\cite{yan2024team,McElfreshKVCRGW23when}. This is especially true when the dataset is not massive or when the model needs to be trained and deployed rapidly. DNNs, on the other hand, often require significant computational resources (\eg, GPUs, longer training times) to achieve comparable performance, making them less ideal in resource-constrained environments~\cite{PangTZZ22,Muller2023MotherNet}.

\noindent{\bf Robustness to Noisy and Missing Data.}
Tree-based models are generally more robust to noisy data and missing values. When training a decision tree, missing values can be handled through optimal splitting that accommodates absent data, and trees can effectively deal with noisy or inconsistent data points~\cite{Grinsztajn2022Why}. DNNs, in contrast, are more sensitive to noise and often require careful preprocessing or specific techniques (\eg, data imputation or noise filtering) to avoid performance degradation with noisy or missing data~\cite{Chen2023Excel,Hollmann2022TabPFN}.

\noindent{\bf Interpretability and Transparency.}
Tree-based models offer a significant advantage in terms of interpretability~\cite{PopovMB20Neural,Chang0G22NODEGAM,ArikP21TabNet}. The decision-making process of models like GBDT can be easily visualized in the form of decision paths, and feature importance can be directly extracted~\cite{chen2016xgboost,Prokhorenkova2018Catboost,ke2017lightgbm}. This transparency makes tree-based models appealing in domains where model explainability is crucial, such as in finance, healthcare, and regulated industries. Although interpretability techniques like LIME~\cite{Ribeiro0G16_LIME} and SHAP~\cite{NIPS2017_SHAP} exist for DNNs, tree-based models still tend to be more intuitive and easier to explain, especially in complex decision-making environments. Recent works~\cite{zhou2019deep,PopovMB20Neural,Badirli2020GrowNet,Marton2024GRANDE} have sought to bridge this gap by enhancing neural network interpretability through emulation of tree-based model behaviors.

\noindent{\bf Handling Outliers and Skewed Data.}
Tree-based methods are often better at handling outliers and skewed distributions in the data. When a feature exhibits extreme values or skewed distributions, decision trees are inherently less sensitive to such anomalies because they create splits based on feature ranges that naturally isolate outliers. This characteristic can make them more robust than DNNs, which may require specialized loss functions or techniques (\eg, robust scaling or outlier removal) to handle such data points~\cite{Tschalzev2024DataCentric,Rubachev2024TabRed}.

%% file: taxonomy2.tex
\begin{table*}[t]
	\caption{ The taxonomy of representation learning for tabular data. The shade color in the last column denotes the subcategory, which is consistent with~\autoref{figure:taxo}.}
	\label{table:taxonomy}
	\centering
	\resizebox{0.88\textwidth}{!}{
		\begin{tabular}{@{}c|c|l|c|p{8cm}<{\centering}@{}}
			\toprule
			\multicolumn{3}{c|}{Algorithm Category} & Reference \\ 
			\midrule
			\midrule
                \multirow{8}{*}[-4.5ex]{ Specialized Methods } &
			\multirow{4}{*}[-2.5ex]{\begin{tabular}[c]{@{}c@{}}\textcolor{red}{$\S~$}\ref{sec:feature}\\ Feature-aspect Methods \end{tabular}} &
			\begin{tabular}[c]{@{}c@{}} {Feature Encoding}\end{tabular}
			& 
			\cellcolor{DeepRed!20}
			\cite{GorishniyRKB21Revisiting,Gorishniy2022On,Zhou2023TabToken} \\ 
			\cmidrule(l){3-4} 
			\multicolumn{1}{c|}{}  & &
			\begin{tabular}[c]{@{}c@{}} {Feature Selection}\end{tabular}
			& \cellcolor{DeepRed!20}
			\cite{Badirli2020GrowNet,PopovMB20Neural,ArikP21TabNet,Chang0G22NODEGAM,Marton2024GRANDE} \\ 
                \cmidrule(l){3-4} 
			\multicolumn{1}{c|}{}  & &
			\begin{tabular}[c]{@{}c@{}} {Feature Projection}\end{tabular}
			& \cellcolor{DeepRed!20}
			\cite{KlambauerUMH17Self,GorishniyRKB21Revisiting,David2024RealMLP,Xu2024BiSHop} \\ 
                \cmidrule(l){3-4} 
			\multicolumn{1}{c|}{}  & &
			\begin{tabular}[c]{@{}c@{}} {Feature Interaction }\end{tabular}
			& \cellcolor{DeepRed!20}
			\cite{WangSCJLHC21DCNv2,SongS0DX0T19AutoInt,Huang2020TabTransformer,ChenLWCW22DAN,Chen2023Excel,Wu2024SwitchTab,cheng2024arithmetic} \\ 
\cmidrule(l){2-4} 
			& \multirow{3}{*}{\begin{tabular}[c]{@{}c@{}}\textcolor{red}{$\S~$}\ref{sec:sample}\\ Sample-aspect Methods\end{tabular}} &
			\begin{tabular}[c]{@{}c@{}} {Sample Interaction}\end{tabular}
			& \cellcolor{Yellow!20}
			\cite{Somepalli2021SAINT,KossenBLGRG21Attention,Bernhard2022Hopular,Chen2023Trompt,PTARL} \\ 
			\cmidrule(l){3-4} 
			\multicolumn{1}{c|}{}  & &
			\begin{tabular}[c]{@{}c@{}} {Neighbor Retrieval}\end{tabular}
			& \cellcolor{Yellow!20} \cite{Kim2019Attentive,NaderSL22DNNR,gorishniy2023tabr,Ye2025ModernNCA} \\
\cmidrule(l){2-4} 
	& \multirow{3}{*}{\begin{tabular}[c]{@{}c@{}}\textcolor{red}{$\S~$}\ref{sec:objective}\\ Objective-aspect Methods\end{tabular}} &
			\begin{tabular}[c]{@{}c@{}} {Training Objective}\end{tabular}
			& \cellcolor{DeepBlue!20}
			\cite{PTARL} \\ 
			\cmidrule(l){3-4} 
			\multicolumn{1}{c|}{}  & &
			\begin{tabular}[c]{@{}c@{}} {Training Regularization}\end{tabular}
			& \cellcolor{DeepBlue!20} \cite{shavitt2018regularization,Kadra2021Well,jeffares2023tangos} \\ 

			\midrule
			\multicolumn{2}{c|}{\multirow{5}{*}[-1.5ex]{\begin{tabular}[c]{@{}c@{}}\textcolor{red}{$\S~$}\ref{sec:transfer}\\ Transferable Methods\end{tabular}}} &
			\begin{tabular}[c]{@{}c@{}} {Homogeneous}\end{tabular}
			& \cellcolor{dreg!20}
			\cite{Huang2020TabTransformer,YoonZJS20VIME,Somepalli2021SAINT,Verma2021DACL,UcarHE21SubTab,Lee2022Self,levin2022transfer,Majmundar2022MET,BahriJTM22Scarf,Hajiramezanali2022STab,Chen2023ReConTab,DDu2023DORA,Sui2024Self} \\ 
			\cmidrule(l){3-4} 
			\multicolumn{2}{c|}{}  & 
			\begin{tabular}[c]{@{}c@{}} {Heterogeneous}\end{tabular} &
			\cellcolor{dreg!20} \cite{Iwata2020Meta,Liu2022Distribution,levin2022transfer,Onishi2023TabRet,shen2023cross,Zhou2023TabToken,zhu2023xtab,Zhang2023Meta} \\ 
			\cmidrule(l){3-4} 
			\multicolumn{2}{c|}{}  &
			\begin{tabular}[c]{@{}c@{}} {Language Model}\end{tabular} &
			\cellcolor{dreg!20} \cite{Wang2022TransTab,Liu2022PTab,Yang2024UniTabE,Ye2024Towards,Yan2024Making,Kim2024CARTE,Cheng2023Binding,Hollmann2023CAAFE,Han2024FeatLLM,Zhang2023TapTap,Dinh2022LIFT,Hegselmann2022TabLLM,Wang2023UniPredict} \\
            \cmidrule(l){3-4} 
			\multicolumn{2}{c|}{}  &
			\begin{tabular}[c]{@{}c@{}} {Vision Model}\end{tabular} &
			\cellcolor{dreg!20} \cite{Sharma2019DeepInsight,Bazgir2020REFINED,buturovic2020novel,Zhu2021IGTD,Lee2023TablEye,Vanesa2024LM-IGTD,Sun2019SuperTML,Mamdouh2025Tab2Visual} \\
                \midrule
            \multicolumn{2}{c|}{\multirow{4}{*}{\begin{tabular}[c]{@{}c@{}}\textcolor{red}{$\S~$}\ref{sec:general}\\ General Mehtods\end{tabular}}} &
			\begin{tabular}[c]{@{}c@{}} {Raw-Feature-based}\end{tabular}
			& \cellcolor{mr!20}
			\cite{Ye2023TabPTM,BonetMGI2024HyperFast,Muller2023MotherNet} \\ 
			\cmidrule(l){3-4} 
			\multicolumn{2}{c|}{}  &
			\begin{tabular}[c]{@{}c@{}} {TabPFN Variants}\end{tabular} &
			\cellcolor{mr!20} \cite{Hollmann2022TabPFN,hollmann2025tabpfn} \\ 
                \cmidrule(l){3-4} 
			\multicolumn{2}{c|}{}  &
			\begin{tabular}[c]{@{}c@{}} {Semantics-based}\end{tabular} &
			\cellcolor{mr!20} \cite{tabula8b,WenZZXB24From,Wen2025ICL,Wang2024MediTab} \\
			\bottomrule
		\end{tabular}
	}
	\vspace{-4mm}
\end{table*}

\begin{figure*}[t]
    \includegraphics[width=0.99\textwidth]{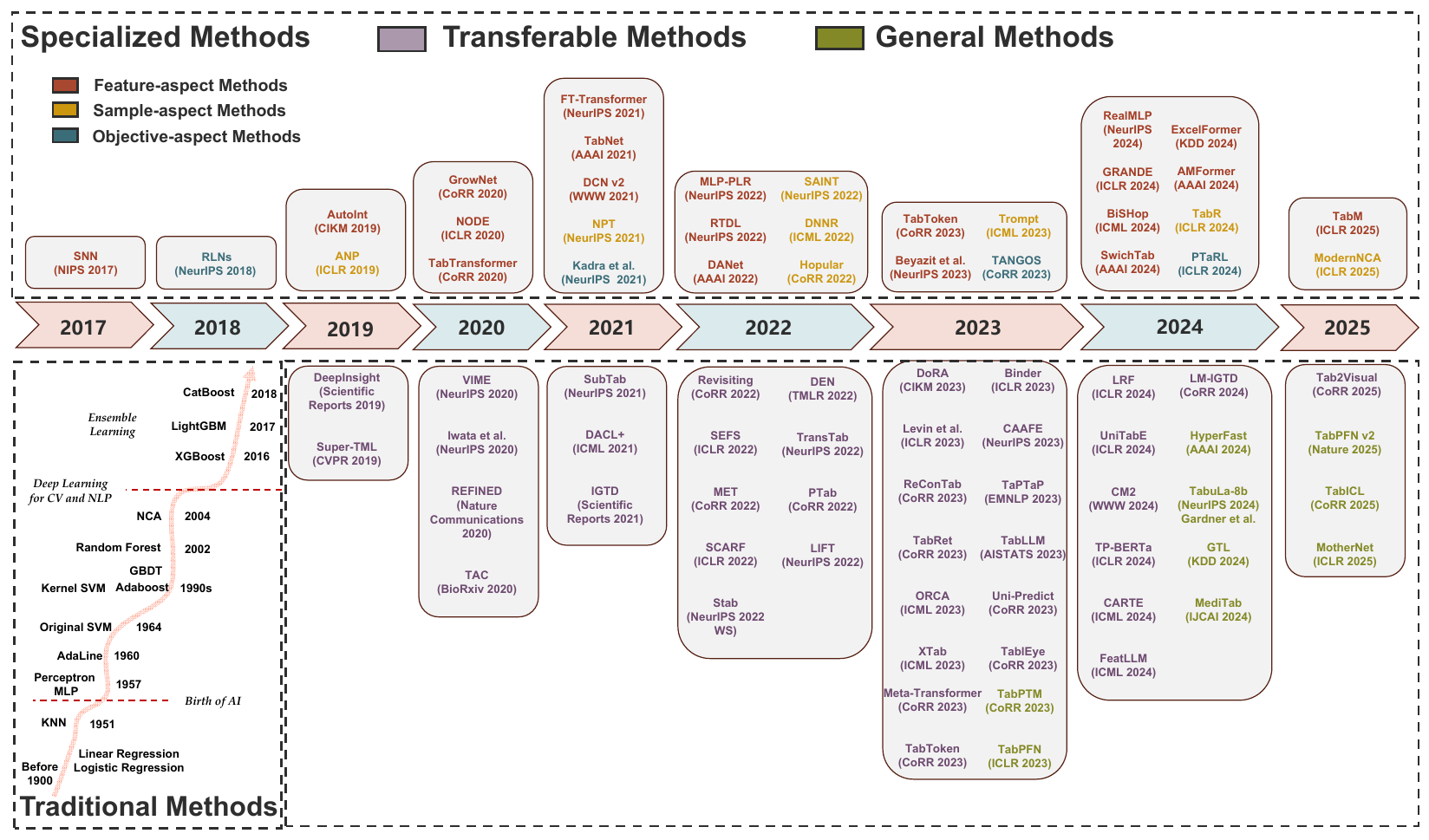}
    \caption{ The roadmap of deep representation learning tabular methods. We organize representative methods chronologically to show the concentration at
different stages. Different colors of these methods denote the sub-categories.} 
    \label{figure:taxo} 
\end{figure*}

%% file: taxonomy.tex
\section{Taxonomy of Specialized Methods}\label{sec:Taxonomy}
Similar to the evolution of deep learning, which progresses from specialized learning to transfer learning and ultimately to foundation models~\cite{bommasani2021opportunities}, we categorize deep tabular methods into three groups, as shown in~\autoref{figure:intro2}: \textit{specialized methods}, \textit{transferable methods}, and \textit{general methods}. This classification reflects both the evolutionary development of deep learning techniques and the increasing generalization capabilities of these models. 

Specialized methods, being the earliest developed and most widely used category, will be our starting point for discussion. Tabular data consists of features (columns), samples (rows), and objectives (labels), which together define the structure and the task objectives. We emphasize detailed strategies for obtaining high-quality representations at both feature- and sample-level for the target task. Specifically, given the input data, according to the general learning objective in~\autoref{eq:objective}, we consider how to transform the tabular input $\x_i$ (feature aspect), how to construct relationships between samples (sample aspect), how to design the objective $\ell(\cdot)$ and regularize $\Omega(\cdot)$ (objective aspect). In particular, 
\begin{itemize}[noitemsep,topsep=0pt,leftmargin=*]
    \item {\bf Feature Aspect.} We focus on how to transform the raw tabular input (in various forms) into intermediate representations. We consider two types of features: numerical and categorical. By explicitly modeling the relationships between the two features (\eg, feature importance and interactions), we are able to enhance the model’s understanding of the input space.

    \item {\bf Sample Aspect.} In addition to features, we explore how to retrieve and utilize neighboring samples to capture inter-sample dependencies, thereby improving predictions. In order to improve the model's ability to make predictions, we explore the relationships between a target sample and its ``extracted neighbors.''
    
    \item {\bf Objective Aspect.} We examine how to modify the loss function and overall objective to introduce inductive biases. By directly guiding the learning process with the target variables, we incorporate prior knowledge or task-specific preferences into the model, thereby improving its generalizability and interpretability.    
\end{itemize}

In specialized methods, we focus solely on learning from pure data, excluding feature semantics considered in transferable methods (in~\autoref{sec:transfer}), as they leverage the capabilities of language models. Since specialized methods encompass a wide range of approaches, and feature-aspect methods are the most extensive part of them, we will first introduce sample-aspect methods and objective-aspect methods in the following subsections. In~\autoref{sec:feature}, we will provide a detailed introduction to feature-aspect methods.

\subsection{Sample-aspect Specialized Methods}\label{sec:sample}
Sample interaction methods take a retrieval-based approach, focusing on relationships between individual samples rather than features. In a tabular dataset, each sample $\x_i$ represents a row with $d$ features, and the goal is to leverage relationships between a target sample and its ``extracted neighbors'' to improve predictions.

The general form for the sample interaction methods can be expressed as:
\begin{equation}
    \hat{y}_i = f\left(\mathcal{R}(\x_i, \mathcal{D}; \Phi)\right),
\end{equation}
where $\mathcal{D}$ is the set of all samples (training data) available for retrieval or learning. $\mathcal{R}(\cdot)$ is the sample interaction module, which retrieves or aggregates information from relevant samples in $\mathcal{S}$ for the target sample $\x_i$. $\Phi$ represents the learnable parameters of $\mathcal{R}$. $f(\cdot)$ is the prediction head that maps the aggregated information to the final output $\hat{y}_i$.

Sample aspect approaches can be broadly categorized into two main strategies. The first approach introduces the modeling of sample relationships $\mathcal{R}$ during representation training, allowing the model to learn better representations by capturing inter-sample dependencies. The second approach is retrieval-based models, which directly predict outcomes by learning how to retrieve and utilize neighbors' relationships $\mathcal{R}$ when testing.

\noindent{\bf Sample Interaction.}
These methods assist in representation learning by allowing the model to capture relationships between samples, which in turn helps generate a more robust representation during training. During testing, the model becomes more sensitive to each sample without interaction.

SAINT~\cite{Somepalli2021SAINT} introduces inter-sample attention beyond inter-attribute attention, which improves row classification by relating each row to others in the table. NPT~\cite{KossenBLGRG21Attention} extends this via non-parametric Transformers, whereas Hopular~\cite{Bernhard2022Hopular} employs Hopfield networks, sharing conceptual alignment with SAINT~\cite{Somepalli2021SAINT}. Unlike nearest-neighbor classification, the distance metric is learned end-to-end. Trompt~\cite{Chen2023Trompt} posits that the feature importance in tabular data is sample-dependent. During feature extraction, it treats the information between samples as prompts. PTaRL~\cite{PTARL} identifies two issues in the representation of tabular data samples: entanglement and localization. It addresses these by modeling global sample relationships through prototype generation and representation projection, helping the model produce clear and consistent decisions.

\noindent{\bf Neighbor Retrieval.}
These methods construct high-quality contexts to aid prediction by retrieving valuable neighbors and designing efficient ways to utilize them based on the relationships between samples. The training data is used to assist during testing.

DNNR~\cite{NaderSL22DNNR} argues that a key advantage of neighbor-based methods is the model's transparency, meaning that the model's decisions can be explained by inspecting its components. It enhances predictive performance by incorporating local gradient estimation and Taylor series approximation into the KNN framework. TabR~\cite{gorishniy2023tabr} proposes that, compared to purely parametric (\eg, retrieval-free) models, retrieval-based models can achieve superior performance while also exhibiting several practically important properties, such as the ability for incremental learning and enhanced robustness. It encodes all candidate samples and then employs an attention-like mechanism to retrieve the samples that aid in making predictions, as explored in \cite{Kim2019Attentive}. ModernNCA~\cite{Ye2025ModernNCA} revitalizes the classic tabular prediction method, Neighbourhood Component Analysis (NCA)~\cite{goldberger2004neighbourhood}, by designing and incorporating deep learning architectures and strategies. The resulting method efficiently leverages neighboring samples for prediction.

\begin{remark}
   The neighborhood-based approach closely resembles the current in-context learning~\cite{BrownMRSKDNSSAA20} mechanism. In particular, the in-context learning used in general models like TabPFN~\cite{Hollmann2022TabPFN,hollmann2025tabpfn} can aslo be considered a form of the neighborhood method. This concept of neighborhood not only helps in standard tasks, but also enhances transferable and general methods. For example, LoCalPFN~\cite{Thomas2024LocalPFN} highlights that employing local linear regression can lead to more expressive decision boundaries, while utilizing local context allows performance to scale with the size of the training dataset.
\end{remark}

\subsection{Objective-aspect Specialized Methods}\label{sec:objective}
The general objective learning $f$ follows the structural risk minimization as in~\autoref{eq:objective}, where $\ell$ is the loss function to set the training objective between the prediction and the ground truth label. $\Omega(\cdot)$ is the regularization on the model, which directs the objective or restricts the complexity of $f$.

In traditional machine learning, models often rely on explicit regularization techniques on $\Omega$ to ensure good generalization. Methods such as decision trees, support vector machines, and linear models typically incorporate regularization terms directly into the loss function to control model complexity and prevent overfitting. For example, in linear regression, regularization methods like L1 (Lasso)~\cite{tibshirani1996regression}, L2 (Ridge)~\cite{hoerl1970ridge}, or Elastic-Nets~\cite{elastic-net} penalize large coefficients, effectively controlling the complexity of the model and helping to maintain a balance between bias and variance.

Objective-aspect methods in deep learning are an extension of these traditional regularization techniques, where inductive bias is introduced by adjusting the loss function $\ell$ or adding regularizers $\Omega$. In the training progress, the goal is to leverage regularization on the model to improve predictions.

\begin{remark}
Pre-train methods such as homogeneous transferable tabular methods in~\autoref{sec:transfer} also change the loss function $\ell$ or the regularization $\Omega$ to help pre-training. We will discuss these methods later.
\end{remark}

Objective-aspect approaches can be broadly categorized into two main strategies. 
The first approach involves training objectives, which enhance the model with a specialized ability. 
The second approach introduces a regularizer, allowing the model to learn strong generalized representations.

\noindent{\bf Training Objective.} 
For training objectives, PTaRL~\cite{PTARL} constructs prototype-based projection space and learns the disentangled representation around global prototypes. PTaRL uses a diversification constraint for representation calibration and introduces a matrix orthogonalization constraint to ensure the independence of global prototypes.

\noindent{\bf Training Regularization.}
For training regularization, RLNs~\cite{shavitt2018regularization} overcome the challenge of an intractable number of hyperparameters during training by introducing an efficient tuning scheme, which minimizes a new ``Counterfactual Loss.'' In RLNs, the regularization coefficients are optimized together with learning the network weight parameters. RLNs produce extremely sparse networks, thus providing more interpretable models and revealing the importance that the network assigns to different inputs. 
~\cite{Kadra2021Well} introduces “cocktails,” dataset-specific combinations of 13 regularization techniques, showing that even simple neural networks can outperform tree-based architectures when optimized with these methods. TANGOS~\cite{jeffares2023tangos} introduces a regularization-based improvement. It regularizes neuron attributions to encourage neurons to specialize and become orthogonal to one another.

\section{Feature-aspect Specialized Methods}\label{sec:feature}
\begin{figure}[t]
    \includegraphics[width=0.49\textwidth]{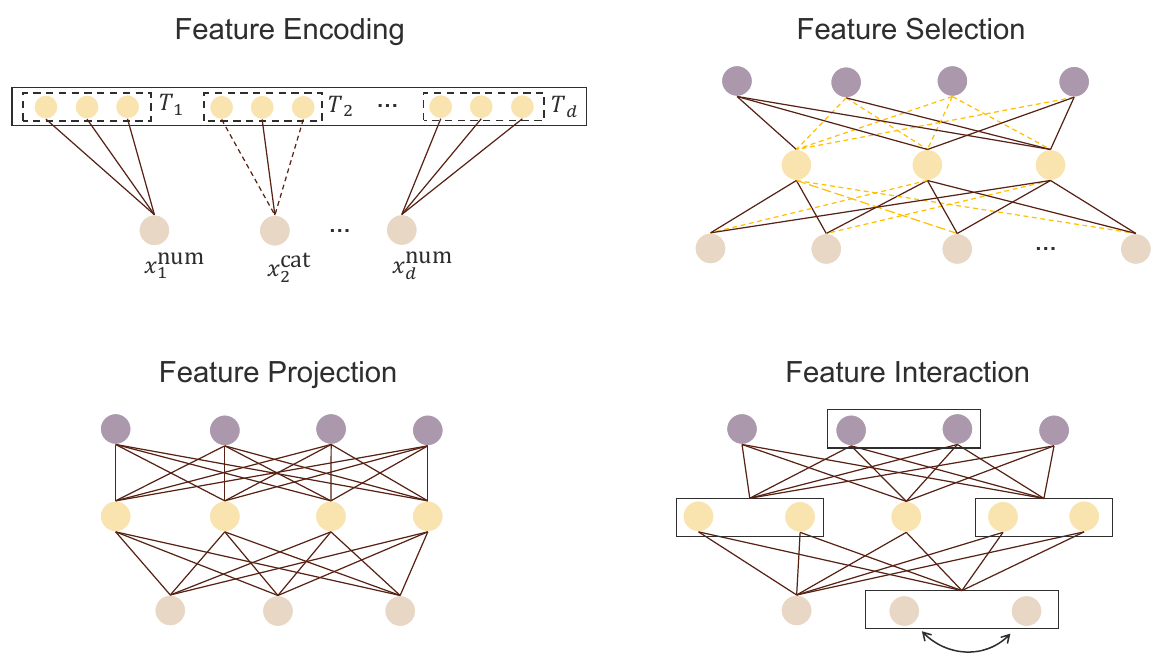}
    \caption{ Illustration of feature-aspect methods, including feature encoding, feature selection, feature projection and feature interaction. } 
    \label{figure:feature} 
    \vspace{-5mm}
\end{figure}

Tabular data is characterized by a diverse set of features, including both categorical and numerical variables. The complexity of tabular data arises from the variety of feature types, their interrelationships, and the high dimensionality often present. Traditional methods often rely on manual feature engineering, using techniques such as encoding categorical variables and selecting relevant features to improve model performance and reduce overfitting.

As deep learning has evolved, these traditional techniques have been integrated and expanded upon. Deep tabular models are capable of automatically learning complex feature representations, reducing the need for explicit feature engineering. Feature-aspect methods, such as feature encoding, selection, projection, and interaction, are essential for transforming raw tabular inputs into more informative intermediate forms. These methods help improve a model's ability to capture intricate relationships between features, thereby enhancing its generalization capabilities.

\subsection{Feature Encoding}
Various encoding strategies have been explored for both categorical and numerical features in tabular data. Additionally, with the advancement of the attention mechanism, feature tokenization, similar to word embeddings in natural language processing, transforms all features into embeddings.

\noindent{\bf Categorical Encoding.} Categorical variables represent types of data which may be divided into groups. Examples of categorical variables are race, sex, age group, and educational level~\cite{hancock2020survey}. The categorical features are usually transformed in an index (integer). The two most popular techniques are an Ordinal Encoding and a One-Hot Encoding. 

Ordinal Encoding assigns each unique category a distinct integer value. This approach is useful when the categorical variable has an inherent order, such as ``low,'' ``medium,'' and ``high.'' The main advantage of Ordinal Encoding is its simplicity and efficiency, as it transforms the categorical variable into a single numeric column. However, it assumes that there is an ordinal relationship between the categories, which may not always be the case. For instance, if the categorical variable represents "color" with categories such as ``red,'' ``blue,'' and ``green,'' applying Ordinal Encoding would introduce an artificial order that does not reflect any meaningful ranking.

On the other hand, One-Hot Encoding creates a new binary column for each unique category in the original categorical variable. For example, for a variable "color" with three categories (red, blue, and green), One-Hot Encoding would generate three binary columns: ``is\_red,'' ``is\_blue,'' and ``is\_green,'' encoding red as (1, 0, 0), blue as (0, 1, 0) and green as (0, 0, 1). Each column indicates the presence or absence of that particular category. One-Hot Encoding is useful for nominal categorical variables, where no order exists between the categories. While One-Hot Encoding avoids the assumption of ordinal relationships, it can lead to a high-dimensional feature space if the categorical variable has many unique values, which may result in increased computational costs and potential issues with overfitting.

In some cases, more advanced encoding techniques are used to address the limitations of these basic approaches. For example, Target Encoding assigns each category a value based on the mean of the target variable for that category. This method can be useful when there is a strong relationship between the categorical features and the target. In Leave-one-out embedding, every category is replaced
with the mean of the target variable of that category, which excludes the current row to avoid overfitting.

\noindent{\bf Numerical Encoding.} 
For encoding, MLP-PLR~\cite{Gorishniy2022On} introduces two numerical encoding methods: Piecewise Linear Encoding (PLE) and Periodic Activation Functions. 
These encoding methods can be integrated with other differentiable layers (\eg, Linear, ReLU) to enhance performance. PLE produces alternative initial representations for the original scalar values and is based on feature binning. Periodic Activation Functions take into account the fact the embedding framework where all features are computed independently of each other forbids mixing features during the embedding process and train the pre-activation coefficients instead of keeping them fixed. 
~\cite{EgeInductiveBias} utilizes tools from spectral analysis, showing that functions described by tabular datasets often have high irregularity, and can be smoothed by transformations such as scaling and ranking to improve performance. They propose ``frequency reduction'' as an inductive bias during training. 

\noindent{\bf Feature Tokenization.} 
Feature tokenizer performs a similar role to the feature extractor in traditional models. It transforms the input features to embeddings~\cite{SongS0DX0T19AutoInt,GorishniyRKB21Revisiting}. Since the feature representations of features are very sparse and high-dimensional, a common way is to represent them into low-dimensional spaces (\eg, word embeddings). 

The general form for feature tokenization can be expressed as:
\begin{equation}
    \mT_{i,j}=\vb_j+\mathcal{T}(x_{i,j}; \Psi)\in\mathbb{R}^t,
\end{equation}
where $\mathcal{T}(\cdot)$ is the feature tokenizer module, which transforms the input feature vector $\x_i \in \mathbb{R}^d$ to a token embedding $\mT_{i,j} \in \mathbb{R}^{t}$. $t$ is the dimension of token embedding. $\vb_j$ is the $j$-th feature bias. $\mathcal{T}$ can be implemented with different forms. $\Psi$ represents the learnable parameters of $\mathcal{T}$.

In AutoInt~\cite{SongS0DX0T19AutoInt}, both the categorical and numerical features are embedded into low-dimensional spaces, which reduces the dimension of the input features and meanwhile allows different types of features to interact with each other. The embeddings of categorical features are computed by multiplying the embedding matrix with the multi-hot vector, while a corresponding embedding vector represents numerical features. TabTransformer~\cite{Huang2020TabTransformer} embed each categorical feature into a parametric embedding of dimension $t$ using Column embedding. An embedding vector is assigned to each feature, and a set of embeddings is constructed for all categorical features. Unlike TabTransformer, SAINT~\cite{Somepalli2021SAINT} proposes projecting numerical features into a $t$-dimensional space before passing their embedding through the transformer encoder. FT-Transformer~\cite{GorishniyRKB21Revisiting} adapts the Transformer architecture for tabular data, where all features are transformed to embeddings and applies a stack of Transformer layers to the embeddings. Specifically, the numerical tokenizer is implemented as the element-wise multiplication $\mT_i^{\textit{\rm num}}=\vb_i^{\textit{\rm num}}+x_i^{\textit{\rm num}}\cdot \mW_i^{\textit{\rm num}}$, and the categorical tokenizer is implemented as the lookup table $\mT_i^{\textit{\rm cat}} =\vb_i^{\textit{\rm cat}} + \e_i^T \mW_i^{\textit{\rm cat}}$, where $\e_i^T$ is a one-hot vector for the corresponding categorical feature. Other transformer-based methods, like \cite{Chen2023Excel,Onishi2023TabRet,zhu2023xtab,cheng2024arithmetic}, use the same feature tokenizer as FT-Transformer.

\subsection{Feature Selection}
The high dimensionality of tabular data often causes overfitting, where the model focuses on irrelevant features and neglects the important ones. Feature selection reduces the number of features, retaining only the most valuable information. This helps prevent overfitting, improves generalization, and reduces computational complexity.

Traditional tree-based models facilitate automatic feature selection by evaluating the impact of each feature on the target during the construction process. Decision trees utilize metrics such as information gain or the Gini index for feature selection, while ensemble methods like random forests determine feature importance by assessing each feature's contribution~\cite{quinlan2014c4,breiman2001random,zhou2004nec4}. Recently, modern deep learning methods for tabular data often mimic trees' structures for feature selection.

GrowNet~\cite{Badirli2020GrowNet} and NODE~\cite{PopovMB20Neural} primarily mimic ensemble techniques. Inspired by GBDT, GrowNet designs a framework for building DNNs with multiple weak learners, where each learner's input consists of the original features plus the penultimate layer output from the previous learner. NODE uses a differentiable Oblivious Decision Tree as the base model, applying Bagging within each layer and Stacking across layers in a multi-layered structure. To make GAM~\cite{hastie1986generalized} scalable and effective, NODE-GAM~\cite{Chang0G22NODEGAM} modifies NODE to be a GAM, allowing GAM to learn quick, non-linear jumps that better match patterns in real data. 

TabNet~\cite{ArikP21TabNet} and GRANDE~\cite{Marton2024GRANDE} focus more on how tree models handle features. TabNet not only retains the representation learning capabilities of DNNs through self-supervised learning, but also incorporates the interpretability of tree models and the benefits of sparse feature selection, with a model structure designed for both feature selection and computation. GRANDE argues that the hard splits used by tree models are a key advantage over deep models, and thus proposes a method for learning hard, axis-aligned tree ensembles using gradient descent. GRANDE combines the beneficial inductive bias of axis-aligned splits with the flexibility provided by gradient descent optimization.

\subsection{Feature Projection}
Feature projection methods aim to project the raw data into a middle form, enhancing the representation ability for later architectures. Feature projection methods can be broadly categorized into two main approaches: MLP variants and special designed architectures. These approaches aim to enhance the model’s ability to represent complex features for underlying feature structures.

\noindent{\bf MLP Variants.}
For model architecture, RTDL~\cite{GorishniyRKB21Revisiting} investigates both ResNet-like and Transformer-based architectures tailored for tabular data, proposing simple yet effective adaptations of these widely-used deep models. In particular, the MLP architecture is constructed by stacking multiple blocks consisting of Linear layers, ReLU activations, and Dropout, which transform the raw tabular features into a fixed-dimensional hidden representation. A final linear layer is then used as the classification head. The paper highlights an important insight: with proper hyperparameter tuning, even simple architectures like MLP and ResNet can achieve competitive performance on tabular benchmarks. 

Another contemporaneous work~\cite{Kadra2021Well} enhances the MLP architecture by equipping it with a comprehensive suite of modern regularization techniques. Instead of introducing architectural innovations, this study focuses on systematically exploring combinations of 13 different regularization methods to identify an effective ``regularization cocktail'' for plain MLPs. The results demonstrate two key findings: (i) a well-regularized vanilla MLP can significantly outperform many recent, specialized neural architectures designed for tabular data; and (ii) such MLPs can even surpass strong traditional machine learning models like XGBoost across a range of benchmarks. For a more comprehensive strategy, RealMLP~\cite{David2024RealMLP} explores multiple aspects including preprocessing, hyperparameters, architecture, regularization, and initialization.

\noindent{\bf Special Designed Architectures.}
For units, motivated by the observation that normalization techniques are prone to disturbances during training, SNN~\cite{KlambauerUMH17Self} proposes the Scaled Exponential Linear Unit (SELU) to improve deep models for tabular data. NAMs~\cite{agarwal2021neural} uses exp-centered (ExU) hidden units to improve the learnability for fitting jumpy functions. 
BiSHop~\cite{Xu2024BiSHop} uses a dual-component approach, sequentially processing data both column-wise and row-wise through two interconnected directional learning modules. They use layers of generalized sparse modern Hopfield layers, a sparse extension of the modern Hopfield model with learnable sparsity.

\subsection{Feature Interaction}
Feature interaction methods aim to model relationships among features to enhance the representation power of deep learning models on tabular data. In tabular datasets, each sample $\x_i \in \mathbb{R}^d$ is described by $d$ features, and the goal is to transform these raw features into enriched representations that improve predictive performance.

The general form for feature interaction methods can be expressed as:
\begin{equation}
    \hat{y}_i = f\left(\mathcal{H}(\x_i; \Theta)\right),
\end{equation}
where $\x_i \in \mathbb{R}^d$ is the input feature vector for a single instance, $\mathcal{H}(\cdot)$ is the feature interaction module, which transforms the input $\x$ by capturing feature dependencies or generating higher-order feature interactions. $\Theta$ represents the learnable parameters of $\mathcal{H}$. $f(\cdot)$ is the prediction head that maps the transformed representation to the final output $\hat{y}$.

Feature interaction methods can be broadly categorized into two main approaches: the design of automatic feature interaction modules and the mining of implicit feature relationships. These approaches aim to enhance the model's ability to learn complex feature interactions and underlying feature structures within tabular data.

\noindent{\bf Automatic Feature Interaction Modules}. These methods do not assume specific feature types within the tabular dataset. Instead, they focus on improving the feature interaction process, enabling the model to learn complex, high-order feature relationships autonomously.  

DCNv2~\cite{WangSCJLHC21DCNv2} improves the learning of the model's feature interaction by improving the ``Cross Network'' structure. It employs low-rank methods to approximate feature crosses in subspaces and then integrates these subspaces using a gating mechanism. AutoInt~\cite{SongS0DX0T19AutoInt} maps the original sparse high-dimensional feature vectors into a low-dimensional space and models high-order feature interactions by stacking interaction layers with a multi-head attention mechanism. Unlike AutoInt, the TabTransformer\cite{Huang2020TabTransformer} only maps categorical features into contextual embeddings and feeds them into a Transformer model, while numerical continuous features are directly concatenated with the interacted contextual embeddings. When tabular data contains only numerical features, TabTransformer behaves in an MLP-like manner. Conversely, when the data contains only categorical features, TabTransformer operates similarly to AutoInt.

\noindent{\bf Implicit Feature Relationships}. Methods in this category typically assume that features in tabular data can be abstracted into implicit types and that it is necessary to design a suitable feature learning process to adapt to the characteristics of different types of features.  

DANets~\cite{ChenLWCW22DAN} propose the existence of underlying feature groups in tabular data, where features within each group are correlated. They learn to group input features and perform further feature abstraction. SwitchTab~\cite{Wu2024SwitchTab} introduces the idea of extracting sample-specific ``Salient Features'' and sample-shared ``Mutual Information'' in tabular features. It leverages self-supervised learning to assist in learning feature representations. ExcelFormer~\cite{Chen2023Excel} argues that while DNN assigns weights to each feature, it does not actively exclude irrelevant features. To address this, it introduces Semi-Permeable Attention for feature interaction, which allows features with lower information content to access information from more informative features while preventing highly informative features from being influenced by less relevant ones. AMFormer~\cite{cheng2024arithmetic} proposes the hypothesis that arithmetic feature interactions are crucial for deep tabular models. Based on the Transformer architecture, it introduces components designed to extract both additive and multiplicative interaction information.

%% file: transfer.tex
\section{From Specialized to Transferable Model}\label{sec:transfer}
Instead of training a tabular model from scratch, learning based on a Pre-Trained Model (PTM) may increase the learning efficacy and reduce the resource and data requirement.
For example, in a house prices prediction task, training a regressor in a certain area may benefit from a well-trained predictor from its neighborhood.

Learning by reusing the PTM usually contains two stages. The first is the pre-training of a tabular model, from one or more upstream tasks. Given the PTM and a downstream task, an adaptation strategy is needed to transform the PTM to the target task or facilitate the learning of the target model. Formally, a well-trained model $g_{\vTheta}$ is often available and can be leveraged to facilitate the training of $f_{\vtheta}$ over $\sD$. Here, $g_{\vTheta}$ is pre-trained on a dataset $\sD' = \{(\x_j', y_j')\}_{j=1}^{N'}$ with instances $\x_j'\in\R^{d'}$ and labels $y_j'\in [C']$. To reuse expert knowledge in $g_{\vTheta}$, an adaptation strategy is applied: $f_{\vtheta}=\textbf{Adapt}(f_{\vtheta_0} \mid \mathcal{D}, g_{\vTheta})$, where $\vtheta_0$ is the initialization of the model. The notation could also be extended to cases with more than one PTM. The main challenge to reuse one or more PTMs is to bridge the gap between the PTM and the target tabular model~\cite{Wang2025Survey}. 
We categorize PTMs into three kinds based on the source of PTM $g_{\vTheta}$. 

\noindent{\bf Homogeneous Transferable Tabular Model}. First, the PTM may come from the same form of task (with $d'=d$ and $C'=C$, but with different distributions $\Pr(\sD') \neq \Pr(\sD)$ or model families $g \neq f$). For example, those pre-trained from other domains~\cite{Rubachev2022revisiting}, or those unlabeled instances~\cite{YoonZJS20VIME,Somepalli2021SAINT}.  

\noindent{\bf Heterogeneous Transferable Tabular Model}. In addition, we consider a PTM pre-trained from a slightly different task with $\sD$. In addition to the previous difference, the PTM $g_{\vTheta}$ may differ from $f_{\vtheta}$ in feature dimension ($d' \neq d$) or target class set ($C' \neq C$), so the adaptation method $\textbf{Adapt}(\cdot)$ must handle such heterogeneity~\cite{Zhou2023TabToken,zhu2023xtab}.

\noindent{\bf Cross-Modal Transferable Tabular Model}. Moreover, the pre-trained model could also be constructed from another modality, such as vision and language domains. The cross-modality PTM is hard to be applied to the tabular prediction task in most cases, so auxiliary information from the tabular task like the semantic meaning of attributes (\ie, the attribute names) are usually assumed to be available in this case, where PTM like large language models may provide the latent semantic meanings as external knowledge~\cite{Wang2022TransTab,shen2023cross}.

The main limitation of the transferable tabular model is the assumption that the data distribution of the well-trained model should be similar to the distribution of the target model. For example in the previous house price prediction task, if the PTM is pre-trained in an area distance from the target area and targets diverse problems, it is hard to utilize the PTM in the target task~\cite{levin2022transfer}. 
Since different tabular tasks may vary in their distribution, feature, or classes, the general assumption is their exist a common ``dimension'' between the PTM and the target task. Only the distribution changes under the shared dimension and classes, or there exists an overlap between the feature or class spaces~\cite{zhu2023xtab}. 
For example, in real-world applications such as healthcare, there are numerous medical diagnostic tables. These tables usually have some features in common such as blood type and blood pressure. For rare diseases with limited data, knowledge transfer from other diagnostic tables with overlapping features becomes beneficial~\cite{Iwata2020Meta}.
When the feature/label semantics are available, two different tasks may be linked through the semantic space, and textual PTMs can be used to map the tabular instance to this space or facilitate the prediction in this space~\cite{Hegselmann2022TabLLM}.

\noindent{\bf Pros and Cons of transferable Models.} 
Learning with a well-trained tabular model has several advantages based on the knowledge encoded in the PTM. First, the training efficiency of the target model is improved and the model may converge fast, as the PTM may provide better initialization weights or optimization paths. Then, the target model will reduce the requirement on the data size, \ie, learning with a few-shot dataset. Training based on a PTM also reduces the number of learnable parameters, leading to parameter-efficient tuning and reducing computational resources.

\subsection{Homogeneous Transferable Tabular Model}
Adapting a tabular model from another domain with different distributions is investigated in the field of unsupervised domain adaptation before the era of deep learning. One representative method is the biased regularization, which minimizes the difference between the weights of the PTM and the target model, \ie,
\begin{equation}
    \min_{\mW} \ell(\mW) + \|\mW - \mW'\|_F^2 = \min_{\Delta\mW} \ell(\Delta\mW+\mW') + \|\Delta\mW\|_F^2\;.
\end{equation}
$\ell(\mW)$ is the loss function on the current weights $\mW$, and the regularize constraint the distance between the target model $\mW$ and the PTM weights $\mW'$. We can reformulate the learning objective as learning the weights residual $\Delta \mW$. Biased regularziation can be extended to the case where $f$ and $g$ are deep neural networks such as MLP, but it fails when the target model has a different architecture with the PTM. In this case, instead of matching two models through their weights, matching their predictions also helps. For example, twice learning~\cite{zhou2004nec4} and knowledge distillation~\cite{hinton2015distilling}. 

\begin{figure}[t]
    \includegraphics[width=0.49\textwidth]{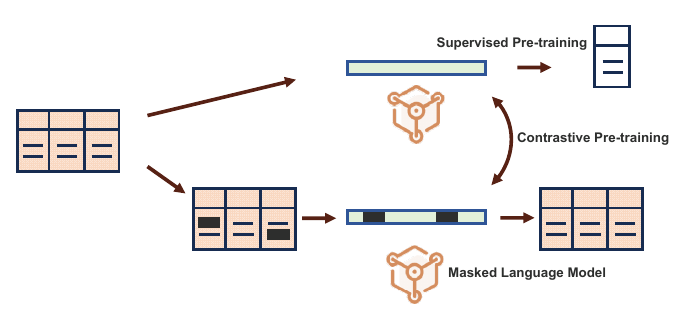}
    \caption{ Illustration of homogeneous transferable tabular methods. The pre-trained model could be constructed from supervised learning or self-supervised learning, which includes masked language model, contrastive pre-training, and hybrid methods.} 
    \label{figure:homo} 
    \vspace{-5mm}
\end{figure}

Benefiting from the strong capacity of deep neural networks, some recent studies focus on pre-training a tabular model from unsupervised instances, and then adapting the model via fine-tuning the PTM on the target (even few-shot) labeled examples. This strategy could be applied in standard supervised learning or semi-supervised learning.

\noindent{\bf Supervised Pre-training Objectives.} 
A straightforward way to incorporate the target variable into the pre-training is by using the input corruption as an augmentation for the standard supervised learning objective. \cite{Rubachev2022revisiting} identifies practices to pre-train tabular deep learning models that can be universally applied to different datasets and architectures. They show that using the object target labels during the pre-training stage benefits the downstream performance and advocates several target-aware pre-training objectives.

\noindent{\bf Self-Supervised Pre-training Objectives}. 
The self-supervised pre-training objectives can be mainly categorized into three categories, including the masked language model, contrastive pre-training, and hybrid methods.

\noindent{\textit{Masked Language Model (MLM).}} MLM is the unsupervised pre-training objective, where a random subset of features is masked for each sample, and the masked values are predicted in a multi-target classification manner~\cite{Huang2020TabTransformer}. 
VIME~\cite{YoonZJS20VIME} estimates mask vectors from corrupted tabular data and reconstructs feature vectors for self-supervised learning. They use the trained encoder to generate multiple augmented samples for each data point by masking each point using several different masks and then imputing the corrupted values for each masked data point. 
SubTab~\cite{UcarHE21SubTab} finds that reconstructing the data from the subset of its features rather than its corrupted version in an autoencoder setting can better capture its underlying latent representation. 
SEFS~\cite{Lee2022Self} reconstructs the original input based on a randomly selected subset of input features, and simultaneously estimates the gate vector that defines which features are selected or not.
MET~\cite{Majmundar2022MET} uses a concatenation of representations for all features instead of averaging and uses adversarial reconstruction loss in addition to the standard loss.

\noindent{\textit{Contrastive Pre-training.}}
Contrastive pre-training uses data augmentations to generate positive pairs or two different augmented views of a given example, and the loss function encourages a feature extractor to map positive pairs to similar features. The key factor in contrastive learning is to generate positive and negative versions of a given instance $\x_i$. 
\cite{Somepalli2021SAINT} utilizes CutMix~\cite{Yun2019CutMix} in the input space and Mixup~\cite{Zhang2018Mixup} in the embedding space to obtain positive pairs, where other instances $\x_{j\neq i}$ are treated as negative ones. 
SCARF~\cite{BahriJTM22Scarf} generates a view for a given input by selecting a random subset of its features and replacing them with random draws from their respective empirical marginal distributions. 
STab~\cite{Hajiramezanali2022STab} relies on two (or multiple) weight-sharing neural networks with different regularizations applied to a single input. By exploiting the stop-gradient operation technique, STab can model invariance with respect to more complicated regularizations while it will not collapse to an undesired trivial solution.
DoRA~\cite{DDu2023DORA} incorporates domain knowledge, training by intra-sample pretext task and inter-sample contrastive learning to learn contextualized representations.
DACL+~\cite{Verma2021DACL}, to overcome the reliance on a particular domain, uses Mixup noise to create similar and dissimilar examples by mixing data samples differently either at the input or hidden-state levels.

\noindent{\textit{Hybrid Methods.}}
\cite{levin2022transfer} explores several pre-training strategies including both supervised and unsupervised ones. 
It considers MLM as the unsupervised pre-training objective, and sets multi-label classification as the supervised pre-training objective.
By fine-tuning the PTM with several choices, including those with frozen feature extractor or not, the paper observes that supervised pre-training leads to more transferable features in the tabular domain.
LFR~\cite{Sui2024Self} conducts pre-training by learning to simultaneously reconstruct multiple randomly generated projection functions. It considers diverse data types to show the wide-ranging applicability of learning from randomness, including tabular, vision, and language.
ReConTab~\cite{Chen2023ReConTab} utilizes both self-supervised learning and semi-supervised learning. It uses regularization techniques for raw feature selection and leverages contrastive learning with labels to distill the most pertinent information for downstream tasks. 
~\cite{Rubachev2022revisiting} focuses on the setup with fully labeled tabular datasets to understand if pretraining helps tabular deep learning in a fully supervised setting and compares pretraining methods to the strong supervised baselines. They show that using the object target labels during the pertaining stage is beneficial for the downstream performance and advocate several target-aware pretraining objectives.
~\cite{Wang2025Survey} provides a systematic review and summarizes the recent progress and challenges of self-supervised learning for non-sequential tabular data.

\subsection{Heterogeneous Transferable Tabular Model}
The main intuition lies in the mapping $f$ and $g$ work in a similar fashion, \ie, predicting the labels with similar mechanisms. Therefore, the main idea to transfer knowledge is to match the target model with the well-trained one, over the weight space or the prediction space.

\begin{figure}[t]
    \includegraphics[width=0.49\textwidth]{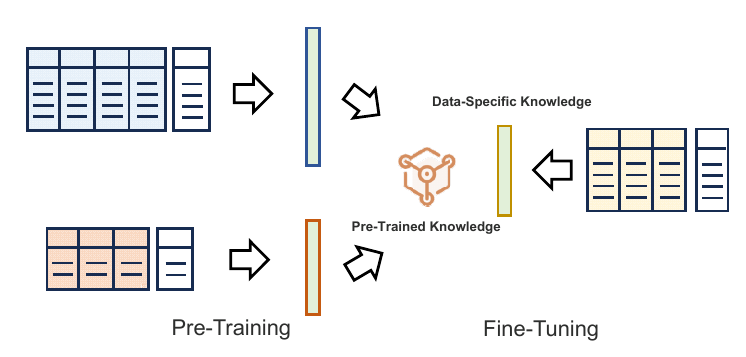}
    \caption{ Illustration of heterogeneous transferable tabular methods. During pre-training on one or multiple datasets, most of the parameters in the PTM are trained. For downstream tasks, only a small subset of parameters is fine-tuned while the rest remain fixed. } 
    \label{figure:heter} 
    \vspace{-5mm}
\end{figure}

Early methods mainly focus on the feature-level heterogeneity between $f$ and $g$. One main assumption is that there exists a shared set of features between the pre-trained task $\sD'$ and the target task $\sD$, then we may directly copy the weights corresponding to the shared features from the PTM. 
Some methods extend bias regularization to deal with heterogeneous feature spaces by padding the weights with zero.
OPID~\cite{hou2017one} is a one-pass learning approach, which only needs to scan each instance once and to deal with evolving streams. In the pre-training stage, OPID compresses important information of vanished features into functions of survived features, and in the adaptation stage, it is expanded to include the augmented features. 
ReForm~\cite{Ye2018ReForm} learns the meta-representation for each feature and based on which calculates the relationship between features in the meta-representation space. ReForm then bridges the feature space gap through optimal transport, which could be further used to transform classifiers with different features and classes. 

A major advantage of neural models is that they are easily fine-tuned in new domains and learn reusable features. For example, as the deep PTM has the ability to extract generalizable features for a tabular task, reusing the knowledge from the PTM can utilize the strategies designed for visual and language domains. In detail, we can fix most of the parameters in the PTM and tune the remaining parts which only have limited parameters, for example, the linear probing or parameter-efficient fine-tuning.

\noindent{\bf Reuse PTM Pre-trained from One Dataset}.
These methods primarily focus on the difference between the pre-trained and down-streaming datasets. TabRet~\cite{Onishi2023TabRet} utilizes masked autoencoding to make the transformer work in downstream tasks. 
To transfer pre-trained large language models to tabular tasks, ORCA~\cite{shen2023cross} trains an embedder to align the source and target distributions. 
TabToken~\cite{Zhou2023TabToken} focuses on improving the quality of the feature tokens, which are an important component in tabular deep models. TabToken leverages a conditional contrastive loss to improve the quality of learned embeddings and demonstrates enhanced transferability of deep learning models for tabular data.

Pseudo-Feature method~\cite{levin2022transfer} utilizes pseudo-feature models individually for each new feature. In detail, given one additional feature in a downstream dataset, it first pre-trains a model on the upstream data without that feature. Then Pseudo-Feature fine-tunes the pre-trained model on downstream data to predict values in the column absent from the upstream data. Next, the fine-tuned model is used back in the upstream datasets to predict and assign pseudo-values of this feature. After supplementing the upstream dataset with the ``unseen'' feature in the downstream task, Pseudo-Feature pre-trains and transfers the feature extractor to the downstream task again. This method is computationally expensive in our broader feature space adaptation scenario.

\noindent{\bf Reuse PTM Pre-trained from Multiple Datasets}.
XTab~\cite{zhu2023xtab} aims to enhance the transferability of the transformer. They address the challenge of inconsistent column types and quantities among tables by utilizing independent features and federated learning to pre-train the shared component. 

Another thread of method learns shared components such as attribute-agnostic transformation across datasets, which provides a good model initialization for partial parameters given a downstream task. \cite{Iwata2020Meta} infers latent representations of each attribute and each response from a few labeled instances using an inference network.  The attribute and response representations are enabled make predictions based on the task-specific properties of attributes and responses even when attribute and response sizes are different across tasks. 
DEN~\cite{Liu2022Distribution} uses a three-block architecture: a covariate transformation block followed by a distribution embedding block and then a classification block. They provide theoretical insights to show
that this architecture allows the embedding and classification blocks to be fixed after pre-training on a diverse set of tasks.
Meta-Transformer~\cite{Zhang2023Meta} leverages a frozen encoder to perform multimodal perception without any paired multimodal training data. In Meta-Transformer, the raw input data from various modalities are mapped into a shared space in meta learning~\cite{ye2022revisiting}, allowing a subsequent encoder with frozen parameters to extract high-level semantic features.

\subsection{Reusing a Pre-trained Language Model}
In some cases, the semantic meaning of features is available, making it natural to leverage pre-trained language models for tabular data. Typically, two types of semantic information can be derived from a tabular dataset $\mathcal{D}$. First, attribute names for each of the $d$ features, $\mathcal{A} = {A_1, \ldots, A_d}$, provide useful context. Additionally, meta-information such as a textual description, denoted as $\text{meta\_descript}$, can further enhance understanding. The learning process is then formulated as:
\begin{equation}
\hat{y}_i = f(\x_i, \mathcal{A} \mid \mathcal{D}, \text{meta\_descript})
\end{equation}
where the semantic information bridges the gap between feature spaces and facilitates knowledge transfer from pre-trained tasks to downstream applications.

\begin{figure}[t]
    \includegraphics[width=0.49\textwidth]{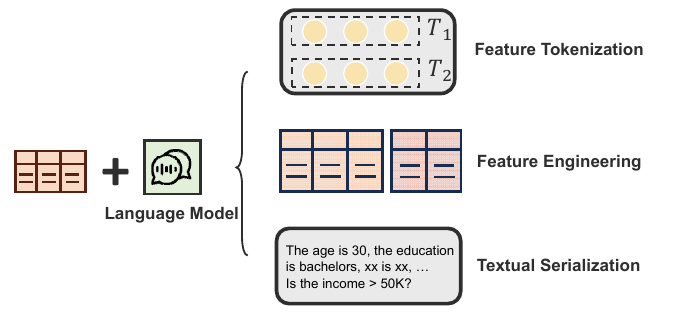}
    \caption{ Illustration of transferable tabular methods with a language model. The language model can be applied at various stages, including feature tokenization, feature engineering, and textual serialization. } 
    \label{figure:language} 
    \vspace{-5mm}
\end{figure}

Although pre-trained language models have demonstrated success in various domains, their application to tabular data remains limited due to the prevalence of numerical values and the scarcity of textual descriptions. Moreover, concerns about data privacy and security may further restrict access to semantic information. Consequently, language models are typically applied to tabular datasets only when textual context is sufficiently available.

\noindent\textbf{Language Models for Feature Tokenization.}
When the feature space changes, language-based methods assume that semantic relationships exist between feature descriptions and rely on large-scale language models to capture these connections. For example, the feature "occupation" in one task may share semantic similarity with the feature "organization" in another, allowing feature-label relationships to be reused across different datasets. By extracting feature embeddings (tokens), tables of varying sizes can be transformed into a standardized set of tokens in a shared space. A pre-trained transformer then encodes transferable knowledge, aiding the fine-tuning process for downstream tasks.

TransTab~\cite{Wang2022TransTab} trains a tokenizer based on the words present in tabular data and incorporates both column descriptions and table cells as raw input to a gated transformer model. The model is pre-trained via self-supervised learning or supervised contrastive loss and is validated on tasks such as transfer learning and feature incremental learning. PTab~\cite{Liu2022PTab} adopts a similar approach, learning contextual representations from multiple tokenized tabular datasets before fine-tuning for downstream tasks. UniTabE~\cite{Yang2024UniTabE} encodes and fuses information from column names, data types, and cell values into a set of tokens, applying an encoder-decoder architecture with Transformer and LSTM components. It is pre-trained using Multi-Cell-Masking and contrastive learning, where a sub-vector of an instance is treated as a positive sample while other instances or their subsets are considered negatives.

CM2~\cite{Ye2024Towards} introduces a cross-table pre-training framework that integrates attribute names and feature values. CM2 uses transformers to process feature tokens and employs a prompt-based Masked Table Modeling (pMTM) self-supervised objective, where column names act as prompts to assist in predicting masked features. TP-BERTa~\cite{Yan2024Making} follows a similar approach but incorporates numerical discretization strategies and magnitude tokenization for feature encoding, fine-tuning smaller pre-trained language models such as RoBERTa~\cite{liu2019roberta} for tabular data prediction. Its pre-training objective includes supervised loss and magnitude-aware triplet loss as a regularizer.

CARTE~\cite{Kim2024CARTE} utilizes a graph representation of tabular data to handle heterogeneous feature spaces, transforming textual information from column names and entries into embeddings. A graph-attentional network is then applied to contextualize entries with column names and neighboring entries. CARTE is pre-trained on the YAGO3 knowledge base~\cite{Mahdisoltani2015YAGO3} by constructing graphlets for tabular data and employing contrastive loss, where the original graphlet and one truncated variant are positives, while other graphlets in the batch serve as negatives. The pre-trained CARTE model is subsequently fine-tuned for downstream tasks.

\noindent\textbf{Language Models for Feature Engineering}.
Discriminative features enhance the effectiveness of subsequent tabular learning models. Binder~\cite{Cheng2023Binding} identifies task input components that are not directly answerable by a model and leverages LLMs to generate auxiliary features, particularly for knowledge grounding tasks. Given that discriminative features are often manually designed, CAAFE~\cite{hollmann2024large} explores the use of LLMs to generate auxiliary features based on task and feature semantics. The quality of these features is then evaluated using a general tabular model, TabPFN~\cite{Hollmann2022TabPFN}. FeatLLM~\cite{han2024large} enhances feature generation by incorporating example-based prompting, enabling LLMs to create new features based on textual descriptions. TaPTaP~\cite{Zhang2023TapTap} is expected to capture a generic tabular data distribution after ongoing pre-training on a large-scale corpus of real-world tabular data, generating high-quality synthetic tables to support various applications on tabular data.

\noindent\textbf{Language Models for Textual Serialization.}
A direct approach to incorporating pre-trained language models involves converting tabular data into a textual format, allowing LLMs to infer relationships between features and labels based on embedded expert knowledge. This concept has been validated in semantic parsing tasks~\cite{Herzig2020TaPas,Yin2020TaBERT}. LIFT~\cite{Dinh2022LIFT} and TabLLM~\cite{Hegselmann2022TabLLM} serialize tabular data by integrating feature names into text and combining them with task descriptions. This enables LLMs to treat tabular prediction tasks as text generation problems. LIFT fine-tunes models on the entire training set, while TabLLM employs few-shot learning for fine-tuning. UniPredict~\cite{Wang2023UniPredict} constructs prompts using metadata, sample serialization, and task instructions, fine-tuning LLMs with confidence-weighted augmented labels predicted by an external model. The approach is validated on multiple in-distribution datasets.

Despite their advantages, textual serialization methods face challenges when the number of features increases, as prompts may become too large to fit within the model's context window. The effectiveness of LLMs in tabular data tasks remains constrained by the availability of semantic information and the capabilities of external tabular models. Further exploration of LLM-based methods will be discussed in the general tabular models in~\autoref{sec:general}.

\subsection{Reusing a Pre-trained Vision Model}
Given the success of deep neural networks (DNNs) in visual tasks, it is intuitive to leverage the strong recognition capabilities of pre-trained vision models for tabular data. Additionally, data augmentation strategies commonly used in image processing can be introduced after transforming tabular data into a visual format. Similar ideas have been explored in time series forecasting~\cite{Chen2024VisionTS} and irregular time series classification~\cite{Li2023Time}.

\begin{figure}[t]
    \includegraphics[width=0.49\textwidth]{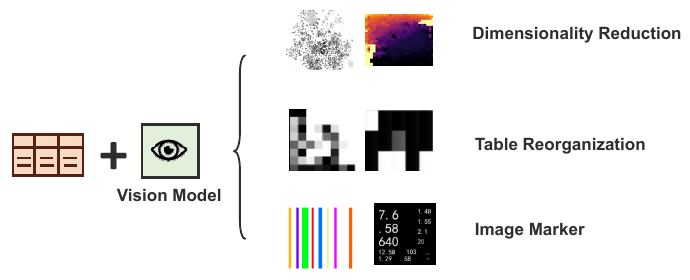}
    \caption{ Illustration of transferable tabular methods with a vision model. Tabular data can be transformed into images through dimensionality reduction, table reorganization, and the use of image markers. } 
    \label{figure:vision} 
    \vspace{-5mm}
\end{figure}

The primary challenge lies in representing tabular instances in an image-compatible format. In natural images, neighboring pixels often share semantic relationships, whereas tabular data lacks inherent spatial structure. Features in a tabular instance are permutation-invariant, meaning that exchanging their order does not alter the instance’s meaning. Various methods have been proposed to transform tabular data into visual representations, enabling the application of pre-trained vision models fine-tuned for tabular tasks. This subsection highlights different transformation strategies that transfer tabular datasets into images.

\noindent\textbf{Dimensionality Reduction Transformation.}
Visualization strategies for tabular data naturally convert tables into images by embedding high-dimensional features into a lower-dimensional space. DeepInsight~\cite{Sharma2019DeepInsight} projects tabular data into a 2D space using t-SNE and constructs images through convex hull analysis, applying translation, rotation, quantization, and normalization. REFINED~\cite{Bazgir2020REFINED} employs Bayesian Metric Multidimensional Scaling to preserve pairwise distances within the low-dimensional representation, ensuring that structurally similar features remain proximate in the transformed image.

\noindent\textbf{Table Reorganization Transformation.}
A tabular dataset $\mathcal{D}$ can be treated as a matrix and represented as a single-channel image or kernel. To enable visual PTMs to recognize meaningful spatial relationships, different strategies have been developed for structuring tabular data into images. Tabular Convolution (TAC)~\cite{buturovic2020novel} arranges data samples into zero-mean square matrices (kernels) of odd integer dimensions. These kernels are then convolved with a fixed ``base image,'' and the resulting images are subsequently fed to a CNN for classification. Image Generator for Tabular Data (IGTD)~\cite{Zhu2021IGTD} and TablEye~\cite{Lee2023TablEye} share a similar idea, generating an image for each data sample where pixel intensities correspond directly to feature values. These methods prioritize placing similar features in close proximity but struggle with high-dimensional tabular tasks. LM-IGTD~\cite{Vanesa2024LM-IGTD} extends IGTD by incorporating stochastic feature generation to enhance robustness and generalization.

\noindent\textbf{Image Marker Transformation.}
Another approach involves encoding feature values as visual markers within an image. Super-TML~\cite{Sun2019SuperTML} assigns feature values to predetermined positions within an image, effectively handling categorical and numerical datasets. Tab2Visual~\cite{Mamdouh2025Tab2Visual} normalizes tabular data and represents each instance as a row of multiple bars, each corresponding to a specific value. Each feature is assigned a unique color to enhance visual differentiation, while bar widths are proportional to feature magnitudes.

By transforming tabular data into images, these methods enable the application of powerful pre-trained vision models to tabular prediction tasks, leveraging established deep learning techniques from the vision domain to enhance tabular model performance.

%% file: general.tex
\section{From Transferable to General Model}\label{sec:general}

The general model (also referred to as the tabular foundation model) represents an advancement over the transferable model. It extends the generalization capabilities of a pre-trained tabular model to a variety of heterogeneous downstream tabular tasks, regardless of their diverse feature and class spaces, {\em without requiring additional fine-tuning}. In other words, given a pre-trained model $g_{\vTheta}$, it can be directly applied to a downstream tabular task $\sD$ to predict the label of a test instance $\x^*$ as follows:
\begin{equation}
    \hat{y}^* = g_{\vTheta}(\x^* \mid \sD)\;.
\end{equation}
Thus, the general model shares similarities with the transferable tabular model, but with a greater emphasis on the ``zero-shot'' ability, aims to construct highly adaptive architectures capable of handling a wide array of heterogeneous datasets simultaneously. Importantly, it does not require an {\bf Adapt} function, which further reduces the computational cost of hyper-parameter tuning. The goal of the general tabular model is to achieve better generalization on downstream tabular datasets $\sD$ when compared to alternative strategies, such as training a tabular model directly on $\sD$ or adapting a transferable model.

\begin{remark}
    Distinguishing between an advanced transferable tabular model, pre-trained on a wide range of heterogeneous tabular tasks, and the general tabular model can be challenging. Some transferable tabular models, based on auxiliary feature semantics, are able to predict labels for downstream test instances directly~\cite{Hegselmann2022TabLLM}. However, their prediction ability is constrained and typically applicable only in specific areas after fine-tuning~\cite{Yan2024Making,Kim2024CARTE}. The general tabular model, on the other hand, is designed to handle a wider range of heterogeneous tabular tasks, sharing similar pre-training challenges with transferable models but without utilizing additional semantics. Fine-tuning a pre-trained general model is also an option for further performance improvements~\cite{WenZZXB24From,Liu2025Beta}.
\end{remark}

Pre-training has revolutionized domains such as vision and language~\cite{Kirillov2023Segment, zhou2024comprehensive}, but its adoption in tabular data remains limited due to the inherent heterogeneity of tabular datasets. Tabular datasets can vary significantly in both dimensionality (\ie, the number of columns) and the semantic meaning of each dimension, even within the same application. For example, different healthcare datasets may capture varying levels of detail and aspects of patient information. Even within the same feature entry (\eg, the $d$-th column), the meaning can vary (\eg, ``age'' vs.~``height''). This contrasts with vision and text data (within the same language), where different data sources typically share the same ``vocabulary'' (\eg, pixels, patches, or sub-words) and similar relationships between vocabulary ``elements'' (\eg, neighboring pixels often share colors). The lack of shared vocabulary and relationships in tabular data makes it challenging to jointly train a model across multiple datasets, let alone apply a pre-trained model directly to new downstream tasks.

\begin{figure}[t]
    \includegraphics[width=0.49\textwidth]{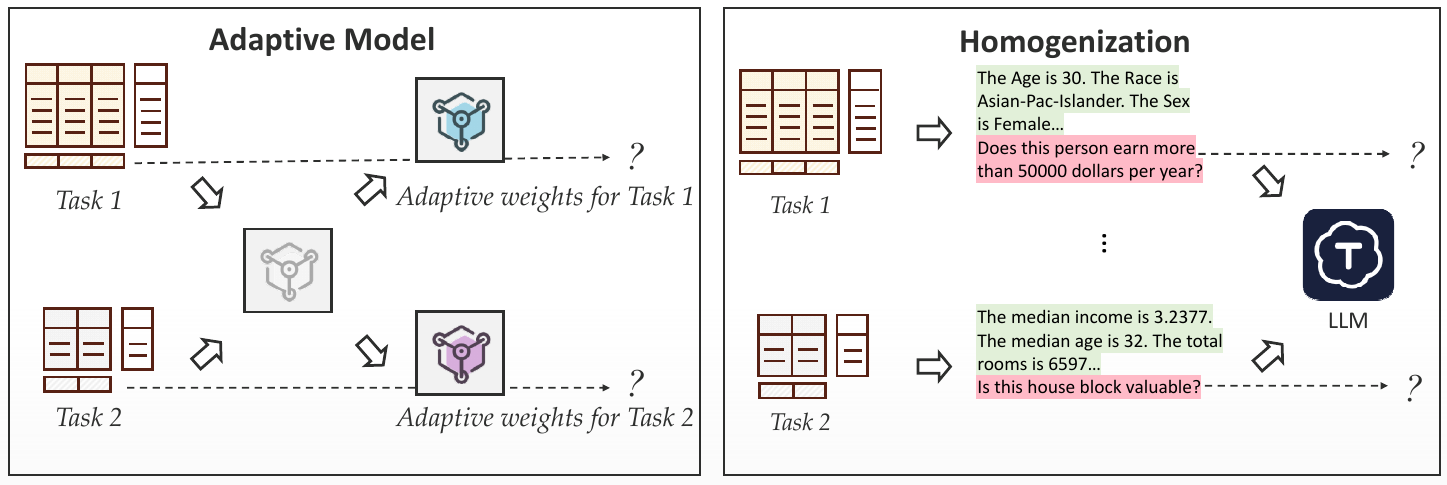}
    \caption{  Illustration of general methods. These methods handle inherent heterogeneity by improving the model’s adaptability or homogenizing the diverse tabular formats. Once pre-trained, they can be directly applied to downstream tasks without fine-tuning.} 
    \label{figure:general} 
    \vspace{-5mm}
\end{figure}

There are two main strategies to address the inherent heterogeneity in tabular datasets: improving the model's adaptability or homogenizing the diverse tabular formats. 
We categorize general tabular models into three parts based on their strategies for achieving generalizability. The first focuses on raw-feature-based approaches, among which TabPFN variants represent a rapidly evolving branch and are thus discussed separately. The third category encompasses semantic-based methods that leverage attribute and task semantics to unify heterogeneous tasks.

\subsection{\bf Raw-Feature-based General Models}
To adapt a general tabular model to heterogeneous tabular datasets during the pre-training and fine-tuning stages, two main strategies can be used from the data-centric and model-centric perspectives. 
From the data-centric perspective, the general model may standardize tabular datasets into a homogeneous form. For instance, TabPTM~\cite{Ye2023TabPTM} transforms all datasets into a uniform format using meta-representation to enable pre-training. The pre-trained model can then be applied directly to a downstream dataset or fine-tuned without introducing additional parameters.

Alternatively, from the model-centric perspective, the general model may improve adaptability by tailoring it to specific tabular tasks. HyperFast~\cite{BonetMGI2024HyperFast} adopts the concept of a Hyper Network~\cite{Ha2017HyperNet} in meta-learning~\cite{Chao2020Meta}, where a mapping from the tabular dataset to the weights of a classifier is learned. This mapping can then be used to predict labels for unseen instances from the task. To address datasets with varying dimensions, HyperFast projects datasets into a fixed size using random projections. To overcome the slow weight generation speed, MotherNet accelerates HyperFast by modifying its architecture with Transformer-like modules~\cite{Muller2023MotherNet}.

\subsection{\bf TabPFN Variants}
The TabPFN family of models~\cite{Hollmann2022TabPFN, hollmann2025tabpfn} leverages the in-context learning capabilities of transformers, directly predicting labels by adapting test instances according to the context of training examples. In the first version of TabPFN, an instance $\x_i$ is padded to a fixed dimension (\eg, 100), and the features are projected to a higher dimension (\eg, $d'$) for further processing. The label $y_i$ is processed similarly and added to the instance embeddings. The embeddings of all $N+1$ instances, including training and test instances, are formulated into a set of $N+1$ tokens with $d'$ dimensions. These tokens are processed through several layers of a Transformer, and the output token corresponding to the test instance is further predicted using a 10-way classifier. TabPFN is pre-trained over synthetically generated datasets with structured causal models (SCM)~\cite{causalinferencelearningbook} and  Bayesian Neural Networks (BNNs)~\cite{neal-bayes96a,0005HPGH22}, enabling the strong in-context learning ability, with the best checkpoint selected based on some real-world datasets. Due to the high complexity of transformers, TabPFN is limited to small-scale tasks, with suggested sizes of $N<1000$, $d<100$, and $C<10$.

TabPFN v2 introduces a specialized feature tokenizer to better handle heterogeneity. Specifically, each cell in the table is projected to a $k$-dimensional vector using a shared mapping, and random position encoding vectors are added to differentiate features~\cite{ye2025closer}. This results in a tensor of size $(N+1)\times (d+1)\times k$ when there is a single test instance. The label of each instance is processed similarly, and the mapped $k$-dimensional token is concatenated with the instance tokens. A dummy label (\eg, the average of all labels) is used for the test instance since its label is unknown. A two-way attention mechanism is used, with each feature attending to the other features in its row and then attending to the same feature across its column~\cite{Iwata2023Meta}.
The output token corresponding to the label of the test instance is further mapped to a 10-class classifier or regressor. Several improvements have been made in TabPFN v2, including increased context size ($N<10000$, $d<500$), automatic feature engineering, and post-hoc ensemble methods. \cite{Nagler2023Statistical} analyzes TabPFN from a bias-variance perspective, shedding light on its generalization capabilities. Various applications have also been explored, including tabular data generation~\cite{Ma2024TabPFGen}, anomaly detection~\cite{RuizVillafrancaGGMM24detection}, data augmentation~\cite{TabMDA}, and time series forecasting~\cite{Shi2024TabPFNSeries}. 

The improvements of TabPFN (especially TabPFN v1) stem from several aspects.\footnote{Some variants of TabPFN are not considered general tabular models, especially the latter parts, as they require additional fine-tuning steps. We place them in this subsection due to their strong relationship with TabPFN.}

\noindent{\bf Pre-training Improvements.}
TabForestPFN~\cite{Breejen2025TabForestPFN} extends TabPFN by pre-training In-Context Learning (ICL)-transformers on a new forest dataset generator that creates unrealistic datasets with complex decision boundaries. 
TabDPT~\cite{Ma2024TabDPT} pre-trains the architecture on real-world datasets using self-supervised learning and retrieval objectives, making it suitable for both classification and regression tasks. 
APT~\cite{wu2025zeroshotmetalearningtabularprediction} is pre-trained utilizing adversarial synthetic data generated by adaptive agents, which systematically modify the underlying data-generating distribution and deliberately challenge the model with diverse synthetic datasets to enhance its robustness and generalization capabilities. TabICL~\cite{qu2025tabicltabularfoundationmodel} integrates tree-based SCMs using XGBoost~\cite{chen2016xgboost} to model complex interactions and employs curriculum learning by progressively increasing synthetic dataset sizes.

\noindent{\bf Scalable Improvements.}
The efficiency of TabPFN is highly sensitive to context size, prompting strategies to enhance scalability and performance~\cite{McElfreshKVCRGW23when}.
These include compressing training data into a compact learned representation using sketching~\cite{Feuer2023ScalePFN} or prompt tuning techniques~\cite{Distillation_PFN,Feuer2024TuneTables}, employing adaptive data selection methods to identify the most pertinent training examples for each test instance~\cite{Xu2024MixPFN,Thomas2024LocalPFN,Ma2024TabDPT,koshil2024towards}, and replacing traditional quadratic attention with computationally efficient linear attention mechanisms~\cite{zeng2024tabflex} and state-space models (SSMs)~\cite{baur2024exploration}.

\noindent{\bf Adaptation Improvements.}
Some approaches improve TabPFN’s performance on downstream tasks by adapting the context~\cite{Thomas2024LocalPFN} or fine-tuning specific parts of the 
model~\cite{Liu2025Beta,Breejen2025TabForestPFN,Xu2024MixPFN,Feuer2024TuneTables}. TabICL~\cite{qu2025tabicltabularfoundationmodel} employs a column-then-row attention mechanism to construct fixed-dimensional embeddings of rows, which are subsequently processed by a transformer like TabPFN v1 to facilitate efficient in-context learning. EquiTabPFN~\cite{arbel2025equitabpfntargetpermutationequivariantprior} introduces self-attention across target components, ensuring that the arbitrary ordering of target dimensions does not influence model predictions, enhancing the performance of TabPFN v1 to some extent.

\subsection{\bf Semantics-based General Models}
By leveraging the semantic structure of tabular data, such as column names, heterogeneous tasks can be projected into a shared language space. This allows a single language model, pre-trained on diverse tabular datasets, to handle unseen tasks in a unified manner. 
TabuLa-8B~\cite{tabula8b} fine-tunes a Llama 3-8B LLM for tabular data prediction (classification and binned regression) using a novel packing and attention scheme for tabular prediction. 
GTL~\cite{WenZZXB24From} transforms tabular datasets into an instruction-oriented language format, facilitating the continued pre-training of LLMs on instruction-oriented tabular data, which demonstrates strong performance in few-shot scenarios. 
GTL-S~\cite{sun2024scaling} unlocks the potential of GTL from a scaling perspective, revealing that scaling datasets and prediction tasks enhance generalization.
\cite{Wen2025ICL} extends GTL by incorporating retrieval-augmented LLMs for tabular data, combined with retrieval-guided instruction-tuning for LLMs. 
MediTab~\cite{Wang2024MediTab} uses a data engine that leverages LLMs to consolidate tabular samples to overcome the barrier across tables with distinct schema. MediTab aligns out-domain data with the target task using a “learn, annotate, and refinement” pipeline, enabling the pre-trained model to infer for arbitrary tabular input in the domain without fine-tuning.

%% file: Ensemble.tex
\section{Tabular Ensemble Methods}\label{sec:ensemble}
Ensemble learning is a natural way to improve the generalization ability of multiple base learners by leveraging their diversity. Classical methods such as Random Forest~\cite{Breiman01RandomForest} and AdaBoost~\cite{freund1995desicion,freund1996experiments} employ bagging and boosting, respectively, by ensembling multiple decision trees. These methods have proven effective for tabular data, as they reduce bias/variance and improve robustness~\cite{zhou2012ensemble}.

In deep tabular learning, ensemble methods can be categorized into two primary approaches: joint-training ensembles, where multiple sub-networks are aggregated within a single training pipeline, and post-hoc ensembles, where the predictions from multiple pre-trained deep tabular models are fused. One major challenge in ensembling deep tabular methods is computational efficiency, as training multiple deep models or sub-models can be computationally expensive and time-consuming.

\subsection{Joint-Training Ensembles}
Joint-training ensemble methods integrate diverse model architectures within a single training process to improve predictive performance while maintaining efficiency. These architectures often combine different types of models, such as linear and non-linear models~\cite{Cheng2016Wide} or tree-based and deep neural network-based approaches~\cite{Huang2020TabTransformer}. Tree-mimic methods leverage this concept by mixing predictions from multiple tree nodes to enhance robustness~\cite{PopovMB20Neural,Badirli2020GrowNet,Marton2024GRANDE}.

To improve efficiency while maintaining predictive power, various techniques have been explored. Some approaches employ parameter-efficient ensembles, such as TabM~\cite{Yury2024TabM}, which uses MLPs as base learners and incorporates BatchEnsemble~\cite{WenTB20BatchEnsemble} to generate multiple diverse base learners efficiently. This prevents a large increase in the number of learnable parameters while maintaining model diversity. Similarly, BETA leverages pre-trained TabPFN by generating multiple base learners through additional parameter tuning~\cite{Liu2025Beta}. Specifically, BETA learns multiple feature projections, feeding the projected training sets into TabPFN and aggregating the results while applying BatchEnsemble to reduce the number of additional learnable parameters.

Some hybrid approaches, such as LLM-Boost and PFN-Boost, have been developed to integrate large language models and TabPFN with gradient-boosted decision trees~\cite{Jayawardhana2025Boost}. In these approaches, LLMs and PFN serve as the initial base learners, and additional base learners are sequentially trained in a boosting manner. This approach leverages the strong prior knowledge from LLMs and TabPFN while maintaining the scalability of gradient-boosted decision trees.

\subsection{Post-Hoc Ensembles}
Post-hoc ensemble (PHE) methods involve combining multiple trained models to improve robustness and accuracy. 
Bagging-based ensembles are one of the most direct post-hoc strategies, where usually multiple models trained with different random seeds are aggregated~\cite{GorishniyRKB21Revisiting,gorishniy2023tabr}. Although this approach improves model robustness, it incurs high computational overhead. Some recent studies have demonstrated that LLM-based methods exhibit diverse prediction behaviors compared to deep tabular models that do not utilize attribute names~\cite{Wen2025ICL}. This difference in prediction styles enhances their complementarity, making them ideal candidates for ensemble methods.

Instead of explicitly training multiple models, perturbation-based approaches create diverse predictions from the same pre-trained model. One such method applies feature permutation with TabPFN, leveraging the fact that TabPFN is not fully feature permutation-invariant~\cite{Hollmann2022TabPFN}. A perturbation-based ensemble can be formed by randomly permuting the feature order in both the training and test sets and making predictions multiple times, generating multiple diverse predictors without additional training costs. TabPFN v2 introduces additional perturbations to enhance diversity among several key factors, including variations in feature encoding, feature quantization, categorical feature shuffling, SVD-based feature compression, outlier removal, and power transformations such as the Yeo–Johnson transformation~\cite{hollmann2025tabpfn}. These randomly selected transformations create diverse prediction patterns, enabling effective ensemble learning without requiring multiple separately trained models.

Another post-hoc ensemble strategy employed in TabPFN v2 is the use of Portfolio-Based Ensemble, where a fixed set of TabPFN configurations is used~\cite{hollmann2025tabpfn}. A greedy ensemble selection technique is then applied to learn optimal weights for aggregating the predictions of different configurations~\cite{Caruana2006GreedyEnsemble}. By combining multiple perturbed models, this method improves generalization without excessive training costs. Some methods applies ensemble techniques to TabPFN v1 to handle large datasets. For instance, TabPFN-Bagging~\cite{Liu2025Beta,wang2025priorfittednetworksscalelarger} divides large datasets into multiple context groups, with the final results averaged to mitigate variance. BoostPFN~\cite{wang2025priorfittednetworksscalelarger} treats TabPFN v1 as weak learners, where each weak learner uses a subset of the training data as context. This approach allows BoostPFN to outperform standard Prior Fitted Networks (PFNs) on large datasets.

%% file: extensions.tex
\section{Extensions}\label{sec:extension}
In this section, we briefly introduce some extensions on deep tabular methods across different complex tasks.

\noindent{\bf Clustering.} 
Traditional clustering approaches often leverage enhanced distance metrics, such as the Gower distance~\cite{gower1971general}, which is specifically designed for mixed data types, and interpretable prototypes, such as K-medoids. Recent advances in tabular data clustering have sought to integrate interpretability constraints with deep representation learning. For example, IDC~\cite{SvirskyL24Interpretable} introduces a deep learning framework for general tabular data that predicts interpretable cluster assignments at both the instance and cluster levels. To address overlapping clusters, TableDC~\cite{Rauf2024TableDC} integrates the Mahalanobis distance, which accounts for variance and correlation within the data. This method provides a similarity measure suitable for tables, rows, or columns in high-dimensional latent spaces.

\noindent{\bf Anomaly Detection.}
Anomaly detection in tabular data is crucial for identifying subtle irregularities in structured datasets, such as fraudulent transactions or equipment failures. While classical techniques like Isolation Forest~\cite{liu2008isolation} and Local Outlier Factor~\cite{breunig2000lof} remain foundational, recent developments have incorporated various methods to capture contextual relationships in high-dimensional data. For instance, \cite{shenkar2022anomaly} introduces a method that learns mappings that maximize mutual information between each sample and the part that is masked out, capturing the structural nuances of samples from a single training class. ADBench~\cite{Han2022ADBench} provides a comprehensive tabular anomaly detection benchmark
with 30 algorithms and 57 benchmark datasets. Additionally, large language models (LLMs) have also been employed for anomaly detection in tabular data~\cite{li2024anomaly}.

\noindent{\bf Tabular Generation.}
Tabular data generation has become an essential tool for synthetic data creation, privacy preservation, and addressing data scarcity. Traditional methods, such as Bayesian networks or GANs, focus on mimicking marginal distributions, while recent advancements emphasize preserving complex feature dependencies and semantic consistency. For instance, tabular diffusion models~\cite{lee2023codi} iteratively refine synthetic data to capture subtle correlations in high-dimensional datasets, outperforming GANs in terms of data fidelity. \cite{tu2024causality} introduces high-order structural causal information as a natural prior knowledge and offers a benchmark framework for evaluating tabular synthesis models. Despite these advances, challenges remain in balancing realism with privacy, such as avoiding identity leakage in sensitive datasets, and scaling to heterogeneous data types. Hybrid neuro-symbolic models~\cite{feinman2020generating} bridge this gap to provide trustworthy synthetic data for downstream tasks.

\noindent{\bf Interpretability.}
Traditional gradient-boosted decision trees (GBDTs) inherently provide interpretability through feature importance scores and decision path visualization. Frameworks such as XGBoost~\cite{chen2016xgboost} and LightGBM~\cite{ke2017lightgbm} quantify feature importance using metrics like split frequency and information gain. SHAP values~\cite{hastie2009elements} enable instance-level explanations by decomposing model predictions into feature contributions. The additive nature of GBDTs allows for partial dependence plots~\cite{greenwell2017pdp} to visualize feature effects while controlling for interactions. NeC4.5~\cite{zhou2004nec4}, a novel decision tree algorithm that integrates the comprehensibility of decision trees with the generalization ability of neural network ensembles. By training a neural network ensemble to generate a new training set, NeC4.5 enhances decision tree performance while maintaining interpretability.

Recent deep models specifically designed for tabular data have introduced novel interpretability mechanisms. For example, NAMs~\cite{agarwal2021neural} combine some of the expressivity of DNNs with the inherent intelligibility of generalized additive models. They learn a linear combination of neural networks that each attend to a single input feature, which are trained jointly and can learn arbitrarily complex relationships between their input feature and the output. TabNet~\cite{ArikP21TabNet} uses sequential attention with learnable feature masks, where each decision step explicitly selects a subset of features via sparse masking. The aggregated feature usage across steps provides global interpretability comparable to GBDT's feature importance. Subsequent variants, such as TabTransformer~\cite{Huang2020TabTransformer}, enhance interpretability by visualizing cross-feature attention patterns. FT-Transformer~\cite{GorishniyRKB21Revisiting} combines feature tokenization with explainable attention, while NODE~\cite{PopovMB20Neural}, NODE-GAM~\cite{Chang0G22NODEGAM} and DOFEN~\cite{DOFEN2024} generalizes ensembles of oblivious decision trees, benefiting from both end-to-end gradient-based optimization and multi-layer hierarchical representation learning.

\noindent{\bf Open-Environment Tabular Machine Learning.}
Research on distribution shifts in tabular data starts with \textit{domain-to-domain shifts}~\cite{gardner2024benchmarking}, which are commonly categorized based on the availability of target domain data. When target data is available, transfer learning techniques such as unsupervised domain adaptation~\cite{sun2016deepcoral} and test-time adaptation~\cite{kim2024adaptable} are widely used. These methods adapt model parameters using test-time inputs but rely on access to target distributions, which may not always be feasible. In contrast, when target data is unavailable, a more practical but challenging scenario, methods aiming to enhance robustness and generalization, using approaches such as domain generalization~\cite{ganin2016domain}, domain robustness~\cite{sagawa2020distributionally,levy2020large}, label robustness~\cite{zhang2021coping} or ensemble strategies~\cite{gorishniy2024tabm}. TableShift~\cite{gardner2024benchmarking} provides a detailed analysis of this scenario.

Beyond domain-to-domain shifts, \textit{temporal shifts} are more general and complex. TabReD~\cite{Rubachev2024TabRed} emphasizes the inherent temporality of real-world tabular data, advocating for temporal splits for training and testing. \cite{cai2025understanding} further propose a refined training protocol focusing on temporal evaluation, significantly improving generalization across models. To address temporal shifts, it's critical to incorporate temporal information~\cite{cai2025understanding}. Drift-Resilient TabPFN~\cite{helli2024drift} models temporal shifts with a secondary SCM, which specifies changes in the primary model parameters. \cite{cai2025understanding} introduce a plug-and-play temporal embedding that effectively captures trend and periodicity patterns, providing an adaptive mechanism to mitigate the impact of temporal shifts. Under temporal shift conditions, most methods experience performance degradation, while TabM~\cite{gorishniy2024tabm} exhibits relative robustness~\cite{Rubachev2024TabRed}. However, \cite{cai2025understanding} demonstrate that with the refined training protocol and temporal embedding, methods such as ModernNCA~\cite{Ye2025ModernNCA} can regain competitiveness.

\noindent{\bf Multi-modal Learning with Tabular Data.}
Text, such as feature names, can be effectively utilized to enhance tabular data learning, as discussed in~\autoref{sec:transfer}. Here, we focus on interactions with the image modality, \eg, in healthcare, where medical images require specialized equipment and expert knowledge, often in tabular form, for accurate diagnosis~\cite{Huang23Multimodal}. To tackle challenges like large medical datasets and high annotation costs, MMCL~\cite{hager2023best} uses a contrastive self-supervised learning framework that integrates images and tabular data. CHARMS~\cite{JiangYW00Z24Tabular} transfers expert knowledge from tabular data to images, improving image predictions even without tabular data during inference, thus reducing reliance on costly expert annotations. TIP~\cite{du2024tip} proposes a self-supervised learning strategy with a tabular encoder for incomplete, heterogeneous data and a multimodal interaction module for inter-modality representation learning.

\noindent{\bf Tabular Understanding.}
Tabular understanding involves comprehending the information contained within a table and can be broken down into several tasks. For example, \textit{Table Detection} (TD)~\cite{gilani2017table, li2020tablebank} refers to identifying the region of the image that contains the table while \textit{Table Structure Recognition} (TSR)~\cite{schreiber2017deepdesrt, salaheldin2024deep} involves the identification of the rows and columns to identify individual table cells, which aims to recognize the cellular structures of tables from table images by extracting the coordinates of cell boxes and row/column spanning information. 
\textit{Table Question Answering} (TQA)~\cite{chen2020open, talmor2021multimodalqa, jin2022survey} refers to providing precise answers from tables to answer a user’s question.
Traditional methods, whether OCR-based~\cite{appalaraju2021docformer, da2023multi, gu2022xylayoutlm} or OCR-free~\cite{nassar2022tableformer, kim2021donut, feng2023unidoc, wan2024omniparser, zhao2024tabpedia}, have made significant strides in TSR and TD, which are relatively simpler tasks.

More complex tasks, such as TQA, have also been the focus of considerable effort. For example, Donut~\cite{kim2021donut} proposes a novel task and a synthetic document image generator to pre-train the model, reducing reliance on large-scale real document images. Monkey and TextMonkey~\cite{li2024monkey, liu2024textmonkey} utilize shifted window attention and use similarity measures to filter out redundant tokens.  mPLUG-DocOwl~\cite{ye2023mplugdoc} adapts mPLUG-Owl for OCR-free document understanding, while TabPedia~\cite{zhao2024tabpedia} constructs low- and high-resolution vision encoders with a concept synergy mechanism for visual table understanding. \cite{deng2024tables} focuses on exploring various table representations and directly prompting LLMs to improve performance. Please refer to \cite{jin2022survey,fang2024large} for more details.

%% file: discussions.tex
\section{Discussions}\label{sec:discussion}
In this section, we discuss several possible future directions for tabular machine learning, particularly in light of the significant potential demonstrated by tabular general/foundation models.

\noindent{\bf The Ability to Handle Dynamic and Open Environments.}  
Tabular models, particularly foundation models, will increasingly need to operate in dynamic, real-world environments where data evolves over time~\cite{Zhou2023Open}. One of the key challenges is dealing with imbalanced datasets~\cite{sauber2022use}, where certain classes may be underrepresented, and the distribution of data may shift over time~\cite{gardner2024benchmarking}. As a result, models need to adapt to these changes and continue providing accurate predictions. Additionally, the emergence of new classes in the data may require the model to evolve and update its predictions in real-time~\cite{Ren0CRWDH24Tablog}. This calls for methods that ensure tabular foundation models can accommodate evolving data, handling both new classes and changing distributions effectively.

\noindent{\bf The Coverage and Scope of Tabular Foundation Models.}  
Current tabular foundation models have demonstrated strong performance on various unseen classification and regression tasks. However, several important questions remain about their capabilities. For instance, in addition to in-context learning~\cite{BrownMRSKDNSSAA20}, are there other prediction strategies that could be employed to further enhance the versatility and performance of tabular foundation models? Beyond classification and regression, can these models be extended to handle related tasks such as clustering, imputation, outlier detection, or even table-based question answering (QA)? Expanding the task scope could increase the model’s utility in a wide range of applications. Furthermore, it is worth investigating whether there is a scaling law~\cite{Kaplan20Scalinglaw} for tabular foundation models. Currently, tabular checkpoints are relatively small compared to foundation models in other modalities, such as language or vision. Understanding the implications of scaling these models---particularly the trade-offs between model size and performance---will be crucial for their future development.

\noindent{\bf Will Foundation Models Always Help?}  
While foundation models have demonstrated impressive generalization abilities, there are inherent trade-offs. Similar to ensemble learning, a single foundation model may provide an ``average'' predictive ability across tasks, potentially losing specialized expertise for specific tasks. To address this, a promising approach could be the development of a ``tabular model zoo''~\cite{Zhou2016Learnware,Zhou2024Learnware}. In this paradigm, different pre-trained models, potentially including models from other domains, could be combined for a specific tabular task. Given a new task, suitable pre-trained models could be selected, adapted if necessary, and integrated for improved performance.

\noindent{\bf Model Efficiency.}  
In many real-world applications, tabular datasets are large and high-dimensional, posing significant challenges for both training and inference~\cite{Hu24AnnotatedTables,Ye2024Closer}. One area of concern is how to handle extreme cases, such as when the data is exceptionally large or sparse. Foundation models should be able to scale effectively in these scenarios without sacrificing performance. Another issue is inference speed. In large-scale problems, timely predictions are essential, especially when deployed in real-time environments~\cite{zeng2024tabflex}. Optimizing the inference process is therefore critical to ensure that predictions can be made quickly on large, complex datasets. Lastly, the computational resources required for training and deploying foundation models can be substantial~\cite{Zhou2023CoRE}. Optimizing resource usage through methods such as model pruning, quantization, and efficient training algorithms will be important to ensure that these models remain practical and accessible for a wide range of applications.

\noindent{\bf Bridging the Gap Between Tabular Data and Other Modalities.}  
Tabular data often coexists with other data modalities, such as images and text. One of the exciting challenges in the field is how to effectively integrate tabular data with foundation models from other domains~\cite{LiangZKYZ22mind}. Combining the strengths of tabular models with those of vision or language models could result in more powerful and versatile models capable of handling multimodal data. Exploring how to seamlessly integrate these modalities---whether through joint embeddings, cross-modal attention mechanisms, or other techniques---could unlock significant advances in tasks that require both structured tabular data and unstructured data sources like images or text.

%% file: conclusion.tex
\section{Conclusion}\label{sec:conclusion}
Tabular data remains a cornerstone of real-world machine learning applications, and the advancement of deep learning has opened new possibilities for effective representation learning in this domain.
In this survey, we present a comprehensive overview of deep tabular representation learning, covering its background, challenges, evaluation benchmarks, and the discussion between tree-based models and DNNs. We systematically categorize existing methods into three categories---specialized, transferable, and general models---based on their generalization capabilities. In addition, we discuss ensemble techniques, extensions, and some promising future directions, such as open-environment and multimodal tabular learning. We hope this survey serves as a valuable reference for understanding the current state of the field and inspires further progress in developing more robust and generalizable tabular learning methods.

%% file: main.bbl
\begin{thebibliography}{100}
\providecommand{\url}[1]{#1}
\csname url@samestyle\endcsname
\providecommand{\newblock}{\relax}
\providecommand{\bibinfo}[2]{#2}
\providecommand{\BIBentrySTDinterwordspacing}{\spaceskip=0pt\relax}
\providecommand{\BIBentryALTinterwordstretchfactor}{4}
\providecommand{\BIBentryALTinterwordspacing}{\spaceskip=\fontdimen2\font plus
\BIBentryALTinterwordstretchfactor\fontdimen3\font minus \fontdimen4\font\relax}
\providecommand{\BIBforeignlanguage}[2]{{%
\expandafter\ifx\csname l@#1\endcsname\relax
\typeout{** WARNING: IEEEtran.bst: No hyphenation pattern has been}%
\typeout{** loaded for the language `#1'. Using the pattern for}%
\typeout{** the default language instead.}%
\else
\language=\csname l@#1\endcsname
\fi
#2}}
\providecommand{\BIBdecl}{\relax}
\BIBdecl

\bibitem{kovalerchuk2005data}
B.~Kovalerchuk and E.~Vityaev, \emph{Data mining in finance: advances in relational and hybrid methods}.\hskip 1em plus 0.5em minus 0.4em\relax Springer Science \& Business Media, 2005.

\bibitem{hyland2020early}
S.~L. Hyland, M.~Faltys, M.~H{\"u}ser, X.~Lyu, T.~Gumbsch, C.~Esteban, C.~Bock, M.~Horn, M.~Moor, B.~Rieck \emph{et~al.}, ``Early prediction of circulatory failure in the intensive care unit using machine learning,'' \emph{Nature medicine}, vol.~26, no.~3, pp. 364--373, 2020.

\bibitem{romero2010educational}
C.~Romero and S.~Ventura, ``Educational data mining: a review of the state of the art,'' \emph{IEEE Transactions on Systems, Man, and Cybernetics}, vol.~40, no.~6, pp. 601--618, 2010.

\bibitem{amatriain2010data}
X.~Amatriain, A.~Jaimes, N.~Oliver, and J.~M. Pujol, ``Data mining methods for recommender systems,'' in \emph{Recommender systems handbook}.\hskip 1em plus 0.5em minus 0.4em\relax Springer, 2010, pp. 39--71.

\bibitem{tibshirani2002diagnosis}
R.~Tibshirani, T.~Hastie, B.~Narasimhan, and G.~Chu, ``Diagnosis of multiple cancer types by shrunken centroids of gene expression,'' \emph{Proceedings of the National Academy of Sciences}, vol.~99, no.~10, pp. 6567--6572, 2002.

\bibitem{ivanciuc2007applications}
O.~Ivanciuc \emph{et~al.}, ``Applications of support vector machines in chemistry,'' \emph{Reviews in computational chemistry}, vol.~23, p. 291, 2007.

\bibitem{ahmed2010empirical}
N.~K. Ahmed, A.~F. Atiya, N.~E. Gayar, and H.~El-Shishiny, ``An empirical comparison of machine learning models for time series forecasting,'' \emph{Econometric reviews}, vol.~29, no. 5-6, pp. 594--621, 2010.

\bibitem{allen2002towards}
M.~R. Allen and D.~A. Stainforth, ``Towards objective probabalistic climate forecasting,'' \emph{Nature}, vol. 419, no. 6903, pp. 228--228, 2002.

\bibitem{Borisov2024Deep}
V.~Borisov, T.~Leemann, K.~Se{\ss}ler, J.~Haug, M.~Pawelczyk, and G.~Kasneci, ``Deep neural networks and tabular data: {A} survey,'' \emph{{IEEE} Transactions Neural Networks and Learning Systems}, vol.~35, no.~6, pp. 7499--7519, 2024.

\bibitem{Aggarwal15DMBook}
C.~C. Aggarwal, \emph{Data Mining - The Textbook}.\hskip 1em plus 0.5em minus 0.4em\relax Springer, 2015.

\bibitem{Ji2014DP}
Z.~Ji, Z.~C. Lipton, and C.~Elkan, ``Differential privacy and machine learning: a survey and review,'' \emph{CoRR}, vol. abs/1412.7584, 2014.

\bibitem{DelgadoCBA14}
M.~F. Delgado, E.~Cernadas, S.~Barro, and D.~G. Amorim, ``Do we need hundreds of classifiers to solve real world classification problems?'' \emph{Journal of Machine Learning Research}, vol.~15, no.~1, pp. 3133--3181, 2014.

\bibitem{bishop2006pattern}
C.~Bishop, \emph{Pattern recognition and machine learning}.\hskip 1em plus 0.5em minus 0.4em\relax Springer, 2006.

\bibitem{HastieTF09ESL}
T.~Hastie, R.~Tibshirani, and J.~H. Friedman, \emph{The Elements of Statistical Learning: Data Mining, Inference, and Prediction, 2nd Edition}.\hskip 1em plus 0.5em minus 0.4em\relax Springer, 2009.

\bibitem{Mohri2012FoML}
M.~Mohri, A.~Rostamizadeh, and A.~Talwalkar, \emph{Foundations of Machine Learning}.\hskip 1em plus 0.5em minus 0.4em\relax {MIT} Press, 2012.

\bibitem{Murphy2012PML}
K.~P. Murphy, \emph{Probabilistic Machine Learning: An introduction}, ser. Adaptive computation and machine learning series.\hskip 1em plus 0.5em minus 0.4em\relax {MIT} Press, 2022.

\bibitem{voulodimos2018deep}
A.~Voulodimos, N.~Doulamis, A.~Doulamis, E.~Protopapadakis \emph{et~al.}, ``Deep learning for computer vision: A brief review,'' \emph{Computational intelligence and neuroscience}, vol. 2018, 2018.

\bibitem{otter2020survey}
D.~W. Otter, J.~R. Medina, and J.~K. Kalita, ``A survey of the usages of deep learning for natural language processing,'' \emph{IEEE transactions on neural networks and learning systems}, vol.~32, no.~2, pp. 604--624, 2020.

\bibitem{bengio2013representation}
Y.~Bengio, A.~Courville, and P.~Vincent, ``Representation learning: A review and new perspectives,'' \emph{IEEE transactions on pattern analysis and machine intelligence}, vol.~35, no.~8, pp. 1798--1828, 2013.

\bibitem{lecun2015deep}
Y.~LeCun, Y.~Bengio, and G.~Hinton, ``Deep learning,'' \emph{nature}, vol. 521, no. 7553, pp. 436--444, 2015.

\bibitem{goodfellow2016deep}
I.~Goodfellow, Y.~Bengio, and A.~Courville, \emph{Deep learning}.\hskip 1em plus 0.5em minus 0.4em\relax MIT press, 2016.

\bibitem{Donahue2014Decaf}
J.~Donahue, Y.~Jia, O.~Vinyals, J.~Hoffman, N.~Zhang, E.~Tzeng, and T.~Darrell, ``Decaf: {A} deep convolutional activation feature for generic visual recognition,'' in \emph{ICML}, 2014, pp. 647--655.

\bibitem{hinton2006reducing}
G.~E. Hinton and R.~R. Salakhutdinov, ``Reducing the dimensionality of data with neural networks,'' \emph{science}, vol. 313, no. 5786, pp. 504--507, 2006.

\bibitem{weston2008deep}
J.~Weston, F.~Ratle, and R.~Collobert, ``Deep learning via semi-supervised embedding,'' in \emph{ICML}, 2008, pp. 1168--1175.

\bibitem{van2009learning}
L.~Van Der~Maaten, ``Learning a parametric embedding by preserving local structure,'' in \emph{AISTATS}, 2009, pp. 384--391.

\bibitem{min2010deep}
M.~R. Min, L.~Maaten, Z.~Yuan, A.~J. Bonner, and Z.~Zhang, ``Deep supervised t-distributed embedding,'' in \emph{ICML}, 2010, pp. 791--798.

\bibitem{ZhangDW16Deep}
W.~Zhang, T.~Du, and J.~Wang, ``Deep learning over multi-field categorical data - - {A} case study on user response prediction,'' in \emph{ECIR}, 2016, pp. 45--57.

\bibitem{Cheng2016Wide}
H.-T. Cheng, L.~Koc, J.~Harmsen, T.~Shaked, T.~Chandra, H.~Aradhye, G.~Anderson, G.~Corrado, W.~Chai, M.~Ispir, R.~Anil, Z.~Haque, L.~Hong, V.~Jain, X.~Liu, and H.~Shah, ``Wide {\&} deep learning for recommender systems,'' in \emph{DLRS}, 2016, pp. 7--10.

\bibitem{mehrotra2017anomaly}
K.~G. Mehrotra, C.~K. Mohan, H.~Huang, K.~G. Mehrotra, C.~K. Mohan, and H.~Huang, \emph{Anomaly detection}.\hskip 1em plus 0.5em minus 0.4em\relax Springer, 2017.

\bibitem{isinkaye2015recommendation}
F.~O. Isinkaye, Y.~O. Folajimi, and B.~A. Ojokoh, ``Recommendation systems: Principles, methods and evaluation,'' \emph{Egyptian informatics journal}, vol.~16, no.~3, pp. 261--273, 2015.

\bibitem{Rangapuram2018DeepSSM}
S.~S. Rangapuram, M.~W. Seeger, J.~Gasthaus, L.~Stella, Y.~Wang, and T.~Januschowski, ``Deep state space models for time series forecasting,'' in \emph{NeurIPS}, 2018, pp. 7796--7805.

\bibitem{lim2021time}
B.~Lim and S.~Zohren, ``Time-series forecasting with deep learning: a survey,'' \emph{Philosophical Transactions of the Royal Society A}, vol. 379, no. 2194, p. 20200209, 2021.

\bibitem{GorishniyRKB21Revisiting}
Y.~Gorishniy, I.~Rubachev, V.~Khrulkov, and A.~Babenko, ``Revisiting deep learning models for tabular data,'' in \emph{NeurIPS}, 2021, pp. 18\,932--18\,943.

\bibitem{David2024RealMLP}
D.~Holzm{\"{u}}ller, L.~Grinsztajn, and I.~Steinwart, ``Better by default: Strong pre-tuned mlps and boosted trees on tabular data,'' in \emph{NeurIPS}, 2024, pp. 26\,577--26\,658.

\bibitem{Ye2025ModernNCA}
H.-J. Ye, H.-H. Yin, D.-C. Zhan, and W.-L. Chao, ``Revisiting nearest neighbor for tabular data: A deep tabular baseline two decades later,'' in \emph{ICLR}, 2025.

\bibitem{Grinsztajn2022Why}
L.~Grinsztajn, E.~Oyallon, and G.~Varoquaux, ``Why do tree-based models still outperform deep learning on typical tabular data?'' in \emph{NeurIPS}, 2022, pp. 507--520.

\bibitem{ZivA22Tabular}
R.~Shwartz-Ziv and A.~Armon, ``Tabular data: Deep learning is not all you need,'' \emph{Information Fusion}, vol.~81, pp. 84--90, 2022.

\bibitem{EgeInductiveBias}
E.~Beyazit, J.~Kozaczuk, B.~Li, V.~Wallace, and B.~Fadlallah, ``An inductive bias for tabular deep learning,'' in \emph{NeurIPS}, 2023, pp. 43\,108--43\,135.

\bibitem{McElfreshKVCRGW23when}
D.~C. McElfresh, S.~Khandagale, J.~Valverde, V.~P. C., G.~Ramakrishnan, M.~Goldblum, and C.~White, ``When do neural nets outperform boosted trees on tabular data?'' in \emph{NeurIPS}, 2023, pp. 76\,336--76\,369.

\bibitem{ye2019learning}
H.-J. Ye, D.-C. Zhan, N.~Li, and Y.~Jiang, ``Learning multiple local metrics: Global consideration helps,'' \emph{IEEE transactions on pattern analysis and machine intelligence}, vol.~42, no.~7, pp. 1698--1712, 2019.

\bibitem{Jesus2022Turning}
S.~M. Jesus, J.~Pombal, D.~Alves, A.~F. Cruz, P.~Saleiro, R.~P. Ribeiro, J.~Gama, and P.~Bizarro, ``Turning the tables: Biased, imbalanced, dynamic tabular datasets for {ML} evaluation,'' in \emph{NeurIPS}, 2022, pp. 33\,563--33\,575.

\bibitem{kohli2024towards}
R.~Kohli, M.~Feurer, K.~Eggensperger, B.~Bischl, and F.~Hutter, ``Towards quantifying the effect of datasets for benchmarking: A look at tabular machine learning,'' in \emph{ICLR Workshop}, 2024.

\bibitem{Tschalzev2024DataCentric}
A.~Tschalzev, S.~Marton, S.~L{\"{u}}dtke, C.~Bartelt, and H.~Stuckenschmidt, ``A data-centric perspective on evaluating machine learning models for tabular data,'' in \emph{NeurIPS Datasets and Benchmarks Track}, 2024.

\bibitem{Ye2024Closer}
H.-J. Ye, S.-Y. Liu, H.-R. Cai, Q.-L. Zhou, and D.-C. Zhan, ``A closer look at deep learning on tabular data,'' \emph{CoRR}, vol. abs/2407.00956, 2024.

\bibitem{Gorishniy2022On}
Y.~Gorishniy, I.~Rubachev, and A.~Babenko, ``On embeddings for numerical features in tabular deep learning,'' in \emph{NeurIPS}, 2022, pp. 24\,991--25\,004.

\bibitem{UcarHE21SubTab}
T.~Ucar, E.~Hajiramezanali, and L.~Edwards, ``Subtab: Subsetting features of tabular data for self-supervised representation learning,'' in \emph{NeurIPS}, 2021, pp. 18\,853--18\,865.

\bibitem{BahriJTM22Scarf}
D.~Bahri, H.~Jiang, Y.~Tay, and D.~Metzler, ``Scarf: Self-supervised contrastive learning using random feature corruption,'' in \emph{ICLR}, 2022.

\bibitem{YoonZJS20VIME}
J.~Yoon, Y.~Zhang, J.~Jordon, and M.~van~der Schaar, ``{VIME:} extending the success of self- and semi-supervised learning to tabular domain,'' in \emph{NeurIPS}, 2020, pp. 11\,033--11\,043.

\bibitem{Wu2024SwitchTab}
J.~Wu, S.~Chen, Q.~Zhao, R.~Sergazinov, C.~Li, S.~Liu, C.~Zhao, T.~Xie, H.~Guo, C.~Ji, D.~Cociorva, and H.~Brunzell, ``Switchtab: Switched autoencoders are effective tabular learners,'' in \emph{AAAI}, 2024, pp. 15\,924--15\,933.

\bibitem{Kadra2021Well}
A.~Kadra, M.~Lindauer, F.~Hutter, and J.~Grabocka, ``Well-tuned simple nets excel on tabular datasets,'' in \emph{NeurIPS}, 2021, pp. 23\,928--23\,941.

\bibitem{WangFFW17DCN}
R.~Wang, B.~Fu, G.~Fu, and M.~Wang, ``Deep {\&} cross network for ad click predictions,'' in \emph{ADKDD}, 2017, pp. 1--7.

\bibitem{KlambauerUMH17Self}
G.~Klambauer, T.~Unterthiner, A.~Mayr, and S.~Hochreiter, ``Self-normalizing neural networks,'' in \emph{NIPS}, 2017, pp. 971--980.

\bibitem{ke2018tabnn}
G.~Ke, J.~Zhang, Z.~Xu, J.~Bian, and T.-Y. Liu, ``Tabnn: A universal neural network solution for tabular data,'' 2018.

\bibitem{WangSCJLHC21DCNv2}
R.~Wang, R.~Shivanna, D.~Z. Cheng, S.~Jain, D.~Lin, L.~Hong, and E.~H. Chi, ``{DCN} {V2:} improved deep {\&} cross network and practical lessons for web-scale learning to rank systems,'' in \emph{WWW}, 2021, pp. 1785--1797.

\bibitem{ChenLWCW22DAN}
J.~Chen, K.~Liao, Y.~Wan, D.~Z. Chen, and J.~Wu, ``Danets: Deep abstract networks for tabular data classification and regression,'' in \emph{AAAI}, 2022, pp. 3930--3938.

\bibitem{Chen2023TabCaps}
J.~Chen, K.~Liao, Y.~Fang, D.~Chen, and J.~Wu, ``Tabcaps: A capsule neural network for tabular data classification with bow routing,'' in \emph{ICLR}, 2023.

\bibitem{Chen2024Team}
J.~Yan, J.~Chen, Q.~Wang, D.~Z. Chen, and J.~Wu, ``Team up gbdts and dnns: Advancing efficient and effective tabular prediction with tree-hybrid mlps,'' in \emph{KDD}, 2024, pp. 3679--3689.

\bibitem{Xu2024BiSHop}
C.~Xu, Y.-C. Huang, J.~Y.-C. Hu, W.~Li, A.~Gilani, H.-S. Goan, and H.~Liu, ``Bishop: Bi-directional cellular learning for tabular data with generalized sparse modern hopfield model,'' in \emph{ICML}, 2024, pp. 55\,048--55\,075.

\bibitem{Badirli2020GrowNet}
S.~Badirli, X.~Liu, Z.~Xing, A.~Bhowmik, and S.~S. Keerthi, ``Gradient boosting neural networks: Grownet,'' \emph{CoRR}, vol. abs/2002.07971, 2020.

\bibitem{PopovMB20Neural}
S.~Popov, S.~Morozov, and A.~Babenko, ``Neural oblivious decision ensembles for deep learning on tabular data,'' in \emph{ICLR}, 2020.

\bibitem{Chang0G22NODEGAM}
C.-H. Chang, R.~Caruana, and A.~Goldenberg, ``{NODE-GAM:} neural generalized additive model for interpretable deep learning,'' in \emph{ICLR}, 2022.

\bibitem{SongS0DX0T19AutoInt}
W.~Song, C.~Shi, Z.~Xiao, Z.~Duan, Y.~Xu, M.~Zhang, and J.~Tang, ``Autoint: Automatic feature interaction learning via self-attentive neural networks,'' in \emph{CIKM}, 2019, pp. 1161--1170.

\bibitem{Huang2020TabTransformer}
X.~Huang, A.~Khetan, M.~Cvitkovic, and Z.~S. Karnin, ``Tabtransformer: Tabular data modeling using contextual embeddings,'' \emph{CoRR}, vol. abs/2012.06678, 2020.

\bibitem{Zhou2023TabToken}
Q.-L. Zhou, H.-J. Ye, L.~Wang, and D.-C. Zhan, ``Unlocking the transferability of tokens in deep models for tabular data,'' \emph{CoRR}, vol. abs/2310.15149, 2023.

\bibitem{Chen2023Excel}
J.~Chen, J.~Yan, Q.~Chen, D.~Z. Chen, J.~Wu, and J.~Sun, ``Can a deep learning model be a sure bet for tabular prediction?'' in \emph{KDD}, 2024, pp. 288--296.

\bibitem{jeffares2023tangos}
A.~Jeffares, T.~Liu, J.~Crabb{\'e}, F.~Imrie, and M.~van~der Schaar, ``Tangos: Regularizing tabular neural networks through gradient orthogonalization and specialization,'' in \emph{ICLR}, 2023.

\bibitem{PTARL}
H.~Ye, W.~Fan, X.~Song, S.~Zheng, H.~Zhao, D.~dan Guo, and Y.~Chang, ``Ptarl: Prototype-based tabular representation learning via space calibration,'' in \emph{ICLR}, 2024.

\bibitem{NaderSL22DNNR}
Y.~Nader, L.~Sixt, and T.~Landgraf, ``{DNNR:} differential nearest neighbors regression,'' in \emph{ICML}, 2022, pp. 16\,296--16\,317.

\bibitem{gorishniy2023tabr}
Y.~Gorishniy, I.~Rubachev, N.~Kartashev, D.~Shlenskii, A.~Kotelnikov, and A.~Babenko, ``Tabr: Tabular deep learning meets nearest neighbors in 2023,'' in \emph{ICLR}, 2024.

\bibitem{Somepalli2021SAINT}
G.~Somepalli, A.~Schwarzschild, M.~Goldblum, C.~B. Bruss, and T.~Goldstein, ``{SAINT}: Improved neural networks for tabular data via row attention and contrastive pre-training,'' in \emph{NeurIPS Workshop}, 2022.

\bibitem{Rubachev2022revisiting}
I.~Rubachev, A.~Alekberov, Y.~Gorishniy, and A.~Babenko, ``Revisiting pretraining objectives for tabular deep learning,'' \emph{CoRR}, vol. abs/2207.03208, 2022.

\bibitem{Onishi2023TabRet}
S.~Onishi, K.~Oono, and K.~Hayashi, ``Tabret: Pre-training transformer-based tabular models for unseen columns,'' \emph{CoRR}, vol. abs/2303.15747, 2023.

\bibitem{shen2023cross}
J.~Shen, L.~Li, L.~M. Dery, C.~Staten, M.~Khodak, G.~Neubig, and A.~Talwalkar, ``Cross-modal fine-tuning: Align then refine,'' in \emph{ICML}, 2023, pp. 31\,030--31\,056.

\bibitem{Zhu2021IGTD}
Y.~Zhu, T.~Brettin, F.~Xia, A.~Partin, M.~Shukla, H.~Yoo, Y.~A. Evrard, J.~H. Doroshow, and R.~L. Stevens, ``Converting tabular data into images for deep learning with convolutional neural networks,'' \emph{Scientific Reports}, vol.~11, no. 11325, 2021.

\bibitem{Lee2023TablEye}
S.~Lee and S.-C. Lee, ``Tableye: Seeing small tables through the lens of images,'' \emph{CoRR}, vol. abs/2307.02491, 2023.

\bibitem{Mamdouh2025Tab2Visual}
A.~Mamdouh, M.~El-Melegy, S.~Ali, and R.~Kikinis, ``Tab2visual: Overcoming limited data in tabular data classification using deep learning with visual representations,'' \emph{CoRR}, vol. abs/2502.07181, 2025.

\bibitem{Wang2022TransTab}
Z.~Wang and J.~Sun, ``Transtab: Learning transferable tabular transformers across tables,'' in \emph{NeurIPS}, 2022, pp. 2902--2915.

\bibitem{Yan2024Making}
J.~Yan, B.~Zheng, H.~Xu, Y.~Zhu, D.~Z. Chen, J.~Sun, J.~Wu, and J.~Chen, ``Making pre-trained language models great on tabular prediction,'' in \emph{ICLR}, 2024.

\bibitem{Ye2024Towards}
C.~Ye, G.~Lu, H.~Wang, L.~Li, S.~Wu, G.~Chen, and J.~Zhao, ``Towards cross-table masked pretraining for web data mining,'' in \emph{WWW}, 2024, pp. 4449--4459.

\bibitem{Hegselmann2022TabLLM}
S.~Hegselmann, A.~Buendia, H.~Lang, M.~Agrawal, X.~Jiang, and D.~Sontag, ``Tabllm: few-shot classification of tabular data with large language models,'' in \emph{AISTATS}, 2023, pp. 5549--5581.

\bibitem{Wen2024Generative}
X.~Wen, H.~Zhang, S.~Zheng, W.~Xu, and J.~Bian, ``From supervised to generative: {A} novel paradigm for tabular deep learning with large language models,'' in \emph{SIGKDD}, 2024, pp. 3323--3333.

\bibitem{Hollmann2023CAAFE}
N.~Hollmann, S.~M{\"{u}}ller, and F.~Hutter, ``Large language models for automated data science: Introducing {CAAFE} for context-aware automated feature engineering,'' in \emph{NeurIPS}, 2023, pp. 44\,753--44\,775.

\bibitem{Han2024FeatLLM}
S.~Han, J.~Yoon, S.~{\"{O}}. Arik, and T.~Pfister, ``Large language models can automatically engineer features for few-shot tabular learning,'' in \emph{ICML}, 2024, pp. 17\,454--17\,479.

\bibitem{zhou2024comprehensive}
C.~Zhou, Q.~Li, C.~Li, J.~Yu, Y.~Liu, G.~Wang, K.~Zhang, C.~Ji, Q.~Yan, L.~He \emph{et~al.}, ``A comprehensive survey on pretrained foundation models: A history from bert to chatgpt,'' \emph{International Journal of Machine Learning and Cybernetics}, pp. 1--65, 2024.

\bibitem{LiangWNJ0SPW24FoundationTS}
Y.~Liang, H.~Wen, Y.~Nie, Y.~Jiang, M.~Jin, D.~Song, S.~Pan, and Q.~Wen, ``Foundation models for time series analysis: {A} tutorial and survey,'' in \emph{SIGKDD}, 2024, pp. 6555--6565.

\bibitem{Ye2023TabPTM}
H.-J. Ye, Q.-L. Zhou, H.-H. Yin, D.-C. Zhan, and W.-L. Chao, ``Rethinking pre-training in tabular data: A neighborhood embedding perspective,'' \emph{CoRR}, vol. abs/2311.00055, 2025.

\bibitem{BonetMGI2024HyperFast}
D.~Bonet, D.~M. Montserrat, X.~G. i~Nieto, and A.~G. Ioannidis, ``Hyperfast: Instant classification for tabular data,'' in \emph{AAAI}, 2024, pp. 11\,114--11\,123.

\bibitem{Muller2023MotherNet}
A.~M{\"{u}}ller, C.~Curino, and R.~Ramakrishnan, ``Mothernet: Fast training and inference via hyper-network transformers,'' in \emph{ICLR}, 2025.

\bibitem{Hollmann2022TabPFN}
N.~Hollmann, S.~M{\"{u}}ller, K.~Eggensperger, and F.~Hutter, ``Tabpfn: {A} transformer that solves small tabular classification problems in a second,'' in \emph{ICLR}, 2023.

\bibitem{Thomas2024LocalPFN}
V.~Thomas, J.~Ma, R.~Hosseinzadeh, K.~Golestan, G.~Yu, M.~Volkovs, and A.~L. Caterini, ``Retrieval {\&} fine-tuning for in-context tabular models,'' in \emph{NeurIPS}, 2024, pp. 108\,439--108\,467.

\bibitem{hollmann2025tabpfn}
N.~Hollmann, S.~M{\"u}ller, L.~Purucker, A.~Krishnakumar, M.~K{\"o}rfer, S.~B. Hoo, R.~T. Schirrmeister, and F.~Hutter, ``Accurate predictions on small data with a tabular foundation model,'' \emph{Nature}, vol. 637, no. 8045, pp. 319--326, 2025.

\bibitem{tabula8b}
J.~Gardner, J.~C. Perdomo, and L.~Schmidt, ``Large scale transfer learning for tabular data via language modeling,'' in \emph{NeurIPS}, 2024, pp. 45\,155--45\,205.

\bibitem{WenZZXB24From}
X.~Wen, H.~Zhang, S.~Zheng, W.~Xu, and J.~Bian, ``From supervised to generative: {A} novel paradigm for tabular deep learning with large language models,'' in \emph{{SIGKDD}}, 2024, pp. 3323--3333.

\bibitem{Wen2025ICL}
X.~Wen, S.~Zheng, Z.~Xu, Y.~Sun, and J.~Bian, ``Scalable in-context learning on tabular data via retrieval-augmented large language models,'' \emph{CoRR}, vol. abs/2502.03147, 2025.

\bibitem{gorishniy2024tabm}
Y.~Gorishniy, A.~Kotelnikov, and A.~Babenko, ``Tabm: Advancing tabular deep learning with parameter-efficient ensembling,'' \emph{CoRR}, vol. abs/2410.24210, 2024.

\bibitem{Liu2025Beta}
S.-Y. Liu and H.-J. Ye, ``Tabpfn unleashed: A scalable and effective solution to tabular classification problems,'' \emph{CoRR}, vol. abs/2502.02527, 2025.

\bibitem{SvirskyL24Interpretable}
J.~Svirsky and O.~Lindenbaum, ``Interpretable deep clustering for tabular data,'' in \emph{ICML}, 2024, pp. 47\,314--47\,330.

\bibitem{Rauf2024TableDC}
H.~T. Rauf, A.~Freitas, and N.~W. Paton, ``Tabledc: Deep clustering for tabular data,'' \emph{CoRR}, vol. abs/2405.17723, 2024.

\bibitem{Han2022ADBench}
S.~Han, X.~Hu, H.~Huang, M.~Jiang, and Y.~Zhao, ``Adbench: Anomaly detection benchmark,'' in \emph{NeurIPS}, 2022, pp. 32\,142--32\,159.

\bibitem{ShenkarW22Anomaly}
T.~Shenkar and L.~Wolf, ``Anomaly detection for tabular data with internal contrastive learning,'' in \emph{ICLR}, 2022.

\bibitem{YinQZW024MCM}
J.~Yin, Y.~Qiao, Z.~Zhou, X.~Wang, and J.~Yang, ``{MCM:} masked cell modeling for anomaly detection in tabular data,'' in \emph{ICLR}, 2024.

\bibitem{HansenSSP23Reimagining}
L.~Hansen, N.~Seedat, M.~van~der Schaar, and A.~Petrovic, ``Reimagining synthetic tabular data generation through data-centric {AI:} {A} comprehensive benchmark,'' in \emph{NeurIPS}, 2023, pp. 33\,781--33\,823.

\bibitem{HouGXQ23Incremental}
C.~Hou, S.~Gu, C.~Xu, and Y.~Qian, ``Incremental learning for simultaneous augmentation of feature and class,'' \emph{{IEEE} Transactions on pattern analysis and machine intelligence}, vol.~45, no.~12, pp. 14\,789--14\,806, 2023.

\bibitem{VeroBV24CuTS}
M.~Vero, M.~Balunovic, and M.~T. Vechev, ``Cuts: Customizable tabular synthetic data generation,'' in \emph{ICML}, 2024, pp. 49\,408--49\,433.

\bibitem{ArikP21TabNet}
S.~{\"{O}}. Arik and T.~Pfister, ``Tabnet: Attentive interpretable tabular learning,'' in \emph{AAAI}, 2021, pp. 6679--6687.

\bibitem{hager2023best}
P.~Hager, M.~J. Menten, and D.~Rueckert, ``Best of both worlds: Multimodal contrastive learning with tabular and imaging data,'' in \emph{CVPR}, 2023, pp. 23\,924--23\,935.

\bibitem{JiangYW00Z24Tabular}
J.-P. Jiang, H.-J. Ye, L.~Wang, Y.~Yang, Y.~Jiang, and D.-C. Zhan, ``Tabular insights, visual impacts: Transferring expertise from tables to images,'' in \emph{ICML}, 2024, pp. 21\,988--22\,009.

\bibitem{Diao2024OEBench}
Y.~Diao, Y.~Yang, Q.~Li, B.~He, and M.~Lu, ``Oebench: Investigating open environment challenges in real-world relational data streams,'' \emph{VLDB}, vol.~17, no.~6, pp. 1283--1296, 2024.

\bibitem{Rubachev2024TabRed}
I.~Rubachev, N.~Kartashev, Y.~Gorishniy, and A.~Babenko, ``Tabred: {A} benchmark of tabular machine learning in-the-wild,'' \emph{CoRR}, vol. abs/2406.19380, 2024.

\bibitem{gardner2024benchmarking}
J.~Gardner, Z.~Popovic, and L.~Schmidt, ``Benchmarking distribution shift in tabular data with tableshift,'' in \emph{NeurIPS}, 2024, pp. 53\,385--53\,432.

\bibitem{zhou2024core}
Z.-H. Zhou, ``Learnability with time-sharing computational resource concerns,'' \emph{National Science Review}, vol.~11, no.~10, p. nwae204, 2024.

\bibitem{jin2022survey}
N.~Jin, J.~Siebert, D.~Li, and Q.~Chen, ``A survey on table question answering: recent advances,'' in \emph{CCKS}, 2022, pp. 174--186.

\bibitem{fang2024large}
X.~Fang, W.~Xu, F.~A. Tan, J.~Zhang, Z.~Hu, Y.~Qi, S.~Nickleach, D.~Socolinsky, S.~Sengamedu, and C.~Faloutsos, ``Large language models (llms) on tabular data: Prediction, generation, and understanding--a survey,'' \emph{CoRR}, vol. abs/2402.17944, 2024.

\bibitem{winship1984regression}
C.~Winship and R.~D. Mare, ``Regression models with ordinal variables,'' \emph{American sociological review}, vol.~49, no.~4, pp. 512--525, 1984.

\bibitem{GutierrezPSFH16}
P.~A. Guti{\'{e}}rrez, M.~P{\'{e}}rez{-}Ortiz, J.~S{\'{a}}nchez{-}Monedero, F.~Fern{\'{a}}ndez{-}Navarro, and C.~Herv{\'{a}}s{-}Mart{\'{\i}}nez, ``Ordinal regression methods: Survey and experimental study,'' \emph{{IEEE} Trans. Knowl. Data Eng.}, vol.~28, no.~1, pp. 127--146, 2016.

\bibitem{Jeffares2024Telescoping}
A.~Jeffares, A.~Curth, and M.~van~der Schaar, ``Deep learning through {A} telescoping lens: {A} simple model provides empirical insights on grokking, gradient boosting {\&} beyond,'' in \emph{NeurIPS}, 2024, pp. 123\,498--123\,533.

\bibitem{Cormode2002Mining}
G.~Cormode, P.~Indyk, N.~Koudas, and S.~Muthukrishnan, ``Fast mining of massive tabular data via approximate distance computations,'' in \emph{ICDE}, 2002, pp. 605--614.

\bibitem{Adelfio2013Schema}
M.~D. Adelfio and H.~Samet, ``Schema extraction for tabular data on the web,'' \emph{VLDB}, vol.~6, no.~6, pp. 421--432, 2013.

\bibitem{Arias1996Efficient}
J.~F. Arias, A.~K. Chhabra, and V.~Misra, ``Efficient interpretation of tabular documents,'' in \emph{ICPR}, 1996, pp. 681--685.

\bibitem{Wang2000Semantic}
H.-L. Wang, S.-H. Wu, K.~K. Wang, C.-L. Sung, W.-L. Hsu, and W.-K. Shih, ``Semantic search on internet tabular information extraction for answering queries,'' in \emph{CIKM}, 2000, pp. 243--249.

\bibitem{Nederhof94Optimal}
M.-J. Nederhof, ``An optimal tabular parsing algorithm,'' in \emph{ACL}, 1994, pp. 117--124.

\bibitem{Arias1996Interpreting}
J.~F. Arias, A.~K. Chhabra, and V.~Misra, ``Interpreting and representing tabular documents,'' in \emph{CVPR}, 1996, pp. 600--605.

\bibitem{Richards2001Discovery}
G.~Richards and V.~J. Rayward-Smith, ``Discovery of association rules in tabular data,'' in \emph{ICDM}, 2001, pp. 465--472.

\bibitem{quinlan1986induction}
J.~R. Quinlan, ``Induction of decision trees,'' \emph{Machine learning}, vol.~1, pp. 81--106, 1986.

\bibitem{breiman1984classification}
L.~Breiman, J.~Friedman, R.~Olshen, and C.~J. Stone, \emph{Classification and Regression Trees}.\hskip 1em plus 0.5em minus 0.4em\relax Chapman and Hall/CRC, 1984.

\bibitem{freund1995desicion}
Y.~Freund and R.~E. Schapire, ``A desicion-theoretic generalization of on-line learning and an application to boosting,'' in \emph{EuroCOLT}, 1995, pp. 23--37.

\bibitem{Breiman01RandomForest}
L.~Breiman, ``Random forests,'' \emph{Machine Learning}, vol.~45, no.~1, pp. 5--32, 2001.

\bibitem{friedman2001greedy}
J.~H. Friedman, ``Greedy function approximation: a gradient boosting machine,'' \emph{Annals of statistics}, pp. 1189--1232, 2001.

\bibitem{friedman2002stochastic}
------, ``Stochastic gradient boosting,'' \emph{Computational statistics \& data analysis}, vol.~38, no.~4, pp. 367--378, 2002.

\bibitem{chen2016xgboost}
T.~Chen and C.~Guestrin, ``Xgboost: A scalable tree boosting system,'' in \emph{KDD}, 2016, pp. 785--794.

\bibitem{ke2017lightgbm}
G.~Ke, Q.~Meng, T.~Finley, T.~Wang, W.~Chen, W.~Ma, Q.~Ye, and T.-Y. Liu, ``Lightgbm: A highly efficient gradient boosting decision tree,'' in \emph{NIPS}, 2017, pp. 3146--3154.

\bibitem{Prokhorenkova2018Catboost}
L.~O. Prokhorenkova, G.~Gusev, A.~Vorobev, A.~V. Dorogush, and A.~Gulin, ``Catboost: unbiased boosting with categorical features,'' in \emph{NeurIPS}, 2018, pp. 6639--6649.

\bibitem{nielsen2016tree}
D.~Nielsen, ``Tree boosting with xgboost-why does xgboost win "every" machine learning competition?'' Master's thesis, NTNU, 2016.

\bibitem{makridakis2022m5}
S.~Makridakis, E.~Spiliotis, and V.~Assimakopoulos, ``M5 accuracy competition: Results, findings, and conclusions,'' \emph{International Journal of Forecasting}, vol.~38, no.~4, pp. 1346--1364, 2022.

\bibitem{larochelle2007empirical}
H.~Larochelle, D.~Erhan, A.~Courville, J.~Bergstra, and Y.~Bengio, ``An empirical evaluation of deep architectures on problems with many factors of variation,'' in \emph{ICML}, 2007, pp. 473--480.

\bibitem{salakhutdinov2007learning}
R.~Salakhutdinov and G.~Hinton, ``Learning a nonlinear embedding by preserving class neighbourhood structure,'' in \emph{AISTATS}, 2007, pp. 412--419.

\bibitem{min2009deep}
R.~Min, D.~A. Stanley, Z.~Yuan, A.~Bonner, and Z.~Zhang, ``A deep non-linear feature mapping for large-margin knn classification,'' in \emph{ICDM}, 2009, pp. 357--366.

\bibitem{ahmed2016survey}
M.~Ahmed, A.~N. Mahmood, and J.~Hu, ``A survey of network anomaly detection techniques,'' \emph{Journal of Network and Computer Applications}, vol.~60, pp. 19--31, 2016.

\bibitem{lu2012recommender}
L.~L{\"u}, M.~Medo, C.~H. Yeung, Y.-C. Zhang, Z.-K. Zhang, and T.~Zhou, ``Recommender systems,'' \emph{Physics reports}, vol. 519, no.~1, pp. 1--49, 2012.

\bibitem{salinas2020deepar}
D.~Salinas, V.~Flunkert, J.~Gasthaus, and T.~Januschowski, ``Deepar: Probabilistic forecasting with autoregressive recurrent networks,'' \emph{International journal of forecasting}, vol.~36, no.~3, pp. 1181--1191, 2020.

\bibitem{huang2025seqfusion}
T.-J. Huang, X.-Y. Chen, and H.-J. Ye, ``Seqfusion: Sequential fusion of pre-trained models for zero-shot time-series forecasting,'' \emph{CoRR}, vol. abs/2503.02836, 2025.

\bibitem{Liu2015CCPM}
Q.~Liu, F.~Yu, S.~Wu, and L.~Wang, ``A convolutional click prediction model,'' in \emph{CIKM}, 2015, pp. 1743--1746.

\bibitem{GuoTYLH17DeepFM}
H.~Guo, R.~Tang, Y.~Ye, Z.~Li, and X.~He, ``Deepfm: {A} factorization-machine based neural network for {CTR} prediction,'' in \emph{IJCAI}, 2017, pp. 1725--1731.

\bibitem{somvanshi2024survey}
S.~Somvanshi, S.~Das, S.~A. Javed, G.~Antariksa, and A.~Hossain, ``A survey on deep tabular learning,'' \emph{CoRR}, vol. abs/2410.12034, 2024.

\bibitem{lane2003introduction}
D.~Lane, D.~Scott, M.~Hebl, R.~Guerra, D.~Osherson, and H.~Zimmer, \emph{Introduction to statistics}.\hskip 1em plus 0.5em minus 0.4em\relax Citeseer, 2003.

\bibitem{karr2006data}
A.~F. Karr, A.~P. Sanil, and D.~L. Banks, ``Data quality: A statistical perspective,'' \emph{Statistical Methodology}, vol.~3, no.~2, pp. 137--173, 2006.

\bibitem{sanchez2020improving}
A.~S{\'a}nchez-Morales, J.-L. Sancho-G{\'o}mez, J.-A. Mart{\'\i}nez-Garc{\'\i}a, and A.~R. Figueiras-Vidal, ``Improving deep learning performance with missing values via deletion and compensation,'' \emph{Neural Computing and Applications}, vol.~32, pp. 13\,233--13\,244, 2020.

\bibitem{chicco2022eleven}
D.~Chicco, L.~Oneto, and E.~Tavazzi, ``Eleven quick tips for data cleaning and feature engineering,'' \emph{PLOS Computational Biology}, vol.~18, no.~12, p. e1010718, 2022.

\bibitem{luo2019autocross}
Y.~Luo, M.~Wang, H.~Zhou, Q.~Yao, W.-W. Tu, Y.~Chen, W.~Dai, and Q.~Yang, ``Autocross: Automatic feature crossing for tabular data in real-world applications,'' in \emph{KDD}, 2019, pp. 1936--1945.

\bibitem{he2009learning}
H.~He and E.~A. Garcia, ``Learning from imbalanced data,'' \emph{IEEE Transactions on knowledge and data engineering}, vol.~21, no.~9, pp. 1263--1284, 2009.

\bibitem{he2013imbalanced}
H.~He and Y.~Ma, \emph{Imbalanced learning: foundations, algorithms, and applications}.\hskip 1em plus 0.5em minus 0.4em\relax John Wiley \& Sons, 2013.

\bibitem{LinGGHD17-FocalLoss}
T.~Lin, P.~Goyal, R.~B. Girshick, K.~He, and P.~Doll{\'{a}}r, ``Focal loss for dense object detection,'' in \emph{{ICCV}}, 2017, pp. 2999--3007.

\bibitem{johnson2019survey}
J.~M. Johnson and T.~M. Khoshgoftaar, ``Survey on deep learning with class imbalance,'' \emph{Journal of big data}, vol.~6, no.~1, pp. 1--54, 2019.

\bibitem{engelmann2021conditional}
J.~Engelmann and S.~Lessmann, ``Conditional wasserstein gan-based oversampling of tabular data for imbalanced learning,'' \emph{Expert Systems with Applications}, vol. 174, p. 114582, 2021.

\bibitem{sauber2022use}
R.~Sauber-Cole and T.~M. Khoshgoftaar, ``The use of generative adversarial networks to alleviate class imbalance in tabular data: a survey,'' \emph{Journal of Big Data}, vol.~9, no.~1, p.~98, 2022.

\bibitem{liu2008exploratory}
X.-Y. Liu, J.~Wu, and Z.-H. Zhou, ``Exploratory undersampling for class-imbalance learning,'' \emph{IEEE Transactions on Systems, Man, and Cybernetics, Part B (Cybernetics)}, vol.~39, no.~2, pp. 539--550, 2008.

\bibitem{Chawla2002SMOTE}
N.~V. Chawla, K.~W. Bowyer, L.~O. Hall, and W.~P. Kegelmeyer, ``{SMOTE:} synthetic minority over-sampling technique,'' \emph{Journal of Artificial Intelligence Research}, vol.~16, pp. 321--357, 2002.

\bibitem{Alberto2018SMOTE}
A.~Fern{\'{a}}ndez, S.~Garc{\'{\i}}a, F.~Herrera, and N.~V. Chawla, ``{SMOTE} for learning from imbalanced data: Progress and challenges, marking the 15-year anniversary,'' \emph{Journal of Artificial Intelligence Research}, vol.~61, pp. 863--905, 2018.

\bibitem{cao2019learning}
K.~Cao, C.~Wei, A.~Gaidon, N.~Arechiga, and T.~Ma, ``Learning imbalanced datasets with label-distribution-aware margin loss,'' in \emph{NeurIPS}, 2019, pp. 1567--1578.

\bibitem{cui2019class}
Y.~Cui, M.~Jia, T.-Y. Lin, Y.~Song, and S.~Belongie, ``Class-balanced loss based on effective number of samples,'' in \emph{CVPR}, 2019, pp. 9268--9277.

\bibitem{xie2021fives}
Y.~Xie, Z.~Wang, Y.~Li, B.~Ding, N.~M. G{\"u}rel, C.~Zhang, M.~Huang, W.~Lin, and J.~Zhou, ``Fives: Feature interaction via edge search for large-scale tabular data,'' in \emph{SIGKDD}, 2021, pp. 3795--3805.

\bibitem{hu2024annotatedtables}
Y.~Hu, I.~Fountalis, J.~Tian, and N.~Vasiloglou, ``Annotatedtables: A large tabular dataset with language model annotations,'' \emph{CoRR}, vol. abs/2406.16349, 2024.

\bibitem{klein2019tabular}
A.~Klein and F.~Hutter, ``Tabular benchmarks for joint architecture and hyperparameter optimization,'' \emph{CoRR}, vol. abs/1905.04970, 2019.

\bibitem{pokhrel2023comparison}
P.~Pokhrel, ``A comparison of automl hyperparameter optimization tools for tabular data,'' Ph.D. dissertation, Youngstown State University, 2023.

\bibitem{hutter2019automated}
F.~Hutter, L.~Kotthoff, and J.~Vanschoren, \emph{Automated machine learning: methods, systems, challenges}.\hskip 1em plus 0.5em minus 0.4em\relax Springer Nature, 2019.

\bibitem{he2021automl}
X.~He, K.~Zhao, and X.~Chu, ``Automl: A survey of the state-of-the-art,'' \emph{Knowledge-based systems}, vol. 212, p. 106622, 2021.

\bibitem{feurer2022auto}
M.~Feurer, K.~Eggensperger, S.~Falkner, M.~Lindauer, and F.~Hutter, ``Auto-sklearn 2.0: Hands-free automl via meta-learning,'' \emph{Journal of Machine Learning Research}, vol.~23, no. 261, pp. 1--61, 2022.

\bibitem{mennella2024ethical}
C.~Mennella, U.~Maniscalco, G.~De~Pietro, and M.~Esposito, ``Ethical and regulatory challenges of ai technologies in healthcare: A narrative review,'' \emph{Heliyon}, vol.~10, no.~4, 2024.

\bibitem{moore2019HIPAA}
W.~Moore and S.~Frye, ``Review of hipaa, part 1: history, protected health information, and privacy and security rules,'' \emph{Journal of nuclear medicine technology}, vol.~47, no.~4, pp. 269--272, 2019.

\bibitem{sittig2011legal}
D.~F. Sittig and H.~Singh, ``Legal, ethical, and financial dilemmas in electronic health record adoption and use,'' \emph{Pediatrics}, vol. 127, no.~4, pp. e1042--e1047, 2011.

\bibitem{amann2020explainability}
J.~Amann, A.~Blasimme, E.~Vayena, D.~Frey, V.~I. Madai, and P.~Consortium, ``Explainability for artificial intelligence in healthcare: a multidisciplinary perspective,'' \emph{BMC medical informatics and decision making}, vol.~20, pp. 1--9, 2020.

\bibitem{caffo2022explainable}
B.~S. Caffo, F.~A. D'Asaro, A.~Garcez, and E.~Raffinetti, ``Explainable artificial intelligence models and methods in finance and healthcare,'' p. 970246, 2022.

\bibitem{guo2017calibration}
C.~Guo, G.~Pleiss, Y.~Sun, and K.~Q. Weinberger, ``On calibration of modern neural networks,'' in \emph{ICML}, 2017, pp. 1321--1330.

\bibitem{helli2024drift}
K.~Helli, D.~Schnurr, N.~Hollmann, S.~M{\"u}ller, and F.~Hutter, ``Drift-resilient tabpfn: In-context learning temporal distribution shifts on tabular data,'' in \emph{NeurIPS}, 2024, pp. 98\,742--98\,781.

\bibitem{Demsar06Statistical}
J.~Demsar, ``Statistical comparisons of classifiers over multiple data sets,'' \emph{Journal of Machine Learning Research}, vol.~7, pp. 1--30, 2006.

\bibitem{Yury2024TabM}
Y.~Gorishniy, A.~Kotelnikov, and A.~Babenko, ``Tabm: Advancing tabular deep learning with parameter-efficient ensembling,'' in \emph{ICLR}, 2025.

\bibitem{glickman1999rating}
M.~E. Glickman and A.~C. Jones, ``Rating the chess rating system,'' \emph{CHANCE-BERLIN THEN NEW YORK-}, vol.~12, pp. 21--28, 1999.

\bibitem{hvattum2010using}
L.~M. Hvattum and H.~Arntzen, ``Using elo ratings for match result prediction in association football,'' \emph{International Journal of forecasting}, vol.~26, no.~3, pp. 460--470, 2010.

\bibitem{Ma2024TabDPT}
J.~Ma, V.~Thomas, R.~Hosseinzadeh, H.~Kamkari, A.~Labach, J.~C. Cresswell, K.~Golestan, G.~Yu, M.~Volkovs, and A.~L. Caterini, ``Tabdpt: Scaling tabular foundation models,'' \emph{CoRR}, vol. abs/2410.18164, 2024.

\bibitem{tschalzev2025unreflected}
A.~Tschalzev, L.~Purucker, S.~L{\"u}dtke, F.~Hutter, C.~Bartelt, and H.~Stuckenschmidt, ``Unreflected use of tabular data repositories can undermine research quality,'' in \emph{ICLR Workshop}, 2025.

\bibitem{Attention-and-contrastive-benchmark}
S.~B. Rabbani, I.~V. Medri, and M.~D. Samad, ``Attention versus contrastive learning of tabular data - {A} data-centric benchmarking,'' \emph{CoRR}, vol. abs/2401.04266, 2024.

\bibitem{Yang2024UniTabE}
Y.~Yang, Y.~Wang, G.~Liu, L.~Wu, and Q.~Liu, ``Unitabe: {A} universal pretraining protocol for tabular foundation model in data science,'' in \emph{ICLR}, 2024.

\bibitem{Eggert2023TabLib}
G.~Eggert, K.~Huo, M.~Biven, and J.~Waugh, ``Tablib: {A} dataset of 627m tables with context,'' \emph{CoRR}, vol. abs/2310.07875, 2023.

\bibitem{deeptables}
H.~W. Jian~Yang, Xuefeng~Li, ``{DeepTables}: { A Deep Learning Python Package for Tabular Data},'' https://github.com/DataCanvasIO/DeepTables, 2022, version 0.2.x.

\bibitem{erickson2020autogluon}
N.~Erickson, J.~Mueller, A.~Shirkov, H.~Zhang, P.~Larroy, M.~Li, and A.~Smola, ``Autogluon-tabular: Robust and accurate automl for structured data,'' \emph{CoRR}, vol. abs/2003.06505, 2020.

\bibitem{PyTorch_Tabular}
M.~Joseph, ``Pytorch tabular: {A} framework for deep learning with tabular data,'' \emph{CoRR}, vol. abs/2104.13638, 2021.

\bibitem{Zaurin_pytorch-widedeep_A_flexible_2023}
J.~R. Zaurin and P.~Mulinka, ``{pytorch-widedeep: A flexible package for multimodal deep learning},'' \emph{Journal of Open Source Software}, vol.~8, no.~86, p. 5027, Jun. 2023.

\bibitem{Liu2024Talent}
S.-Y. Liu, H.-R. Cai, Q.-L. Zhou, and H.-J. Ye, ``{TALENT:} {A} tabular analytics and learning toolbox,'' \emph{CoRR}, vol. abs/2407.04057, 2024.

\bibitem{akiba2019optuna}
T.~Akiba, S.~Sano, T.~Yanase, T.~Ohta, and M.~Koyama, ``Optuna: A next-generation hyperparameter optimization framework,'' in \emph{KDD}, 2019, pp. 2623--2631.

\bibitem{morgan1989advances}
N.~Morgan and H.~Bourlard, ``Generalization and parameter estimation in feedforward nets: Some experiments,'' in \emph{NeuIPS}, 1989, pp. 630--637.

\bibitem{arlot2009survey}
S.~Arlot and A.~Celisse, ``A survey of cross-validation procedures for model selection,'' \emph{CoRR}, vol. abs/0907.4728, 2009.

\bibitem{Chen2023Trompt}
K.-Y. Chen, P.-H. Chiang, H.-R. Chou, T.-W. Chen, and T.-H. Chang, ``Trompt: Towards a better deep neural network for tabular data,'' in \emph{ICML}, 2023, pp. 4392--4434.

\bibitem{Marton2024GRANDE}
S.~Marton, S.~L{\"{u}}dtke, C.~Bartelt, and H.~Stuckenschmidt, ``{GRANDE:} gradient-based decision tree ensembles for tabular data,'' in \emph{ICLR}, 2024.

\bibitem{Jiang2024ProtoGate}
X.~Jiang, A.~Margeloiu, N.~Simidjievski, and M.~Jamnik, ``Protogate: Prototype-based neural networks with global-to-local feature selection for tabular biomedical data,'' in \emph{ICML}, 2024, pp. 21\,844--21\,878.

\bibitem{CawleyT10}
G.~C. Cawley and N.~L.~C. Talbot, ``On over-fitting in model selection and subsequent selection bias in performance evaluation,'' \emph{Journal of Machine Learning Research}, vol.~11, pp. 2079--2107, 2010.

\bibitem{Dietterich98}
T.~G. Dietterich, ``Approximate statistical tests for comparing supervised classification learning algorithms,'' \emph{Neural Computation}, vol.~10, no.~7, pp. 1895--1923, 1998.

\bibitem{Model_eval_Raschka}
S.~Raschka, ``Model evaluation, model selection, and algorithm selection in machine learning,'' \emph{CoRR}, vol. abs/1811.12808, 2018.

\bibitem{Constructing_Hornung}
H.~Schulz{-}K{\"{u}}mpel, S.~Fischer, T.~Nagler, A.~Boulesteix, B.~Bischl, and R.~Hornung, ``Constructing confidence intervals for 'the' generalization error - a comprehensive benchmark study,'' \emph{CoRR}, vol. abs/2409.18836, 2024.

\bibitem{nagler2024reshuffling}
T.~Nagler, L.~Schneider, B.~Bischl, and M.~Feurer, ``Reshuffling resampling splits can improve generalization of hyperparameter optimization,'' in \emph{NeurIPS}, 2024.

\bibitem{Feng0Z18_mGBDT}
J.~Feng, Y.~Yu, and Z.~Zhou, ``Multi-layered gradient boosting decision trees,'' in \emph{NeurIPS}, 2018, pp. 3555--3565.

\bibitem{padhi2021tabular}
I.~Padhi, Y.~Schiff, I.~Melnyk, M.~Rigotti, Y.~Mroueh, P.~Dognin, J.~Ross, R.~Nair, and E.~Altman, ``Tabular transformers for modeling multivariate time series,'' in \emph{ICASSP}, 2021, pp. 3565--3569.

\bibitem{di2023explainable}
F.~Di~Martino and F.~Delmastro, ``Explainable ai for clinical and remote health applications: a survey on tabular and time series data,'' \emph{Artificial Intelligence Review}, vol.~56, no.~6, pp. 5261--5315, 2023.

\bibitem{van2022three}
G.~M. Van~de Ven, T.~Tuytelaars, and A.~S. Tolias, ``Three types of incremental learning,'' \emph{Nature Machine Intelligence}, vol.~4, no.~12, pp. 1185--1197, 2022.

\bibitem{zhou2024class}
D.-W. Zhou, Q.-W. Wang, Z.-H. Qi, H.-J. Ye, D.-C. Zhan, and Z.~Liu, ``Class-incremental learning: A survey,'' \emph{IEEE transactions on pattern analysis and machine intelligence}, vol.~46, no.~12, pp. 9851--9873, 2024.

\bibitem{yosinski2014transferable}
J.~Yosinski, J.~Clune, Y.~Bengio, and H.~Lipson, ``How transferable are features in deep neural networks?'' in \emph{NIPS}, vol.~27, 2014.

\bibitem{dar2020transfer}
S.~U.~H. Dar, M.~{\"O}zbey, A.~B. {\c{C}}atl{\i}, and T.~{\c{C}}ukur, ``A transfer-learning approach for accelerated mri using deep neural networks,'' \emph{Magnetic resonance in medicine}, vol.~84, no.~2, pp. 663--685, 2020.

\bibitem{cao2019towards}
Y.~Cao, Z.~Fang, Y.~Wu, D.-X. Zhou, and Q.~Gu, ``Towards understanding the spectral bias of deep learning,'' \emph{CoRR}, vol. abs/1912.01198, 2019.

\bibitem{basri2020frequency}
R.~Basri, M.~Galun, A.~Geifman, D.~Jacobs, Y.~Kasten, and S.~Kritchman, ``Frequency bias in neural networks for input of non-uniform density,'' in \emph{ICML}, 2020, pp. 685--694.

\bibitem{NEURIPS2023_ac01e21b}
F.~Matteucci, V.~Arzamasov, and K.~B\"{o}hm, ``A benchmark of categorical encoders for binary classification,'' in \emph{NeurIPS}, 2023, pp. 54\,855--54\,875.

\bibitem{yan2024team}
J.~Yan, J.~Chen, Q.~Wang, D.~Z. Chen, and J.~Wu, ``Team up gbdts and dnns: Advancing efficient and effective tabular prediction with tree-hybrid mlps,'' in \emph{SIGKDD}, 2024, pp. 3679--3689.

\bibitem{PangTZZ22}
M.~Pang, K.~M. Ting, P.~Zhao, and Z.~Zhou, ``Improving deep forest by screening,'' \emph{{IEEE} Transactions on Knowledge and Data Engineering.}, vol.~34, no.~9, pp. 4298--4312, 2022.

\bibitem{Ribeiro0G16_LIME}
M.~T. Ribeiro, S.~Singh, and C.~Guestrin, ``"why should {I} trust you?": Explaining the predictions of any classifier,'' in \emph{{KDD}}, 2016, pp. 1135--1144.

\bibitem{NIPS2017_SHAP}
S.~M. Lundberg and S.~Lee, ``A unified approach to interpreting model predictions,'' in \emph{{NIPS}}, 2017, pp. 4765--4774.

\bibitem{zhou2019deep}
Z.-H. Zhou and J.~Feng, ``Deep forest,'' \emph{National science review}, vol.~6, no.~1, pp. 74--86, 2019.

\bibitem{cheng2024arithmetic}
Y.~Cheng, R.~Hu, H.~Ying, X.~Shi, J.~Wu, and W.~Lin, ``Arithmetic feature interaction is necessary for deep tabular learning,'' in \emph{AAAI}, 2024, pp. 11\,516--11\,524.

\bibitem{KossenBLGRG21Attention}
J.~Kossen, N.~Band, C.~Lyle, A.~N. Gomez, T.~Rainforth, and Y.~Gal, ``Self-attention between datapoints: Going beyond individual input-output pairs in deep learning,'' in \emph{NeurIPS}, 2021, pp. 28\,742--28\,756.

\bibitem{Bernhard2022Hopular}
B.~Sch{\"{a}}fl, L.~Gruber, A.~Bitto-Nemling, and S.~Hochreiter, ``Hopular: Modern hopfield networks for tabular data,'' \emph{CoRR}, vol. abs/2206.00664, 2022.

\bibitem{Kim2019Attentive}
H.~Kim, A.~Mnih, J.~Schwarz, M.~Garnelo, S.~M.~A. Eslami, D.~Rosenbaum, O.~Vinyals, and Y.~W. Teh, ``Attentive neural processes,'' in \emph{{ICLR}}, 2019.

\bibitem{shavitt2018regularization}
I.~Shavitt and E.~Segal, ``Regularization learning networks: deep learning for tabular datasets,'' in \emph{NeurIPS}, 2018, pp. 1386--1396.

\bibitem{Verma2021DACL}
V.~Verma, T.~Luong, K.~Kawaguchi, H.~Pham, and Q.~V. Le, ``Towards domain-agnostic contrastive learning,'' in \emph{ICML}, 2021, pp. 10\,530--10\,541.

\bibitem{Lee2022Self}
C.~Lee, F.~Imrie, and M.~van~der Schaar, ``Self-supervision enhanced feature selection with correlated gates,'' in \emph{ICLR}, 2022.

\bibitem{levin2022transfer}
R.~Levin, V.~Cherepanova, A.~Schwarzschild, A.~Bansal, C.~B. Bruss, T.~Goldstein, A.~G. Wilson, and M.~Goldblum, ``Transfer learning with deep tabular models,'' in \emph{ICLR}, 2023.

\bibitem{Majmundar2022MET}
K.~Majmundar, S.~Goyal, P.~Netrapalli, and P.~Jain, ``{MET:} masked encoding for tabular data,'' \emph{CoRR}, vol. abs/2206.08564, 2022.

\bibitem{Hajiramezanali2022STab}
E.~Hajiramezanali, N.~L. Diamant, G.~Scalia, and M.~W. Shen, ``Stab: Self-supervised learning for tabular data,'' in \emph{NeurIPS Workshop}, 2022.

\bibitem{Chen2023ReConTab}
S.~Chen, J.~Wu, N.~Hovakimyan, and H.~Yao, ``Recontab: Regularized contrastive representation learning for tabular data,'' \emph{CoRR}, vol. abs/2310.18541, 2023.

\bibitem{DDu2023DORA}
W.-W. Du, W.-Y. Wang, and W.-C. Peng, ``Dora: Domain-based self-supervised learning framework for low-resource real estate appraisal,'' in \emph{CIKM}, 2023, pp. 4552--4558.

\bibitem{Sui2024Self}
Y.~Sui, T.~Wu, J.~C. Cresswell, G.~Wu, G.~Stein, X.~S. Huang, X.~Zhang, and M.~Volkovs, ``Self-supervised representation learning from random data projectors,'' in \emph{ICLR}, 2024.

\bibitem{Iwata2020Meta}
T.~Iwata and A.~Kumagai, ``Meta-learning from tasks with heterogeneous attribute spaces,'' in \emph{NeurIPS}, 2020, pp. 6053--6063.

\bibitem{Liu2022Distribution}
L.~Liu, M.~M. Fard, and S.~Zhao, ``Distribution embedding networks for generalization from a diverse set of classification tasks,'' \emph{Transactions on Machine Learning Research}, 2022.

\bibitem{zhu2023xtab}
B.~Zhu, X.~Shi, N.~Erickson, M.~Li, G.~Karypis, and M.~Shoaran, ``Xtab: Cross-table pretraining for tabular transformers,'' in \emph{ICML}, 2023, pp. 43\,181--43\,204.

\bibitem{Zhang2023Meta}
Y.~Zhang, K.~Gong, K.~Zhang, H.~Li, Y.~Qiao, W.~Ouyang, and X.~Yue, ``Meta-transformer: {A} unified framework for multimodal learning,'' \emph{CoRR}, vol. abs/2307.10802, 2023.

\bibitem{Liu2022PTab}
G.~Liu, J.~Yang, and L.~Wu, ``Ptab: Using the pre-trained language model for modeling tabular data,'' \emph{CoRR}, vol. abs/2209.08060, 2022.

\bibitem{Kim2024CARTE}
M.~J. Kim, L.~Grinsztajn, and G.~Varoquaux, ``{CARTE:} pretraining and transfer for tabular learning,'' in \emph{ICML}, 2024, pp. 23\,843--23\,866.

\bibitem{Cheng2023Binding}
Z.~Cheng, T.~Xie, P.~Shi, C.~Li, R.~Nadkarni, Y.~Hu, C.~Xiong, D.~Radev, M.~Ostendorf, L.~Zettlemoyer, N.~A. Smith, and T.~Yu, ``Binding language models in symbolic languages,'' in \emph{ICLR}, 2023.

\bibitem{Zhang2023TapTap}
T.~Zhang, S.~Wang, S.~Yan, L.~Jian, and Q.~Liu, ``Generative table pre-training empowers models for tabular prediction,'' in \emph{EMNLP}, 2023.

\bibitem{Dinh2022LIFT}
T.~Dinh, Y.~Zeng, R.~Zhang, Z.~Lin, M.~Gira, S.~Rajput, J.~yong Sohn, D.~S. Papailiopoulos, and K.~Lee, ``{LIFT:} language-interfaced fine-tuning for non-language machine learning tasks,'' in \emph{NeurIPS}, 2022, pp. 11\,763--11\,784.

\bibitem{Wang2023UniPredict}
R.~Wang, Z.~Wang, and J.~Sun, ``Unipredict: Large language models are universal tabular predictors,'' \emph{CoRR}, vol. abs/2310.03266, 2023.

\bibitem{Sharma2019DeepInsight}
A.~Sharma, E.~Vans, D.~Shigemizu, K.~A. Boroevich, and T.~Tsunoda, ``Deepinsight: A methodology to transform a non-image data to an image for convolution neural network architecture,'' \emph{Scientific reports}, vol.~9, no.~1, p. 11399, 2019.

\bibitem{Bazgir2020REFINED}
O.~Bazgir, R.~Zhang, S.~R. Dhruba, R.~Rahman, S.~Ghosh, and R.~Pal, ``Representation of features as images with neighborhood dependencies for compatibility with convolutional neural networks,'' \emph{Nature communications}, vol.~11, no.~1, p. 4391, 2020.

\bibitem{buturovic2020novel}
L.~Buturovi{\'c} and D.~Miljkovi{\'c}, ``A novel method for classification of tabular data using convolutional neural networks,'' \emph{BioRxiv}, pp. 2020--05, 2020.

\bibitem{Vanesa2024LM-IGTD}
V.~G{\'{o}}mez-Mart{\'{\i}}nez, F.~J. Lara-Abelenda, P.~Peiro-Corbacho, D.~Chushig-Muzo, C.~Granja, and C.~Soguero-Ru{\'{\i}}z, ``{LM-IGTD:} a 2d image generator for low-dimensional and mixed-type tabular data to leverage the potential of convolutional neural networks,'' \emph{CoRR}, vol. abs/2406.14566, 2024.

\bibitem{Sun2019SuperTML}
B.~Sun, L.~Yang, W.~Zhang, M.~Lin, P.~Dong, C.~Young, and J.~Dong, ``Supertml: Two-dimensional word embedding for the precognition on structured tabular data,'' in \emph{{CVPR} Workshops}, 2019.

\bibitem{Wang2024MediTab}
Z.~Wang, C.~Gao, C.~Xiao, and J.~Sun, ``Meditab: Scaling medical tabular data predictors via data consolidation, enrichment, and refinement,'' in \emph{IJCAI}, 2024, pp. 6062--6070.

\bibitem{bommasani2021opportunities}
R.~Bommasani, D.~A. Hudson, E.~Adeli, R.~Altman, S.~Arora, S.~von Arx, M.~S. Bernstein, J.~Bohg, A.~Bosselut, E.~Brunskill \emph{et~al.}, ``On the opportunities and risks of foundation models,'' \emph{CoRR}, vol. abs/2108.07258, 2021.

\bibitem{goldberger2004neighbourhood}
J.~Goldberger, G.~E. Hinton, S.~Roweis, and R.~R. Salakhutdinov, ``Neighbourhood components analysis,'' in \emph{NIPS}, vol.~17, 2004.

\bibitem{BrownMRSKDNSSAA20}
T.~B. Brown, B.~Mann, N.~Ryder, M.~Subbiah, J.~Kaplan, P.~Dhariwal, A.~Neelakantan, P.~Shyam, G.~Sastry, A.~Askell, S.~Agarwal, A.~Herbert{-}Voss, G.~Krueger, T.~Henighan, R.~Child, A.~Ramesh, D.~M. Ziegler, J.~Wu, C.~Winter, C.~Hesse, M.~Chen, E.~Sigler, M.~Litwin, S.~Gray, B.~Chess, J.~Clark, C.~Berner, S.~McCandlish, A.~Radford, I.~Sutskever, and D.~Amodei, ``Language models are few-shot learners,'' in \emph{NeurIPS}, 2020, pp. 1877--1901.

\bibitem{tibshirani1996regression}
R.~Tibshirani, ``Regression shrinkage and selection via the lasso,'' \emph{Journal of the Royal Statistical Society Series B: Statistical Methodology}, vol.~58, no.~1, pp. 267--288, 1996.

\bibitem{hoerl1970ridge}
A.~E. Hoerl and R.~W. Kennard, ``Ridge regression: Biased estimation for nonorthogonal problems,'' \emph{Technometrics}, vol.~12, no.~1, pp. 55--67, 1970.

\bibitem{elastic-net}
H.~Zou and T.~Hastie, ``Zou h, hastie t. regularization and variable selection via the elastic net.'' \emph{Journal of the Royal Statistical Society: Series B (Statistical Methodology)}, vol.~67, pp. 301 -- 320, 2005.

\bibitem{hancock2020survey}
J.~T. Hancock and T.~M. Khoshgoftaar, ``Survey on categorical data for neural networks,'' \emph{Journal of big data}, vol.~7, no.~1, p.~28, 2020.

\bibitem{quinlan2014c4}
J.~R. Quinlan, \emph{C4.5: programs for machine learning}.\hskip 1em plus 0.5em minus 0.4em\relax Elsevier, 2014.

\bibitem{breiman2001random}
L.~Breiman, ``Random forests,'' \emph{Machine learning}, vol.~45, pp. 5--32, 2001.

\bibitem{zhou2004nec4}
Z.-H. Zhou and Y.~Jiang, ``Nec4. 5: Neural ensemble based c4. 5,'' \emph{IEEE Transactions on knowledge and data engineering}, vol.~16, no.~6, pp. 770--773, 2004.

\bibitem{hastie1986generalized}
T.~Hastie and R.~Tibshirani, ``Generalized additive models,'' \emph{Statistical science}, vol.~1, no.~3, pp. 297--310, 1986.

\bibitem{agarwal2021neural}
R.~Agarwal, L.~Melnick, N.~Frosst, X.~Zhang, B.~Lengerich, R.~Caruana, and G.~E. Hinton, ``Neural additive models: Interpretable machine learning with neural nets,'' in \emph{NeurIPS}, 2021, pp. 4699--4711.

\bibitem{Wang2025Survey}
W.-Y. Wang, W.-W. Du, D.~Xu, W.~Wang, and W.-C. Peng, ``A survey on self-supervised learning for non-sequential tabular data,'' \emph{Machine Learning}, vol. 114, no.~1, p.~16, 2025.

\bibitem{hinton2015distilling}
G.~Hinton, O.~Vinyals, and J.~Dean, ``Distilling the knowledge in a neural network,'' \emph{CoRR}, vol. abs/1503.02531, 2015.

\bibitem{Yun2019CutMix}
S.~Yun, D.~Han, S.~Chun, S.~J. Oh, Y.~Yoo, and J.~Choe, ``Cutmix: Regularization strategy to train strong classifiers with localizable features,'' in \emph{ICCV}, 2019, pp. 6023--6032.

\bibitem{Zhang2018Mixup}
H.~Zhang, M.~Ciss{\'{e}}, Y.~N. Dauphin, and D.~Lopez-Paz, ``mixup: Beyond empirical risk minimization,'' in \emph{ICLR}, 2018.

\bibitem{hou2017one}
C.~Hou and Z.-H. Zhou, ``One-pass learning with incremental and decremental features,'' \emph{IEEE transactions on pattern analysis and machine intelligence}, vol.~40, no.~11, pp. 2776--2792, 2017.

\bibitem{Ye2018ReForm}
H.-J. Ye, D.-C. Zhan, Y.~Jiang, and Z.-H. Zhou, ``Rectify heterogeneous models with semantic mapping,'' in \emph{ICML}, 2018, pp. 5630--5639.

\bibitem{ye2022revisiting}
H.-J. Ye, L.~Han, and D.-C. Zhan, ``Revisiting unsupervised meta-learning via the characteristics of few-shot tasks,'' \emph{IEEE Transactions on Pattern Analysis and Machine Intelligence}, vol.~45, no.~3, pp. 3721--3737, 2022.

\bibitem{liu2019roberta}
Y.~Liu, M.~Ott, N.~Goyal, J.~Du, M.~Joshi, D.~Chen, O.~Levy, M.~Lewis, L.~Zettlemoyer, and V.~Stoyanov, ``Roberta: A robustly optimized bert pretraining approach,'' \emph{CoRR}, vol. abs/1907.11692, 2019.

\bibitem{Mahdisoltani2015YAGO3}
F.~Mahdisoltani, J.~Biega, and F.~M. Suchanek, ``{YAGO3:} {A} knowledge base from multilingual wikipedias,'' in \emph{CIDR}, 2015.

\bibitem{hollmann2024large}
N.~Hollmann, S.~M{\"u}ller, and F.~Hutter, ``Large language models for automated data science: Introducing caafe for context-aware automated feature engineering,'' in \emph{NeurIPS}, 2023, pp. 44\,753--44\,775.

\bibitem{han2024large}
S.~Han, J.~Yoon, S.~O. Arik, and T.~Pfister, ``Large language models can automatically engineer features for few-shot tabular learning,'' in \emph{ICML}, 2024, pp. 17\,454--17\,479.

\bibitem{Herzig2020TaPas}
J.~Herzig, P.~K. Nowak, T.~M{\"{u}}ller, F.~Piccinno, and J.~M. Eisenschlos, ``Tapas: Weakly supervised table parsing via pre-training,'' in \emph{ACL}, 2020, pp. 4320--4333.

\bibitem{Yin2020TaBERT}
P.~Yin, G.~Neubig, W.~tau Yih, and S.~Riedel, ``Tabert: Pretraining for joint understanding of textual and tabular data,'' in \emph{ACL}, 2020, pp. 8413--8426.

\bibitem{Chen2024VisionTS}
M.~Chen, L.~Shen, Z.~Li, X.~J. Wang, J.~Sun, and C.~Liu, ``Visionts: Visual masked autoencoders are free-lunch zero-shot time series forecasters,'' \emph{CoRR}, vol. abs/2408.17253, 2024.

\bibitem{Li2023Time}
Z.~Li, S.~Li, and X.~Yan, ``Time series as images: Vision transformer for irregularly sampled time series,'' in \emph{NeurIPS}, 2023, pp. 49\,187--49\,204.

\bibitem{Kirillov2023Segment}
A.~Kirillov, E.~Mintun, N.~Ravi, H.~Mao, C.~Rolland, L.~Gustafson, T.~Xiao, S.~Whitehead, A.~C. Berg, W.-Y. Lo, P.~Doll{\'{a}}r, and R.~B. Girshick, ``Segment anything,'' in \emph{ICCV}, 2023, pp. 3992--4003.

\bibitem{Ha2017HyperNet}
D.~Ha, A.~M. Dai, and Q.~V. Le, ``Hypernetworks,'' in \emph{ICLR}, 2017.

\bibitem{Chao2020Meta}
W.-L. Chao, H.-J. Ye, D.-C. Zhan, M.~E. Campbell, and K.~Q. Weinberger, ``Revisiting meta-learning as supervised learning,'' \emph{CoRR}, vol. abs/2002.00573, 2020.

\bibitem{causalinferencelearningbook}
J.~Peters, D.~Janzing, and B.~Sch{\"o}lkopf, \emph{Elements of causal inference: foundations and learning algorithms}.\hskip 1em plus 0.5em minus 0.4em\relax The MIT Press, 2017.

\bibitem{neal-bayes96a}
R.~Neal, \emph{Bayesian Learning for Neural Networks}, ser. lncs.\hskip 1em plus 0.5em minus 0.4em\relax springer, 1996.

\bibitem{0005HPGH22}
S.~M{\"{u}}ller, N.~Hollmann, S.~Pineda{-}Arango, J.~Grabocka, and F.~Hutter, ``Transformers can do bayesian inference,'' in \emph{{ICLR}}, 2022.

\bibitem{ye2025closer}
H.-J. Ye, S.-Y. Liu, and W.-L. Chao, ``A closer look at tabpfn v2: Strength, limitation, and extension,'' \emph{CoRR}, vol. abs/2502.17361, 2025.

\bibitem{Iwata2023Meta}
T.~Iwata and A.~Kumagai, ``Meta-learning of semi-supervised learning from tasks with heterogeneous attribute spaces,'' \emph{CoRR}, vol. abs/2311.05088, 2023.

\bibitem{Nagler2023Statistical}
T.~Nagler, ``Statistical foundations of prior-data fitted networks,'' in \emph{ICML}, A.~Krause, E.~Brunskill, K.~Cho, B.~Engelhardt, S.~Sabato, and J.~Scarlett, Eds., 2023, pp. 25\,660--25\,676.

\bibitem{Ma2024TabPFGen}
J.~Ma, A.~Dankar, G.~Stein, G.~Yu, and A.~L. Caterini, ``Tabpfgen - tabular data generation with tabpfn,'' \emph{CoRR}, vol. abs/2406.05216, 2024.

\bibitem{RuizVillafrancaGGMM24detection}
S.~Ruiz-Villafranca, J.~R. G{\'{o}}mez, J.~M.~C. G{\'{o}}mez, J.~C. Mond{\'{e}}jar, and J.~L. Mart{\'{\i}}nez, ``A tabpfn-based intrusion detection system for the industrial internet of things,'' \emph{The Journal of Supercomputing}, vol.~80, no.~14, pp. 20\,080--20\,117, 2024.

\bibitem{TabMDA}
A.~Margeloiu, A.~Bazaga, N.~Simidjievski, P.~Li{\`{o}}, and M.~Jamnik, ``Tabmda: Tabular manifold data augmentation for any classifier using transformers with in-context subsetting,'' \emph{CoRR}, vol. abs/2406.01805, 2024.

\bibitem{Shi2024TabPFNSeries}
S.~B. Hoo, S.~Müller, D.~Salinas, and F.~Hutter, ``The tabular foundation model tabpfn outperforms specialized time series forecasting models based on simple features,'' \emph{CoRR}, vol. abs/2501.02945, 2025.

\bibitem{Breejen2025TabForestPFN}
F.~den Breejen, S.~Bae, S.~Cha, and S.-Y. Yun, ``Fine-tuned in-context learning transformers are excellent tabular data classifiers,'' \emph{CoRR}, vol. abs/2405.13396v2, 2025.

\bibitem{wu2025zeroshotmetalearningtabularprediction}
Y.~Wu and D.~L. Bergman, ``Zero-shot meta-learning for tabular prediction tasks with adversarially pre-trained transformer,'' \emph{CoRR}, vol. abs/2502.04573, 2025.

\bibitem{qu2025tabicltabularfoundationmodel}
J.~Qu, D.~Holzmüller, G.~Varoquaux, and M.~L. Morvan, ``Tabicl: A tabular foundation model for in-context learning on large data,'' \emph{CoRR}, vol. abs/2502.05564, 2025.

\bibitem{Feuer2023ScalePFN}
B.~Feuer, C.~Hegde, and N.~Cohen, ``Scaling tabpfn: Sketching and feature selection for tabular prior-data fitted networks,'' \emph{CoRR}, vol. abs/2311.10609, 2023.

\bibitem{Distillation_PFN}
J.~Ma, V.~Thomas, G.~Yu, and A.~L. Caterini, ``In-context data distillation with tabpfn,'' \emph{CoRR}, vol. abs/2402.06971, 2024.

\bibitem{Feuer2024TuneTables}
B.~Feuer, R.~T. Schirrmeister, V.~Cherepanova, C.~Hegde, F.~Hutter, M.~Goldblum, N.~Cohen, and C.~White, ``Tunetables: Context optimization for scalable prior-data fitted networks,'' in \emph{NeurIPS}, 2024, pp. 83\,430--83\,464.

\bibitem{Xu2024MixPFN}
D.~Xu, O.~Cirit, R.~Asadi, Y.~Sun, and W.~Wang, ``Mixture of in-context prompters for tabular pfns,'' \emph{CoRR}, vol. abs/2405.16156, 2024.

\bibitem{koshil2024towards}
M.~Koshil, T.~Nagler, M.~Feurer, and K.~Eggensperger, ``Towards localization via data embedding for tab{PFN},'' in \emph{NeurIPS Workshop}, 2024.

\bibitem{zeng2024tabflex}
Y.~Zeng, W.~Kang, and A.~C. Mueller, ``Tabflex: Scaling tabular learning to millions with linear attention,'' in \emph{NeurIPS Workshop}, 2024.

\bibitem{baur2024exploration}
S.~K. Baur and S.~Kim, ``Exploration of autoregressive models for in-context learning on tabular data,'' in \emph{NeurIPS Workshop}, 2024.

\bibitem{arbel2025equitabpfntargetpermutationequivariantprior}
M.~Arbel, D.~Salinas, and F.~Hutter, ``Equitabpfn: A target-permutation equivariant prior fitted networks,'' \emph{CoRR}, vol. abs/2502.06684, 2025.

\bibitem{sun2024scaling}
Y.~Sun, X.~Wen, S.~Zheng, X.~Jia, and J.~Bian, ``Scaling generative tabular learning for large language models,'' in \emph{NeurIPS Workshop}, 2024.

\bibitem{freund1996experiments}
Y.~Freund, R.~E. Schapire \emph{et~al.}, ``Experiments with a new boosting algorithm,'' in \emph{ICML}, vol.~96, 1996, pp. 148--156.

\bibitem{zhou2012ensemble}
Z.-H. Zhou, \emph{Ensemble methods: foundations and algorithms}.\hskip 1em plus 0.5em minus 0.4em\relax CRC press, 2012.

\bibitem{WenTB20BatchEnsemble}
Y.~Wen, D.~Tran, and J.~Ba, ``Batchensemble: an alternative approach to efficient ensemble and lifelong learning,'' in \emph{{ICLR}}, 2020.

\bibitem{Jayawardhana2025Boost}
M.~Jayawardhana, Renbo, S.~Dooley, V.~Cherepanova, A.~G. Wilson, F.~Hutter, C.~White, T.~Goldstein, and M.~Goldblum, ``Transformers boost the performance of decision trees on tabular data across sample sizes,'' \emph{CoRR}, vol. abs/2502.02672v2, 2025.

\bibitem{Caruana2006GreedyEnsemble}
R.~Caruana, A.~Munson, and A.~Niculescu-Mizil, ``Getting the most out of ensemble selection,'' in \emph{ICDM}, 2006, pp. 828--833.

\bibitem{wang2025priorfittednetworksscalelarger}
Y.~Wang, B.~Jiang, Y.~Guo, Q.~Gan, D.~Wipf, X.~Huang, and X.~Qiu, ``Prior-fitted networks scale to larger datasets when treated as weak learners,'' \emph{CoRR}, vol. abs/2503.01256, 2025.

\bibitem{gower1971general}
J.~C. Gower, ``A general coefficient of similarity and some of its properties,'' \emph{Biometrics}, pp. 857--871, 1971.

\bibitem{liu2008isolation}
F.~T. Liu, K.~M. Ting, and Z.-H. Zhou, ``Isolation forest,'' in \emph{ICDM}, 2008, pp. 413--422.

\bibitem{breunig2000lof}
M.~M. Breunig, H.-P. Kriegel, R.~T. Ng, and J.~Sander, ``Lof: identifying density-based local outliers,'' in \emph{SIGMOD}, 2000, pp. 93--104.

\bibitem{shenkar2022anomaly}
T.~Shenkar and L.~Wolf, ``Anomaly detection for tabular data with internal contrastive learning,'' in \emph{ICLR}, 2022.

\bibitem{li2024anomaly}
A.~Li, Y.~Zhao, C.~Qiu, M.~Kloft, P.~Smyth, M.~Rudolph, and S.~Mandt, ``Anomaly detection of tabular data using llms,'' \emph{CoRR}, vol. abs/2406.16308, 2024.

\bibitem{lee2023codi}
C.~Lee, J.~Kim, and N.~Park, ``Codi: Co-evolving contrastive diffusion models for mixed-type tabular synthesis,'' in \emph{ICML}, 2023, pp. 18\,940--18\,956.

\bibitem{tu2024causality}
R.~Tu, Z.~Senane, L.~Cao, C.~Zhang, H.~Kjellstr{\"o}m, and G.~E. Henter, ``Causality for tabular data synthesis: A high-order structure causal benchmark framework,'' \emph{CoRR}, vol. abs/2406.08311, 2024.

\bibitem{feinman2020generating}
R.~Feinman and B.~M. Lake, ``Generating new concepts with hybrid neuro-symbolic models,'' \emph{CoRR}, vol. abs/2003.08978, 2020.

\bibitem{hastie2009elements}
T.~Hastie, ``The elements of statistical learning: data mining, inference, and prediction,'' 2009.

\bibitem{greenwell2017pdp}
B.~M. Greenwell \emph{et~al.}, ``pdp: An r package for constructing partial dependence plots,'' \emph{R Journal}, vol.~9, no.~1, p. 421, 2017.

\bibitem{DOFEN2024}
K.-Y. Chen, P.-H. Chiang, H.-R. Chou, C.-S. Chen, and D.~T.-H. Chang, ``Dofen: Deep oblivious forest ensemble,'' in \emph{NeurIPS}, 2024, pp. 44\,624--44\,677.

\bibitem{sun2016deepcoral}
B.~Sun and K.~Saenko, ``Deep {CORAL:} correlation alignment for deep domain adaptation,'' in \emph{{ECCV} Workshops {(3)}}, 2016, pp. 443--450.

\bibitem{kim2024adaptable}
C.~Kim, T.~Kim, S.~Woo, J.~Y. Yang, and E.~Yang, ``Adaptable: Test-time adaptation for tabular data via shift-aware uncertainty calibrator and label distribution handler,'' \emph{CoRR}, vol. abs/2407.10784, 2024.

\bibitem{ganin2016domain}
Y.~Ganin, E.~Ustinova, H.~Ajakan, P.~Germain, H.~Larochelle, F.~Laviolette, M.~Marchand, and V.~S. Lempitsky, ``Domain-adversarial training of neural networks,'' \emph{J. Mach. Learn. Res.}, vol.~17, pp. 59:1--59:35, 2016.

\bibitem{sagawa2020distributionally}
S.~Sagawa, P.~W. Koh, T.~B. Hashimoto, and P.~Liang, ``Distributionally robust neural networks,'' in \emph{{ICLR}}, 2020.

\bibitem{levy2020large}
D.~Levy, Y.~Carmon, J.~C. Duchi, and A.~Sidford, ``Large-scale methods for distributionally robust optimization,'' in \emph{NeurIPS}, 2020, pp. 8847--8860.

\bibitem{zhang2021coping}
J.~Zhang, A.~K. Menon, A.~Veit, S.~Bhojanapalli, S.~Kumar, and S.~Sra, ``Coping with label shift via distributionally robust optimisation,'' in \emph{{ICLR}}, 2021.

\bibitem{cai2025understanding}
H.-R. Cai and H.-J. Ye, ``Understanding the limits of deep tabular methods with temporal shift,'' \emph{CoRR}, vol. abs/2502.20260, 2025.

\bibitem{Huang23Multimodal}
W.~Huang, ``Multimodal contrastive learning and tabular attention for automated alzheimer's disease prediction,'' in \emph{{ICCV} (Workshops)}, 2023, pp. 2465--2474.

\bibitem{du2024tip}
S.~Du, S.~Zheng, Y.~Wang, W.~Bai, D.~P. O’Regan, and C.~Qin, ``Tip: Tabular-image pre-training for multimodal classification with incomplete data,'' in \emph{ECCV}, 2024, pp. 478--496.

\bibitem{gilani2017table}
A.~Gilani, S.~R. Qasim, I.~Malik, and F.~Shafait, ``Table detection using deep learning,'' in \emph{ICDAR}, 2017, pp. 771--776.

\bibitem{li2020tablebank}
M.~Li, L.~Cui, S.~Huang, F.~Wei, M.~Zhou, and Z.~Li, ``Tablebank: Table benchmark for image-based table detection and recognition,'' in \emph{LREC}, 2020, pp. 1918--1925.

\bibitem{schreiber2017deepdesrt}
S.~Schreiber, S.~Agne, I.~Wolf, A.~Dengel, and S.~Ahmed, ``Deepdesrt: Deep learning for detection and structure recognition of tables in document images,'' in \emph{ICDAR}, 2017, pp. 1162--1167.

\bibitem{salaheldin2024deep}
M.~s. Kasem, A.~Abdallah, A.~Berendeyev, E.~Elkady, M.~Mahmoud, M.~Abdalla, M.~Hamada, S.~Vascon, D.~Nurseitov, and I.~Taj-eddin, ``Deep learning for table detection and structure recognition: A survey,'' \emph{ACM Computing Surveys}, vol.~56, no.~12, pp. 1--41, 2024.

\bibitem{chen2020open}
W.~Chen, M.-W. Chang, E.~Schlinger, W.~Wang, and W.~W. Cohen, ``Open question answering over tables and text,'' \emph{CoRR}, vol. abs/2010.10439, 2020.

\bibitem{talmor2021multimodalqa}
A.~Talmor, O.~Yoran, A.~Catav, D.~Lahav, Y.~Wang, A.~Asai, G.~Ilharco, H.~Hajishirzi, and J.~Berant, ``Multimodalqa: Complex question answering over text, tables and images,'' \emph{CoRR}, vol. abs/2104.06039, 2021.

\bibitem{appalaraju2021docformer}
S.~Appalaraju, B.~Jasani, B.~U. Kota, Y.~Xie, and R.~Manmatha, ``Docformer: End-to-end transformer for document understanding,'' in \emph{ICCV}, 2021, pp. 993--1003.

\bibitem{da2023multi}
C.~Da, P.~Wang, and C.~Yao, ``Multi-granularity prediction with learnable fusion for scene text recognition,'' \emph{CoRR}, vol. abs/2307.13244, 2023.

\bibitem{gu2022xylayoutlm}
Z.~Gu, C.~Meng, K.~Wang, J.~Lan, W.~Wang, M.~Gu, and L.~Zhang, ``Xylayoutlm: Towards layout-aware multimodal networks for visually-rich document understanding,'' in \emph{CVPR}, 2022, pp. 4583--4592.

\bibitem{nassar2022tableformer}
A.~Nassar, N.~Livathinos, M.~Lysak, and P.~Staar, ``Tableformer: Table structure understanding with transformers,'' in \emph{CVPR}, 2022, pp. 4614--4623.

\bibitem{kim2021donut}
G.~Kim, T.~Hong, M.~Yim, J.~Park, J.~Yim, W.~Hwang, S.~Yun, D.~Han, and S.~Park, ``Donut: Document understanding transformer without ocr,'' \emph{CoRR}, vol. abs/2111.15664, 2021.

\bibitem{feng2023unidoc}
H.~Feng, Z.~Wang, J.~Tang, J.~Lu, W.~Zhou, H.~Li, and C.~Huang, ``Unidoc: A universal large multimodal model for simultaneous text detection, recognition, spotting and understanding,'' \emph{CoRR}, vol. abs/2308.11592, 2023.

\bibitem{wan2024omniparser}
J.~Wan, S.~Song, W.~Yu, Y.~Liu, W.~Cheng, F.~Huang, X.~Bai, C.~Yao, and Z.~Yang, ``Omniparser: A unified framework for text spotting key information extraction and table recognition,'' in \emph{CVPR}, 2024, pp. 15\,641--15\,653.

\bibitem{zhao2024tabpedia}
W.~Zhao, H.~Feng, Q.~Liu, J.~Tang, S.~Wei, B.~Wu, L.~Liao, Y.~Ye, H.~Liu, W.~Zhou \emph{et~al.}, ``Tabpedia: Towards comprehensive visual table understanding with concept synergy,'' \emph{CoRR}, vol. abs/2406.01326, 2024.

\bibitem{li2024monkey}
Z.~Li, B.~Yang, Q.~Liu, Z.~Ma, S.~Zhang, J.~Yang, Y.~Sun, Y.~Liu, and X.~Bai, ``Monkey: Image resolution and text label are important things for large multi-modal models,'' in \emph{CVPR}, 2024, pp. 26\,763--26\,773.

\bibitem{liu2024textmonkey}
Y.~Liu, B.~Yang, Q.~Liu, Z.~Li, Z.~Ma, S.~Zhang, and X.~Bai, ``Textmonkey: An ocr-free large multimodal model for understanding document,'' \emph{CoRR}, vol. abs/2403.04473, 2024.

\bibitem{ye2023mplugdoc}
J.~Ye, A.~Hu, H.~Xu, Q.~Ye, M.~Yan, Y.~Dan, C.~Zhao, G.~Xu, C.~Li, J.~Tian \emph{et~al.}, ``mplug-docowl: Modularized multimodal large language model for document understanding,'' \emph{CoRR}, vol. abs/2307.02499, 2023.

\bibitem{deng2024tables}
N.~Deng, Z.~Sun, R.~He, A.~Sikka, Y.~Chen, L.~Ma, Y.~Zhang, and R.~Mihalcea, ``Tables as images? exploring the strengths and limitations of llms on multimodal representations of tabular data,'' \emph{CoRR}, vol. abs/2402.12424, 2024.

\bibitem{Zhou2023Open}
Z.-H. Zhou, ``Open-environment machine learning,'' \emph{National Science Review}, vol.~9, no.~8, p. nwac123, 07 2022.

\bibitem{Ren0CRWDH24Tablog}
W.~Ren, X.~Li, H.~Chen, V.~Rakesh, Z.~Wang, M.~Das, and V.~G. Honavar, ``Tablog: Test-time adaptation for tabular data using logic rules,'' in \emph{{ICML}}, 2024, pp. 42\,417--42\,427.

\bibitem{Kaplan20Scalinglaw}
J.~Kaplan, S.~McCandlish, T.~Henighan, T.~B. Brown, B.~Chess, R.~Child, S.~Gray, A.~Radford, J.~Wu, and D.~Amodei, ``Scaling laws for neural language models,'' \emph{CoRR}, vol. abs/2001.08361, 2020.

\bibitem{Zhou2016Learnware}
Z.-H. Zhou, ``Learnware: on the future of machine learning,'' \emph{Frontiers of Computer Science}, vol.~10, no.~4, pp. 589--590, 2016.

\bibitem{Zhou2024Learnware}
Z.-H. Zhou and Z.-H. Tan, ``Learnware: small models do big,'' \emph{Science China Information Science}, vol.~67, no.~1, 2024.

\bibitem{Hu24AnnotatedTables}
Y.~Hu, I.~Fountalis, J.~Tian, and N.~Vasiloglou, ``Annotatedtables: {A} large tabular dataset with language model annotations,'' \emph{CoRR}, vol. abs/2406.16349, 2024.

\bibitem{Zhou2023CoRE}
Z.-H. Zhou, ``Learnability with time-sharing computational resource concerns,'' \emph{National Science Review}, vol.~11, no.~10, p. nwae204, 06 2024.

\bibitem{LiangZKYZ22mind}
W.~Liang, Y.~Zhang, Y.~Kwon, S.~Yeung, and J.~Y. Zou, ``Mind the gap: Understanding the modality gap in multi-modal contrastive representation learning,'' in \emph{NeurIPS}, 2022.

\end{thebibliography}
